\documentclass{article}
\usepackage[journal=FoCM,lang=american]{ems-journal}
\makeatletter
\renewcommand*\ems@ps@titlepage@FoCM{\let\@oddhead\@empty\let\@evenhead\@empty}
\makeatother
\usepackage{microtype}
\usepackage{array}
\usepackage{algorithm}
\usepackage{algpseudocode}
\usepackage[nameinlink,capitalise]{cleveref}
\crefname{algorithm}{Algorithm}{Algorithms}

\theoremstyle{definition}
\newtheorem{theorem}{Theorem}[section]
\newtheorem{lemma}[theorem]{Lemma}
\newtheorem{proposition}[theorem]{Proposition}
\newtheorem{corollary}[theorem]{Corollary}
\newtheorem{definition}[theorem]{Definition}
\newtheorem{example}[theorem]{Example}
\newtheorem{remark}[theorem]{Remark}
\numberwithin{equation}{section}
\begin{document}
\title{Optimal recovery from moments and polynomial-time density estimation for Gaussian location mixtures}
\titlemark{Moment recovery and density estimation for Gaussian mixtures}
\emsauthor{1}{Do Ngoc Tuan}{T.~N.~Do}
\emsauthor{2}{Dao Quang Hiep}{H.~Q.~Dao}
\Emsaffil{1}{%
  \department{1}{Phenikaa School of Information Technology}%
  \organisation{1}{Phenikaa University}%
  \zip{1}{12116}%
  \city{1}{Hanoi}%
  \country{1}{Vietnam}%
  \affemail{1}{tuan.dongoc@phenikaa-uni.edu.vn}}
\Emsaffil{2}{%
  \department{1}{College of Computing and Data Science}%
  \organisation{1}{Nanyang Technological University}%
  \zip{1}{639798}%
  \city{1}{Singapore}%
  \affemail{1}{quanghiep.dao@ntu.edu.sg}}
\begin{abstract}
We study Gaussian location mixtures with a common spherical covariance from two sides: how accurately a Gaussian convolution can be recovered from finitely many moments of its mixing law, and at what cost the mixture can be learned from samples. Under an exponential tail budget of scale \(s\), Hermite truncation diverges uniformly for \(s>1\). We show that the minimax total variation error of recovery from moments through degree \(N\) has exponential base \(\sqrt{s/(s+2)}\) whenever the dimension is \(o(N)\), and we determine its exact order on the line. For mixtures of \(k\) components with arbitrary means and weights, we give an improper learner with sample and arithmetic cost polynomial in the dimension, \(k\) and the inverse accuracy, in the real-number model and in the bit model. The best previous bound without separation was quasi-polynomial. The learner combines a cover whose radius depends on the component weight, which gives every cell a leakage profile of constant variance, with a conformal summation of the local Hermite series from low-rank moment estimates.
\end{abstract}
\maketitle
\section{Introduction}\label{sec:intro}

How much information about a Gaussian mixture survives in a finite list of moments? We observe independent samples from \(F=\sum_{i=1}^k w_i N(\mu_i,I_d)\) and seek a distribution \(H\) within total variation distance \(\varepsilon\) of \(F\). Nearly coincident means and vanishing weights can make components difficult to identify while hardly changing the density. Moving mass \(w\) to distance \(D\) costs at most \(w\) in total variation and contributes \(wD^n\) to a moment of degree \(n\). Moments are the standard information for mixtures \cite{HK13}, and this imbalance governs their use.

For a component at distance \(\sqrt{sN}\) with weight \(e^{-N/2}\), the squared normalized Hermite coefficient at degree \(N\) is about \(s^N/\sqrt{2\pi N}\). For \(s>1\) it grows as the displaced mass vanishes. This raises two questions. The first is one of optimal recovery \cite{MR77,TWW88}: what is the least error of any reconstruction from the moments through degree \(N\)? The second is one of complexity: can a learner reach that accuracy from samples at polynomial cost, without forming a tensor of size \(d^N\)?

\subsection{Main results}\label{sec:results}

\textbf{Information.} Let \(\mathcal C_{s,d}(B)\) be the mixing laws \(\nu\) on \(\mathbb R^d\) with \(\int e^{|u|^2/(2s)}\,d\nu(u)\le B\), and let \(E_{N,d}(s,B)\) be the minimax total variation error for recovering \(\nu*N(0,I_d)\) from the moments of \(\nu\) through degree \(N\).

\emph{Theorem A} (\cref{thm:recovery,thm:line}). The error \(E_{N,d}(s,B)\) has exponential base \(q=\sqrt{s/(s+2)}\) whenever \(d=o(N)\), and on the line it is of exact order \(q^{N+1}/\sqrt{N+1}\). Hermite truncation has worst-case error of exact order \(s^{N/2}/\sqrt N\) and diverges for \(s>1\) (\cref{thm:criticality}).

For the upper bound we expand the Gaussian channel between a latent Gaussian of variance \(s/2\) and its observation law of variance \(1+s/2\). The tail budget controls the diagonal Mehler kernel, and a derivative of the latent Gaussian annihilates all measured polynomials and gives the matching lower exponent. On the line, extrapolating the last observed coefficient of each parity closes the prefactor gap. The Mehler reconstruction forms \(\binom{N+d}{d}\) coefficients. \Cref{prop:uniform} gives a geometric rate that is uniform in dimension, and \cref{cor:responses} treats noisy responses.

\textbf{Computation.} \emph{Theorem B} (\cref{thm:main,thm:bit}). For \(F=\sum_{i\le k}w_iN(\mu_i,I_d)\) with arbitrary means and weights, an improper learner outputs a sampler within total variation \(\varepsilon\) of \(F\) with probability \(1-\delta\). It uses \(\mathrm{poly}(d,k,1/\varepsilon)\log(1/\delta)\) samples and operations on a real-number machine, and \(\mathrm{poly}(d,k,1/\varepsilon,\beta)\log(1/\delta)\) bit operations on inputs rounded to \(O(\log(dk/\varepsilon))\) fractional bits, where \(\beta\) is the largest encoding length of a reported coordinate. The exponent is explicit (\cref{rem:exponent}). Theorem B establishes polynomial complexity for this problem in the real-number model of \cite{BCSS98} and in the bit model.

The learner multiplies each original Hermite tensor of a cell by a scalar taper (\cref{thm:taper}, \cref{prop:rational}). Each retained level is a contraction available from thinned observations, and \cref{sec:trains} estimates it in low rank. Selection and fitting use conditionally independent Gaussian views. \Cref{cor:variance} treats an unknown common spherical variance by validating candidate scales.

\textbf{Complements.} \Cref{sec:complements} gives a sharp bound for posterior centroid fits, an observable certificate for a fitted partition, positive reconstruction of the mixing law and the cumulant order that reveals the mean span under an unknown common covariance. \Cref{sec:numerics} evaluates each ingredient numerically and runs the cover at small scale.

\subsection{Comparison with the quasi-polynomial bound}\label{sec:key-step}

The best previous guarantee for this problem without separation or radius restrictions is quasi-polynomial in \(k\) \cite{DK20}, and polynomial cost is known when the means lie in a ball of radius \(O(\sqrt{\log k})\) \cite{DK24}. Combining a cover with a local fit meets the following obstruction. A fit centered at an anchor costs \(e^{O(R^2)}\) when a component may sit at distance \(R\) from its nearest anchor (\cref{cor:anchors}). A uniform radius that reaches components of weight \(2^{-U}\) forces \(R^2\gtrsim U\), so the fitting batch has size polynomial in \(2^U\). Components lighter than \(2^{-U}\) must be absent from that batch, which needs a batch much smaller than \(2^U\).

We let the radius depend on the weight. The cover of \cref{thm:cover} places every component within \(r_i^2\lesssim\log(dk/(\delta w_i))\) of an anchor. \Cref{prop:profile} turns this into a leakage profile for every cell whose variance is an absolute constant, because a component at radius \(r_i\) has weight at most \(e^{-r_i^2/164}\) up to a polynomial factor. Since the rational taper of \cref{prop:rational} has degree linear in \(\log(1/\varepsilon)\) at any fixed leak variance, the fitting batch is then polynomial in \(U\), and the coupling in the proof of \cref{thm:main} closes. \Cref{fig:roadmap} lists the steps.

\begin{figure}
\centering
\footnotesize
\[
\begin{array}{c}
\fbox{\cref{prop:cells}: leakage}\ \to\
\fbox{\cref{lem:pair-test}: pair test}\ \to\
\fbox{\cref{thm:cover}: weight-adaptive cover}\\[1.2ex]
\to\ \fbox{\cref{prop:profile}: constant leak variance}\ \to\
\fbox{\cref{prop:rational}: rational taper}\\[1.2ex]
\to\ \fbox{\cref{thm:levels}: Hermite levels}\ \to\
\fbox{\cref{cor:anchors}: fit from anchors}\ \to\
\fbox{\cref{thm:reduction}: rejection}\\[1.2ex]
\to\ \fbox{\cref{thm:main}: real-number learner}\ \to\
\fbox{\cref{thm:confidence,thm:bit}: confidence, bits}
\end{array}
\]
\caption{Logical dependence of the main line. The step that removes the dependence of the fitting cost on the weight cutoff is \cref{prop:profile}.}\label{fig:roadmap}
\end{figure}

\subsection{Relation to earlier work}\label{sec:related}

Gaussian mixture algorithms under separation \cite{Das99,VW04} preceded density estimation without separation \cite{FOS06} and polynomial learnability for fixed \(k\) \cite{MV10}. Parameter recovery admits polynomial-time algorithms at polynomial separation \cite{KSS18,DKS18} and separation of order \(\log^{1/2+c} k\) under weight conditions \cite{LL22}. Its lower bounds and those for low-degree clustering address stronger targets \cite{RV17,EGV24}. For arbitrary covariances, robust learning is polynomial only for fixed \(k\) \cite{BDJ26}, and statistical-query lower bounds show that an exponent growing with \(k\) is then unavoidable \cite{DKS17}. A bound polynomial jointly in \(k\) therefore requires an assumption such as the common spherical covariance made here.

For density estimation with one common identity covariance, fix a ball of radius \(R\) and at most \(k\) components. Doss, Wu, Yang and Zhou proved the minimax Hellinger rate \(\min(1,\sqrt{d/n})\) up to constants depending on \(k,R\), with an efficient sharp estimator for \(k=2\) \cite{DWYZ23}. He and Cheng attain the rate for every fixed \(k,R\), with arbitrary weights and no separation \cite{HC26}. Their Hermite contractions identify a subspace in which they fit a proper mixture with at most \(k\) atoms. Their arithmetic cost is polynomial in \(n,d\) for fixed \(k,R\), with an exponent that grows exponentially in \(k\) through the final search. Hellinger accuracy implies TV accuracy up to a universal factor. \Cref{thm:main} returns an improper sampler, allows unbounded means and gives a single polynomial sample and arithmetic bound in \(d,k,1/\varepsilon\). The two results are complementary: \cite{HC26} is proper and rate-sharp at fixed \(k\), and \cref{thm:main} is polynomial jointly in \(k\).

For the variable-\(k\) TV problem, sample bounds are near \(dk/\varepsilon^2\) up to logarithmic factors \cite{SOAJ14,ABH+20}. Diakonikolas and Kane gave a quasi-polynomial unrestricted algorithm \cite{DK20}. Their implicit tensor algorithm has joint polynomial cost when all means lie in a ball of radius \(O(\sqrt{\log k})\) \cite{DK24}. A diffusion sampler attains TV accuracy in quasi-polynomial time under a minimum-weight condition \cite{GKL25}, and score matching handles more general covariances under parameter bounds \cite{CKS25}. Chewi and coauthors obtain quasi-polynomial PAC density evaluation for Gaussian location mixtures \cite{CKMM26}. In one dimension Compton and Li allow arbitrary variances and obtain squared Hellinger error of order \(k/n\) up to logarithmic factors in near-linear time, with an improper piecewise polynomial density \cite{CL26}.

Moment comparison and denoising control the latent and smoothed laws \cite{WY20,DWYZ23}. Finite-atomic approximation bounds the number of components needed when the whole mixing law is available \cite{MWY25}. \Cref{cor:responses} gives an explicit coefficient estimator under a radial exponential-moment budget. When the responses have the noise level of the target (\(t=1\), so \(\gamma=1/2\)), its expected TV error is of order \((\log n)^{d/2}n^{-1/2}\), and for noisier responses the exponent \(\gamma\) is smaller. In the terms of information-based complexity \cite{TWW88}, \cref{sec:recovery} measures information cost in supplied moment degrees, and \crefrange{sec:machine}{sec:bit} measure the combinatorial cost of learning and evaluating local tensors.

\section{Statistical model and preliminaries}\label{sec:model}

\subsection{Learning protocol}\label{sec:protocol}

We observe independent draws from \(F=\sum_{i=1}^kw_iN(\mu_i,I_d)\), with \(w_i\ge0\), \(\sum_iw_i=1\) and \(\mu_i\in\mathbb R^d\), where the locations may coincide. With probability at least \(1-\delta\) over training and fitting randomness, the conditional output law \(H\) must satisfy \(d_{\mathrm{TV}}(F,H)\le\varepsilon\). For densities, \(d_{\mathrm{TV}}(f,g)=\tfrac12\int|f-g|\).

This is density estimation in the probably approximately correct sense. The learner is \emph{improper}: \(H\) need not be a mixture of \(k\) Gaussians. Its error has two sources. The \emph{approximation error} is the distance from \(F\) to the best output the method can represent, here governed by a polynomial degree \(A\). The \emph{estimation error} comes from fitting that output on finitely many samples, and its sample cost grows exponentially in \(A\). The paper is organized around keeping \(A\) logarithmic in \(dk/\varepsilon\).

For the cost, the real-arithmetic input is \(d,k,\varepsilon,\delta\) and a sample oracle. We count training and sampling costs separately. Whitening a known common positive definite covariance preserves TV. \Cref{cor:variance} treats one unknown common variance. \Cref{sec:bit} measures precision in units of the known noise standard deviation.

\subsection{Notation}\label{sec:notation}

\(\gamma=N(0,I_d)\) and \(\gamma_a=N(a,I_d)\), used both for the measure and its density. Write \(g_t\) for the density of \(N(0,tI_d)\) and \(\Phi\) for the standard normal distribution function. \(\operatorname{He}_n\) is the probabilists' Hermite polynomial and \(h_n=\operatorname{He}_n/\sqrt{n!}\). For \(h\in\mathbb R^d\), \(\operatorname{He}_n(h)\) also denotes the Hermite tensor, defined by \(\langle\delta^{\otimes n},\operatorname{He}_n(h)\rangle=|\delta|^n\operatorname{He}_n(\langle\delta,h\rangle/|\delta|)\) for \(\delta\ne0\). Tensors carry the Frobenius norm \(|\cdot|\). The logarithm \(\log\) is natural.

\Cref{tab:notation} lists the symbols used with one meaning across sections. All other letters are local, as are these letters wherever a statement redefines them explicitly. Each section or proof introduces its local letters before use, and a letter that recurs in another section need not denote the same quantity. For example, \(N\) is a moment degree in \cref{sec:criticality,sec:recovery} and a batch size from \cref{sec:trains} on. The letter \(m\) equals \(N+1\) in \cref{sec:recovery}, is most often a cell index, and also serves as a matrix size, a pair count and a moment order where stated. The letter \(n\) is a Hermite level, the degree of the pair test in \cref{sec:cover}, or a sample size in \cref{sec:complements}.

\begin{table}
\centering\small
\begin{tabular*}{\linewidth}{@{\extracolsep{\fill}}l>{\raggedright\arraybackslash}p{.7\linewidth}@{}}
\toprule
Symbol & Meaning \\
\midrule
\(d,k,\varepsilon,\delta\) & dimension, component count, TV accuracy, failure probability \\
\(w_i,\mu_i,F\) & weights, means and the target mixture \\
\(\gamma,\gamma_a,g_t,\varphi,\Phi\) & Gaussian laws and densities, standard normal density and distribution function \\
\(\nu,r_\nu\) & displacement (mixing) measure and its density ratio \\
\(s,B,\mathcal C_{s,d}(B)\) & tail scale, tail budget and the class \(\int e^{|u|^2/(2s)}d\nu\le B\) \\
\((K,R,v')\) & profile of a cell law: constant, radius and leak variance \\
\(a_m,r_i,\pi_i(m),P(m)\) & anchors, component radii, assignment probabilities, cell masses \\
\(M_n,Y_n,\Lambda_n,A\) & moment tensors, Hermite levels, taper weights and taper degree \\
\(p\) & the odd integer of the rational taper, \(2^p\ge8v'+1\) \\
\(\lg,\exp^+\) & the one-sided approximants of \cref{lem:approximants} \\
\(S_0,\rho_g,Q_c,v_\ast\) & resolution, norm gate, cover constant \(2dk/\delta_c\) and the leak variance \(90\) of \cref{sec:cover} \\
\(U\) & the weight cutoff \(2^{-U}\) \\
\(\varsigma\) & standard deviation of a response view \\
\bottomrule
\end{tabular*}
\caption{Global notation. Symbols not listed are local to a section or a proof.}\label{tab:notation}
\end{table}

For a probability measure \(\nu\) on \(\mathbb R^d\) (a \emph{displacement measure}), its \emph{density ratio} is

\[
r_\nu(h)=\frac{d(\nu*\gamma)}{d\gamma}(h)=\int e^{\langle\delta,h\rangle-|\delta|^2/2}\,d\nu(\delta).
\]

We say that \(\nu=\sum_l\rho_l\delta_{\delta_l}\) has \emph{profile} \((K,R,v')\) if it has at most \(k\) atoms and

\[
\rho_l\le K\exp\big(-(|\delta_l|-R)_+^2/(2v')\big)\qquad\text{for all }l,
\]

with \(K\ge1\), \(R\ge0\) and \emph{leak variance} \(v'\ge8\).

\subsection{Thinning and cells}\label{sec:thinning}

We first split each observation into two views. For each sample \(X\) draw \(g\sim\gamma\) independently and set \(X^{(1)}=X+g\), \(X^{(2)}=X-g\). Gaussian data fission and thinning construct independent views for selective inference \cite{LDWR23,NDGW24}. Thinning is sample splitting inside one observation: \(X^{(1)}\) selects a cell and \(X^{(2)}\) is used for estimation, so selection does not bias the estimate. The independence holds conditional on the latent component.

The first view assigns the observation to a cell. An \emph{anchor list} is a finite set \(a_1,\dots,a_M\in\mathbb R^d\) computed from data independent of the samples being processed. The \emph{radius} of component \(i\) is \(r_i=\min_m|\mu_i-a_m|\), and the list is an \emph{\(R\)-cover} if \(r_i\le R\) for all \(i\). The radius will depend on the component weight, as in list-decodable Gaussian mean estimation \cite{DKS18}. A sample is assigned to the cell

\[
m(X)=\arg\min_m|X^{(1)}-a_m|
\]

(ties broken by index). Write \(\pi_i(m)=\Pr[m(X)=m\mid X\text{ from component }i]\), \(P(m)=\sum_iw_i\pi_i(m)\), and for \(P(m)>0\)

\[
\rho^m_i=\frac{w_i\pi_i(m)}{P(m)},\qquad \nu_m=\sum_i\rho_i^m\,\delta_{\mu_i-a_m},\qquad F_m=\nu_m*\gamma_{a_m}=\sum_i\rho_i^mN(\mu_i,I).
\]

\begin{proposition}[thinned cells]\label{prop:cells}

\begin{enumerate}
\item
  Given that \(X\) comes from component \(i\), \(X^{(1)}\) and \(X^{(2)}\) are independent, each \(N(\mu_i,2I)\).
\item
  Given \(m(X)=m\), the law of \(X^{(2)}\) is \(\sum_i\rho_i^mN(\mu_i,2I)\). Moreover \(F=\sum_mP(m)F_m\).
\item
  For every component and every cell, \(\pi_i(m)\le\exp\!\big(-(|\mu_i-a_m|-r_i)_+^2/16\big)\). In particular, if the anchors are an \(R\)-cover then \(\nu_m\) has profile \((1/P(m),R,8)\).
\end{enumerate}

\end{proposition}

\begin{proof} (1) Given the component, \(X=\mu_i+z\) with \(z\sim\gamma\) independent of \(g\). The pair \((z+g,z-g)\) is jointly Gaussian with cross-covariance \(I-I=0\) and each coordinate block has covariance \(2I\).

(2) The event \(\{m(X)=m\}\) is a function of \(X^{(1)}\) and the anchors. By (1), conditionally on the component it is independent of \(X^{(2)}\), so \(\Pr[i\mid m]=\rho^m_i\) and the law of \(X^{(2)}\) given \((i,m)\) is \(N(\mu_i,2I)\). For the second claim, \(\sum_mP(m)F_m=\sum_i w_i\big(\sum_m\pi_i(m)\big)N(\mu_i,I)=F\).

(3) Write \(R=r_i\), let \(D=|\mu_i-a_m|>R\) and let \(a_p\) be an anchor with \(|\mu_i-a_p|=R\). If \(m(X)=m\) then \(|X^{(1)}-a_m|\le|X^{(1)}-a_p|\), that is \(\langle X^{(1)}-\tfrac{a_m+a_p}2,\,a_m-a_p\rangle\ge0\). Write \(X^{(1)}=\mu_i+\sqrt2\,\zeta\) with \(\zeta\sim\gamma\) and \(u=(a_m-a_p)/|a_m-a_p|\). Since \(\langle\mu_i-\tfrac{a_m+a_p}2,a_m-a_p\rangle=\tfrac12(|\mu_i-a_p|^2-|\mu_i-a_m|^2)\le\tfrac12(R^2-D^2)\) and \(|a_m-a_p|\le D+R\), the event requires

\[
\langle\zeta,u\rangle\ \ge\ \frac{D^2-R^2}{2\sqrt2\,(D+R)}=\frac{D-R}{2\sqrt2}.
\]

The Gaussian tail bound \(\Pr[N(0,1)\ge t]\le e^{-t^2/2}\) gives \(e^{-(D-R)^2/16}\). 
\end{proof}

Part 2 reduces learning \(F\) to learning each cell law \(F_m\) from independent responses with law \(\nu_m*N(a_m,2I)\). We call an assignment of component \(i\) to a cell other than that of its nearest anchor \emph{foreign}, and its probability a \emph{leak}. Part 3 bounds the leak at distances beyond \(r_i\). \Cref{sec:cover} bounds \(r_i^2\) by \(O(\log(dk/(\delta w_i)))\), giving a common leak variance (\cref{prop:profile}).

\begin{lemma}[multi-thinning]\label{lem:multi-thinning}
Given \(v\), let \(Y\sim N(v,s^2I)\) and draw \(g_1,\dots,g_J\) independently from \(\gamma\) and independently of \(Y\). With \(\bar g\) their average, the \emph{parts} \(Y^{(i)}=Y+s\sqrt J(g_i-\bar g)\) are independent \(N(v,Js^2I)\) given \(v\).
\end{lemma}

\begin{proof} Given \(v\), the parts are jointly Gaussian with cross-covariance \(s^2I+Js^2(\mathbf1_{\{i=j\}}-1/J)I=Js^2\mathbf1_{\{i=j\}}I\).\end{proof}

\section{The criticality barrier}\label{sec:criticality}

The Hermite series of a density ratio \(r_\nu\) has coefficient tensors \(T_n=\sum_l\rho_l\delta_l^{\otimes n}/\sqrt{n!}\) and energy \(\sum_n\|T_n\|^2=\sum_{l,l'}\rho_l\rho_{l'}e^{\langle\delta_l,\delta_{l'}\rangle}\). A component at distance \(D\) with weight \(\rho\) contributes \(\rho^2e^{D^2}\), so uniform control in \(D\) requires \(\rho=O(e^{-D^2/2})\). We call this decay rate \emph{critical}. Truncating the series at degree \(N\) is the orthogonal-series density estimator, and its approximation error is the tail of the series. For a fixed finite mixture that tail tends to zero in \(L^2(\gamma)\) \cite{Hil40}. A learner needs the tail small uniformly over the cell laws it may meet, in \(L^1(\gamma)\), which is twice the total variation distance.

\begin{theorem}[criticality barrier]\label{thm:criticality}

\textbf{(a) A set that holds mass \(p\) of one Gaussian holds mass at least \(\Phi(\Phi^{-1}(p)-D/\sigma)\) of a translate at distance \(D\).} Let \(\sigma>0\), \(\mu,\mu'\in\mathbb R^d\) with \(|\mu-\mu'|=D\), and let \(A\subseteq\mathbb R^d\) be measurable with \(N(\mu,\sigma^2I)(A)=p\in(0,1)\). Then

\[
N(\mu',\sigma^2I)(A)\ \ge\ \Phi\big(\Phi^{-1}(p)-D/\sigma\big).
\]

In particular, if \(p\ge1/2\) then \(N(\mu',\sigma^2I)(A)\ge\Phi(-D/\sigma)\ge\dfrac{D/\sigma}{1+D^2/\sigma^2}\cdot\dfrac{e^{-D^2/(2\sigma^2)}}{\sqrt{2\pi}}\).

\textbf{(b) Truncation has worst-case \(L^1\) error of exact order \(s^{N/2}/\sqrt N\).} For \(s>0\) and \(B>1\) let \(\mathcal C_{s,d}(B)\) be the Borel probability measures on \(\mathbb R^d\) such that \(\int e^{|\delta|^2/(2s)}d\nu\le B\). Define \(S_Nr_\nu\) by integrating the degree-\(N\) Hermite truncation of each translated kernel against \(\nu\), which only requires its finite moments. There are constants \(0<c\le C\) depending only on \((s,B)\) such that for all \(d,N\ge1\)

\[
c\,\frac{s^{N/2}}{\sqrt N}\ \le\ \sup_{\nu\in\mathcal C_{s,d}(B)}\big\|r_\nu-S_Nr_\nu\big\|_{L^1(\gamma)}\ \le\ C\,\frac{s^{N/2}}{\sqrt N}.
\]

The lower bound is witnessed by two atoms, \(\nu=(1-\lambda)\delta_0+\lambda\delta_{De_1}\) with \(D=\sqrt{sN}\) and \(\lambda=\min(1,B-1)\,e^{-N/2}\), for which \(\|r_\nu-1\|_{L^1(\gamma)}\le2\lambda\).

\end{theorem}

\begin{figure}
\centering
\includegraphics[width=0.67\textwidth]{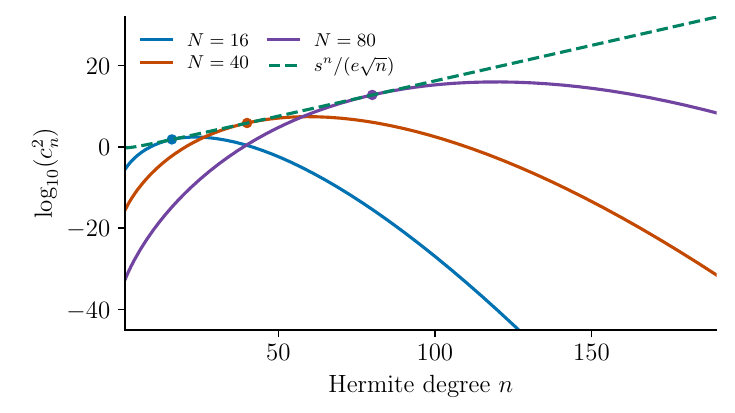}
\caption{Squared Hermite coefficients \(c_n^2=\lambda^2D^{2n}/n!\) for the two-atom law in \cref{thm:criticality}(b), with \(s=1.5\) and \(B\ge2\). Dots mark \(n=N=D^2/s\), where the curves exceed \(s^n/(e\sqrt n)\). Each curve peaks near \(n=D^2\).}\label{fig:energy}
\end{figure}

\begin{proof} (a) At \(D=0\) both probabilities equal \(p\). For \(D>0\) let \(P_0=N(\mu,\sigma^2I)\), \(P_1=N(\mu',\sigma^2I)\), \(u=(\mu'-\mu)/D\) and \(T(x)=\langle x-\mu,u\rangle/\sigma\). Under \(P_0\), \(T\sim N(0,1)\), under \(P_1\), \(T\sim N(D/\sigma,1)\), and \(\frac{dP_1}{dP_0}=\exp(\tfrac D\sigma T-\tfrac{D^2}{2\sigma^2})=:L\) is increasing in \(T\). Let \(t_p=\Phi^{-1}(p)\), \(H=\{T\le t_p\}\) and \(L^\ast=\exp(\tfrac D\sigma t_p-\tfrac{D^2}{2\sigma^2})\). On \(A\setminus H\) we have \(L\ge L^\ast\) and on \(H\setminus A\) we have \(L\le L^\ast\), so \((1_A-1_H)(L-L^\ast)\ge0\) pointwise. Integrating against \(P_0\),

\[
P_1(A)-P_1(H)\ \ge\ L^\ast\big(P_0(A)-P_0(H)\big)=0 .
\]

Finally \(P_1(H)=\Pr[N(D/\sigma,1)\le t_p]=\Phi(t_p-D/\sigma)\). For \(p\ge1/2\), \(t_p\ge0\), and the last inequality is the Mills-ratio bound \(\Phi(-t)\ge\frac{t}{1+t^2}\frac{e^{-t^2/2}}{\sqrt{2\pi}}\), \(t>0\).

(b) \emph{Upper bound.} Put \(m=N+1\). For one atom \(\delta\) with \(|\delta|=D\), the kernel \(e^{\langle\delta,h\rangle-D^2/2}\) depends on \(h\) only through \(x=\langle\delta,h\rangle/D\), and truncation by degree commutes with rotations, so its truncation error is one-dimensional. Since \(\operatorname{He}_n\varphi=(-1)^n\varphi^{(n)}\), the generating function \(e^{Dx-D^2/2}=\sum_{n\ge0}D^n\operatorname{He}_n(x)/n!\) shows that the truncated kernel times \(\varphi\) is the Taylor polynomial of degree \(N\) of \(D\mapsto\varphi(x-D)\). The integral form of the remainder bounds the \(L^1(dx)\) error by \(D^m\|\varphi^{(m)}\|_1/m!\). By Plancherel's theorem \(2\pi\|\varphi^{(m)}\|_2^2=\Gamma(m+\tfrac12)\) and \(2\pi\|x\varphi^{(m)}\|_2^2=(1+\tfrac1{4m-2})\Gamma(m+\tfrac12)\), so the Cauchy--Schwarz inequality with weight \(1+x^2\) gives \(\|\varphi^{(m)}\|_1\le C_0\,\Gamma(m+\tfrac12)^{1/2}\le C_0\,m^{-1/4}\sqrt{m!}\). Integrate over \(\nu\) and use \(\int|\delta|^md\nu\le B\max_{D\ge0}D^me^{-D^2/(2s)}=B(sm/e)^{m/2}\). Stirling's formula then gives \(\|r_\nu-S_Nr_\nu\|_{L^1(\gamma)}\le C_1B\,s^{m/2}/\sqrt m\).

\emph{Lower bound.} For the two-atom law, \(\int e^{|\delta|^2/(2s)}d\nu=1-\lambda+\lambda e^{N/2}\le B\) and \(\|r_\nu-1\|_{L^1(\gamma)}\le\lambda+\lambda\,\mathbb{E}_\gamma[e^{Dh_1-D^2/2}]=2\lambda\). Test the signed error against the bounded function \(e^{itx}\) with \(t=\sqrt N\). Since \(\int \operatorname{He}_n(x)e^{itx}d\gamma(x)=(it)^ne^{-t^2/2}\), with \(a=Dt=\sqrt s\,N\),

\[
\begin{aligned}
\big\|r_\nu-S_Nr_\nu\big\|_{L^1(\gamma)} &\ge W_{N,s}:=\lambda e^{-N/2}\Big|e^{ia}-\sum_{j\le N}\frac{(ia)^j}{j!}\Big|\\ &\hphantom{{}\ge W_{N,s}{}}=\lambda e^{-N/2}\,\frac{a^{N+1}}{N!}\,\Big|\int_0^1(1-u)^Ne^{iau}\,du\Big|.
\end{aligned}
\]

Integration by parts shows that its imaginary part is \(\tfrac Na\int_0^1(1-u)^{N-1}(1-\cos au)\,du>0\). By dominated convergence \(N\int_0^1(1-u)^Ne^{iau}du\to\int_0^\infty e^{-y+i\sqrt sy}dy=(1-i\sqrt s)^{-1}\), and Stirling's formula gives

\[
W_{N,s}\ \sim\ \min(1,B-1)\,\sqrt{\frac{s}{2\pi(1+s)}}\,\frac{s^{N/2}}{\sqrt N}\qquad(N\to\infty).
\]

For \(B\ge2\) the orthonormal Hermite coefficients of this law are \(c_n=\lambda D^n/\sqrt{n!}\) for \(n\ge1\), and \(c_N^2=e^{-N}(sN)^N/N!\ge s^N/(e\sqrt N)\) (\cref{fig:energy}).
\end{proof}

Part (a) is a lower bound on the leak of any selection rule, with \(\sigma^2=1\) for raw and \(2\) for thinned selection. A cell that captures half of its own component also captures a fraction of order \(e^{-D^2/(2\sigma^2)}\) of an equally heavy component at distance \(D\), which exceeds the critical weight \(e^{-D^2/2}\) when \(\sigma^2=2\). Part (b) shows that such weights defeat truncation. At \(s=1\), uniform accuracy costs degree of order \(1/\varepsilon^2\), and for \(s>1\) the worst-case error grows exponentially in the degree.

\begin{example}[exact anchors]\label{ex:exact-anchors}
Take \(d=1\), equal weights, means \(0\) and \(D=\sqrt{8N}\), anchors at the two means, and the cell of \(0\) selected on \(X+g\). The cell has probability \(1/2\) and measure \(\nu=(1-\lambda)\delta_0+\lambda\delta_D\) with \(\lambda=\Phi(-D/(2\sqrt2))=\Phi(-\sqrt N)\), so \(\|r_\nu-1\|_{L^1(\gamma)}\le2\lambda\to0\). Testing against \(e^{i\sqrt Nx}\) as in the proof of part (b), now with \(a=\sqrt8\,N\), and using Mills' ratio for \(\lambda\) and \(|1-i\sqrt8|=3\),

\[
\big\|r_\nu-S_Nr_\nu\big\|_{L^1(\gamma)}\ \ge\ \lambda e^{-N/2}\,\frac{a^{N+1}}{N!}\,\Big|\int_0^1(1-u)^Ne^{iau}\,du\Big|=\Big(\frac{\sqrt8}{6\pi}+o(1)\Big)\frac{8^{N/2}}{N}.
\]

For \(N=4,8,16,32\) the leak \(\lambda\) is \(2.3\times10^{-2}\), \(2.3\times10^{-3}\), \(3.2\times10^{-5}\), \(7.7\times10^{-9}\) and the lower bound is \(2.03\), \(69.7\), \(1.49\times10^5\), \(1.28\times10^{12}\).

\end{example}

\begin{figure}
\centering
\includegraphics[width=0.67\textwidth]{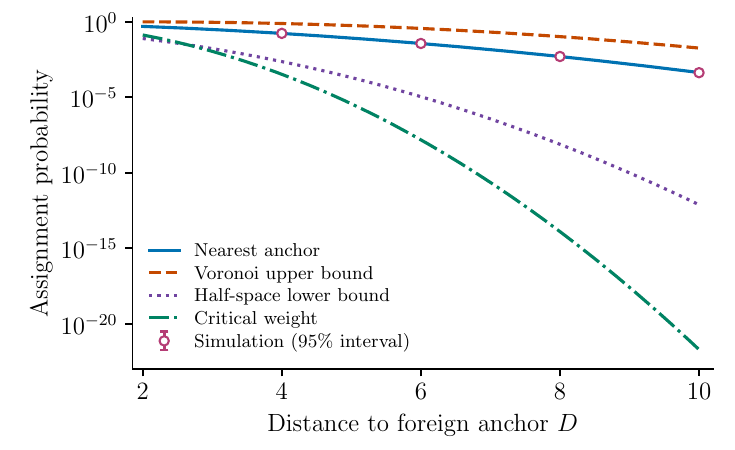}
\caption{Foreign assignment with thinned data (\(\sigma^2=2\)), cover radius \(R=2\), and \(d=30\). The lower bound uses \(p=1/2\) in \cref{thm:criticality}(a). The upper bound is from \cref{prop:cells}(3). The critical reference weight is \(e^{-D^2/2}\). Bars are exact pointwise binomial intervals.}\label{fig:leak}
\end{figure}

\section{Conformal resummation and optimal moment recovery}\label{sec:recovery}

In \cref{sec:taper} the taper degree is \(A\), \(\tau\) bounds normalized projections, \(s\) and \(z\) are generating variables and \(w=v'/(v'+1)\). From \cref{sec:exponent} on, \(N\) is the largest supplied moment degree, \(m=N+1\), \(v=s/2\) and \(q=\sqrt{s/(s+2)}\).

\subsection{A positive taper on the original Hermite levels}\label{sec:taper}

For an atom \(\delta\) and a point \(h\) write \(u=|\delta|^2\), \(t=\langle\delta,h\rangle\), and let \(y_n(\delta,h)\) be the degree-\(n\) Hermite level of the kernel of \(r_\nu\):

\begin{equation}\label{eq:generating}
e^{st-s^2u/2}=\sum_{n\ge0}s^n\,y_n(\delta,h),\qquad y_n(\delta,h)=\frac{\langle\delta^{\otimes n},\operatorname{He}_n(h)\rangle}{n!}.
\end{equation}

For a displacement measure put \(M_n=\int\delta^{\otimes n}d\nu\) and

\[
Y_n(h)=\int y_n(\delta,h)\,d\nu=\langle M_n,\operatorname{He}_n(h)\rangle/n! .
\]

Hermite truncation is \(\sum_{n\le N}Y_n\). We keep the quantities \(Y_n\) and change their weights.

\begin{definition}[conformal taper]\label{def:taper}
Fix \(v'\ge1\) and \(w=v'/(v'+1)\). For an integer \(A\ge0\) let \(\Lambda_0=1\) and

\[
\Lambda_n=(1-w)^{n/2}\sum_{i=0}^{\lfloor(A-n)/2\rfloor}\frac{\Gamma(n/2+i)}{\Gamma(n/2)\,i!}\,w^i\quad(n\ge1),\qquad q_\nu(h)=\sum_{n=0}^{A}\Lambda_n\,Y_n(h).
\]

\(\Lambda_n\) is the probability that a negative binomial variable with shape \(n/2\) and success probability \(1-w\) is at most \((A-n)/2\). In particular, \(0\le\Lambda_n\le1\) and \(\Lambda_A=(1-w)^{A/2}\) for \(A\ge1\).
\end{definition}

\begin{theorem}[conformal resummation]\label{thm:taper}
Let \(\nu=\sum_{l\le k}\rho_l\delta_{\delta_l}\) be a probability measure with \(\rho_l\le K\exp(-(|\delta_l|-R)_+^2/(2v'))\), \(K\ge1\), \(R\ge0\), and \(v'\ge1\). Let \(\tau>0\) and let \(h\) satisfy \(|\langle\delta_l,h\rangle|\le\tau|\delta_l|\) for all \(l\). For the degree-\(A\) taper of \cref{def:taper},

\[
\big|r_\nu(h)-q_\nu(h)\big|\ \le\ 4(v'+1)\,kK\,\exp\!\Big(\frac{(R+2v'\tau)^2}{v'}-\frac{A+1}{4(v'+1)}\Big).
\]

\end{theorem}

\begin{lemma}[conformal expansion]\label{lem:conformal}
Let \(s(z)=z/\sqrt{v'(1-z^2)}\) and \(z_0=\sqrt w\). Then \(s(z_0)=1\), and for \(|z|<1\)

\[
e^{s(z)t-s(z)^2u/2}=\sum_{m\ge0}z^m\,\Psi_m(u,t),
\]

where (i) \(\Psi_m=\sum_{n\le m}\kappa_{m,n}y_n\) with \(\kappa_{m,n}\ge0\) and \(\sum_{m\le A}z_0^m\kappa_{m,n}=\Lambda_n\). Moreover, (ii) for every \(r\in(0,1)\),

\[
|\Psi_m(u,t)|\le r^{-m}\exp\!\Big(\frac{r^2u}{2v'(1+r^2)}+\frac{r|t|}{\sqrt{v'(1-r^2)}}\Big).
\]

\end{lemma}

\begin{proof} \(s(z_0)^2=w/(v'(1-w))=1\) because \(w/(1-w)=v'\). The function \(s\) is analytic in the unit disc with \(s(0)=0\), and for \(n\ge1\)

\[
s(z)^n=v'^{-n/2}z^n(1-z^2)^{-n/2}=v'^{-n/2}\sum_{i\ge0}\frac{\Gamma(n/2+i)}{\Gamma(n/2)\,i!}\,z^{n+2i},
\]

a series with nonnegative coefficients. Substituting \(s=s(z)\) in \eqref{eq:generating} and collecting powers of \(z\) (the double series converges absolutely, being dominated by \(\sum_ns(|z|)^n|y_n|<\infty\)) gives (i) with \(\kappa_{n+2i,n}=v'^{-n/2}\Gamma(n/2+i)/(\Gamma(n/2)i!)\), \(\kappa_{0,0}=1\), and \(\sum_{m\le A}z_0^m\kappa_{m,n}=\sum_{i\le(A-n)/2}w^{n/2+i}v'^{-n/2}\frac{\Gamma(n/2+i)}{\Gamma(n/2)i!}=\Lambda_n\) since \(w/v'=1-w\). For (ii), by Cauchy's estimate \(|\Psi_m|\le r^{-m}\max_{|z|=r}\exp(\mathrm{Re}(s(z))\,t-\mathrm{Re}(s(z)^2)\,u/2)\). On \(|z|=r\), \(|s(z)|\le r/\sqrt{v'(1-r^2)}\). Moreover \(s(z)^2=\frac1{v'}\big(\frac1{1-z^2}-1\big)\) and \(\mathrm{Re}\frac1{1-\zeta}\ge\frac1{1+|\zeta|}\), so \(-\mathrm{Re}(s(z)^2)\le\frac1{v'}\big(1-\frac1{1+r^2}\big)=\frac{r^2}{v'(1+r^2)}\). 
\end{proof}

\begin{proof}[Proof of \cref{thm:taper}] We evaluate the conformal expansion at the point \(z_0\) where it reproduces the kernel, and bound its tail by the Cauchy estimate (ii). At \(z=z_0\) the expansion and (i) read

\[
e^{t-u/2}=\sum_{m\ge0}z_0^m\Psi_m,\qquad \sum_{m\le A}z_0^m\Psi_m=\sum_{n\le A}\Lambda_ny_n .
\]

Hence for one atom, with \(x=|\delta_l|\) and \(|t|\le\tau x\), and any \(r\in(z_0,1)\),

\[
\Big|e^{t-u/2}-\sum_{n\le A}\Lambda_ny_n\Big|\le\sum_{m>A}\Big(\frac{z_0}{r}\Big)^m e^{\alpha x^2+\beta x},\qquad\alpha=\frac{r^2}{2v'(1+r^2)},\ \ \beta=\frac{r\tau}{\sqrt{v'(1-r^2)}} .
\]

Take \(r=w^{1/4}\). Since \(1-(1+y)^{-1/2}\ge\frac{y}{2(1+y)}\) and \(1-(1+y)^{-1/4}\ge\frac{y}{4(1+y)}\) for \(y\ge0\) (integrate the corresponding derivative bounds from \(0\)), with \(y=1/v'\) we get \(1-r^2\ge\frac1{2(v'+1)}\) and \(1-z_0/r=1-w^{1/4}\ge\frac1{4(v'+1)}\). Therefore \(\beta\le\tau\sqrt{2(v'+1)/v'}\le2\tau\) and \(\sum_{m>A}(z_0/r)^m\le4(v'+1)w^{(A+1)/4}\le4(v'+1)e^{-(A+1)/(4(v'+1))}\), using \(\log(1/w)\ge1/(v'+1)\).

Multiply by \(\rho_l\le Ke^{-(x-R)_+^2/(2v')}\). For \(x\ge R\) the exponent \(f(x)=-\frac{(x-R)^2}{2v'}+\alpha x^2+\beta x\) is a concave quadratic with leading coefficient \(-\frac1{2v'(1+r^2)}\), and its maximum is

\[
\frac{(R+v'\beta)^2(1+r^2)-R^2}{2v'}\le\frac{(R+2v'\tau)^2}{v'} .
\]

For \(x\le R\), \(\alpha x^2+\beta x\le f(R)\). Summing over at most \(k\) atoms gives the bound. 
\end{proof}

The taper is a linear summability method: it keeps the Hermite levels \(Y_n\), which are the quantities a learner can estimate, and replaces the sharp cutoff by weights that decrease across a width of order \(\sqrt A\). The weights are the partial sums of the power series of the kernel after a conformal change of variable, in the manner of conformal summation \cite{Har49} and conformal quadrature \cite{HT08}. The Gaussian decay of the leak controls a distant atom on the new contour. Hermite approximation of functions analytic in a strip \cite{WZ26} uses analyticity in the spatial argument, whereas the taper uses analyticity in the scale of the displacement, which is what a tail budget on the mixing law provides.

\begin{figure}
\centering
\includegraphics{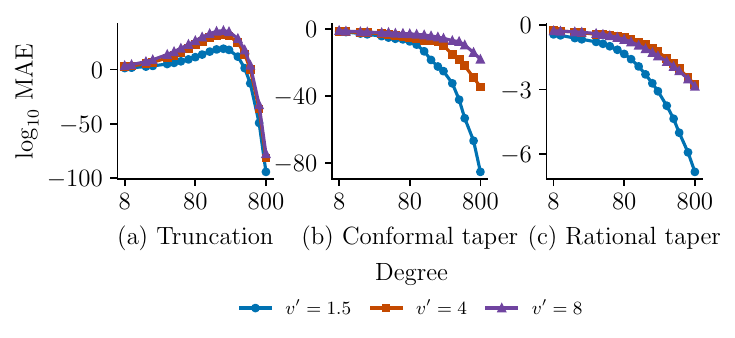}
\caption{Density ratio error for ordinary truncation (a), the original conformal taper from \cref{def:taper} (b), and the rational taper in \eqref{eq:rational} (c). At each leak variance the three reconstructions use identical moments of the same finite law.}\label{fig:tapers}
\end{figure}

A change in the conformal map lets the quadratic term absorb the linear term on the contour and improves the bound for large \(v'\).

\begin{proposition}[rational taper]\label{prop:rational}
Under the hypotheses of \cref{thm:taper}, let \(p\) be the least odd positive integer with \(2^p\ge 8v'+1\), set \(w_p=1-2^{-p}\) and \(a=1-1/p\). Use \(\Lambda^{(p)}_0=1\) and

\[
\Lambda^{(p)}_n=2^{-(p-1)n/2}
 \sum_{j=0}^{\lfloor(A-n)/2\rfloor}
 \frac{(an/2)_j}{j!}\,w_p^j,\qquad 1\le n\le A ,
\]

where \((x)_j\) is the rising factorial. These rational weights lie in \([0,1]\), since the infinite binomial sum cancels the leading factor. Their weighted sum \(q_{\nu,p}=\sum_{n=0}^A\Lambda_n^{(p)}Y_n\) satisfies

\begin{equation}\label{eq:rational}
\left|r_\nu(h)-q_{\nu,p}(h)\right|
\le 2^{p+1}kK\exp\!\left(\frac{R^2}{2v'}+
             \frac{p\tau^2}{2}-\frac{A+1}{2^{p+1}}\right).
\end{equation}
\end{proposition}

\begin{proof} Set \(z_0=\sqrt{w_p}\) and \(S(z)=(z/z_0)((1-w_p)/(1-z^2))^{a/2}\), using the branch positive at zero for the fractional power. Then \(S(z_0)=1\). For \(\zeta=z^2\) in the open unit disc and \(c=2/(2^p-1)\),

\[
S(z)^2=c\big((1-\zeta)^{-a}-(1-\zeta)^{1-a}\big).
\]

Since \(|\arg(1-\zeta)|<\pi/2\), the real part of the first summand inside parentheses is at least \(\sin(\pi/(2p))|1-\zeta|^{-a}\ge |\zeta(1-\zeta)^{-a}|/p\). The absolute value of the second is at most two. Thus

\[
\operatorname{Re}(S(z)^2)+1/(2v')\ge |S(z)|^2/p .
\]

For \(x=|\delta_l|\), the profile gives \(\rho_l\le K\exp(R^2/(2v')-x^2/(4v'))\). Completing the square bounds the modulus of the weighted kernel at \(S(z)\) by \(K\exp(R^2/(2v')+p\tau^2/2)\) on the whole disc. Cauchy's estimate, with contour radius tending to one, bounds each Taylor coefficient by the same quantity. At \(z_0\) the retained coefficients sum to the stated weighted Hermite levels by the binomial series. The tail is at most the coefficient bound times \(z_0^{A+1}/(1-z_0)\). Now \(1-z_0\ge2^{-p-1}\) and \(z_0\le e^{-2^{-p-1}}\). Sum over the atoms to obtain \eqref{eq:rational}. 
\end{proof}

\subsection{Exact recovery exponent from global moments}\label{sec:exponent}

Let \(\mathcal I_N(\nu)\) contain all ordinary moments through total degree \(N\). The minimax recovery error over probability-valued reconstructions \(H\) is

\[
E_{N,d}(s,B)=\inf_H\sup_{\nu\in\mathcal C_{s,d}(B)}
d_{\mathrm{TV}}\big(H(\mathcal I_N(\nu)),\nu*\gamma\big).
\]

The input consists of exact moments, so \(E_{N,d}\) is a pure approximation error: it is the error of the best estimator in the limit of infinitely many samples when only moments through degree \(N\) are used. \Cref{cor:responses} adds the estimation error. Finite-atomic approximation grants access to the whole mixing law and restricts the output component count \cite{MWY25}.

\begin{theorem}[sharp exponential rate for moment recovery]\label{thm:recovery}
Fix \(s>0\), \(B>1\), set \(m=N+1\) and \(q=\sqrt{s/(s+2)}\). For every \(d\ge1\) and \(N\ge1\),

\[
\frac{c\min(1,B-1)}{\sqrt{m+1}}q^m
\ \le\ E_{N,d}(s,B)\ \le\
 Bq^m\inf_{q^2\le b<1} b^{-m/2}(1-b^2)^{-d/4},
\]

where \(c>0\) is universal. Both bounds apply if the class is restricted to finitely supported laws with an unrestricted number of atoms. In particular, if \(d=d_N=o(N)\), including every fixed dimension, then

\[
\lim_{N\to\infty} E_{N,d_N}(s,B)^{1/(N+1)}
=\sqrt{\frac{s}{s+2}}.
\]

For fixed \(d\), the least sufficient degree for accuracy \(\varepsilon\) is asymptotic to \(2\log(1/\varepsilon)/\log(1+2/s)\) as \(\varepsilon\downarrow0\).

\end{theorem}

\begin{proof} We first prove the upper bound. Write \(v=s/2\) and let \(h_\alpha\) denote the product of the one-dimensional normalized Hermite polynomials indexed by \(\alpha\). With \(y=\delta/\sqrt v\) and \(z=x/\sqrt{1+v}\), the Gaussian channel has the expansion

\[
\frac{g_1(x-\delta)}{g_{1+v}(x)}
=\sum_{\alpha\in\mathbb N_0^d}q^{|\alpha|}h_\alpha(y)h_\alpha(z).
\]

Indeed, under \(x\sim N(\delta,I_d)\), \(z\) is Gaussian with mean \(qy\) and covariance \((1-q^2)I_d\). Taking the expectation of the Hermite generating function at \(z\) gives that at \(y\) with its variable multiplied by \(q\). These are the coefficients of the displayed ratio in the orthonormal basis for \(g_{1+v}\). The ratio belongs to \(L^2(g_{1+v})\) for each \(\delta\), since \(1+v>1\). The same expansion for a Gaussian channel with correlation \(\sqrt b\), followed by Parseval and completion of the square in each coordinate, gives the diagonal Mehler identity

\[
\sum_\alpha b^{|\alpha|}h_\alpha(y)^2
=(1-b^2)^{-d/2}\exp\!\left(\frac{b|y|^2}{1+b}\right),\qquad 0<b<1 .
\]

In particular the squared \(L^2(g_{1+v})\) tail beyond degree \(N\) at \(y\) is bounded by

\[
(q^2/b)^m(1-b^2)^{-d/2}
\exp\!\left(\frac{b|y|^2}{1+b}\right),\qquad q^2\le b<1 .
\]

Its square root bounds the \(L^1\) error in the corresponding density. Since \(b/(1+b)\le1/2\), the exponential in the square root is at most \(e^{|\delta|^2/(4v)}=e^{|\delta|^2/(2s)}\). Integration over \(\nu\) therefore costs at most \(B\). Each retained coefficient \(q^{|\alpha|}\int h_\alpha(\delta/\sqrt v)\,d\nu\) uses only \(\mathcal I_N(\nu)\). If \(p\) denotes the resulting signed density, it integrates to one. The positive part \(p_+\) has mass at least one. For any probability density \(f\), \(\|p_+-f\|_1\le\|p-f\|_1\) and \(|\int p_+-1|\le\|p_+-f\|_1\). Thus \(p_+/\int p_+\) achieves the asserted total variation error. This normalization requires no additional moment.

We turn to the lower bound. Embed a one-dimensional latent law on the first coordinate axis. Its convolved law is the product of its one-dimensional convolution and a standard Gaussian on the other coordinates. This preserves total variation, so it suffices to construct the pair on the line. Use the signed measure \(d\sigma(x)=g_v^{(m)}(x)\,dx\). Integration by parts annihilates every polynomial of degree below \(m\). Its weighted variation \(W=\int e^{x^2/(2s)}\,d|\sigma|(x)\) satisfies

\[
W=\frac{v^{-m/2}}{\sqrt{2\pi}}
\int_{\mathbb R}|\operatorname{He}_m(t)|e^{-t^2/4}\,dt
\le C v^{-m/2}\sqrt{m!}\,(m+1)^{1/4}.
\]

For the inequality, apply Cauchy--Schwarz after multiplying and dividing by \((1+t^2/(m+1))^{1/2}\). The integral of \((1+t^2/(m+1))^{-1}\) is \(\pi\sqrt{m+1}\). The other squared factor integrates against \(e^{-t^2/2}\). Hermite orthogonality and \(t\operatorname{He}_m=\operatorname{He}_{m+1}+m\operatorname{He}_{m-1}\) give \(\int t^2 \operatorname{He}_m^2\,d\gamma=(2m+1)m!\). These two estimates give the stated universal constant.

Write \(\sigma=\sigma_+-\sigma_-\) and \(u=\sigma_+(\mathbb R)=\sigma_-(\mathbb R)\). With \(a=\min(1,B-1)/W\), the laws \(\nu_\pm=(1-au)\delta_0+a\sigma_\pm\) are probabilities with the same data \(\mathcal I_N\). Indeed \(au\le1/2\) and their exponential moments are at most \(1+aW\le B\). Their convolved difference is \(a g_{1+v}^{(m)}\). Its characteristic function at \(t=\sqrt{m/(1+v)}\) has modulus

\[
a |t|^m e^{-(1+v)t^2/2}
=a(1+v)^{-m/2}(m/e)^{m/2}.
\]

This is at most twice their distance in total variation. Any single reconstruction incurs at least half that distance on one member of the pair. By Stirling's inequality \((m/e)^{m/2}\ge c_0\sqrt{m!}(m+1)^{-1/4}\). Substituting the estimate on \(W\) proves the lower bound.

The same argument holds with finite support. Replace each law \(\nu_\pm\) by a positive quadrature preserving the expectations of \(x,\ldots,x^N\), \(e^{x^2/(2s)}\), \(\cos(tx)\) and \(\sin(tx)\) at the test frequency above. The expectation of an integrable vector in \(\mathbb R^j\) lies in the convex hull of its range, so Carathéodory's theorem gives such a quadrature with at most \(j+1\) points. Here \(j=N+3\), so at most \(N+4\) atoms per law preserve both the budget and the distinguishing Fourier value.

Finally, if \(d=o(m)\), take \(b^2=m/(m+d)\), which is at least \(q^4\) for all sufficiently large \(m\). The logarithm of the upper prefactor apart from \(B\) becomes

\[
\tfrac14\{m\log(1+d/m)+d\log(1+m/d)\}=o(m).
\]

The lower prefactor has logarithm \(O_B(\log m)\). Taking \(m\)th roots proves the limit. For fixed \(d\), monotonicity of the minimax error and this limit give the degree asymptotic. 
\end{proof}

The upper bound uses the full tail budget at \(v=s/2\). Convolution takes the moment-annihilating derivative at exactly that variance to a derivative of variance \(1+v\), with the same exponential loss. The upper bound is the \(L^1\) error \(\|p-f\|_1\) of the signed reconstruction, which dominates the total variation error after positive-part normalization. The quantity \(U_m\) in \cref{ex:mehler} uses the same convention.

For perturbed coefficients \(\widehat a_\alpha\) with \(\widehat a_0=1\), orthogonality under \(g_{1+v}\) and Cauchy--Schwarz bound the additional \(L^1\) error by

\[
\left(\sum_{1\le|\alpha|\le N}q^{2|\alpha|}
|\widehat a_\alpha-a_\alpha|^2\right)^{1/2},
\qquad a_\alpha=\int h_\alpha(\delta/\sqrt v)\,d\nu.
\]

The same penalty applies to total variation after normalization.

For \(N\ge2\), the moments determine the covariance rank and mean. In the orthogonal complement of the covariance range the latent projection is constant, so the smoothed factor is a known independent Gaussian. Projecting onto the range reduces displacement norms and preserves the budget. The upper bound therefore holds with covariance rank in place of \(d\), with exact recovery at rank zero. The full Mehler array has \(\binom{N+d}{d}\) coefficients. \Cref{sec:trains} avoids forming it for finite mixtures.

\begin{corollary}[identification of atomic laws]\label{cor:atomic-identification}
Moments through degree \(2k-1\) identify every probability law supported on at most \(k\) points.
\end{corollary}

\begin{proof} Project the difference of two such laws in a direction separating its at most \(2k\) support points. The projected moment equations give a full-column-rank Vandermonde system in the signed masses.\end{proof}

\begin{figure}
\centering
\includegraphics{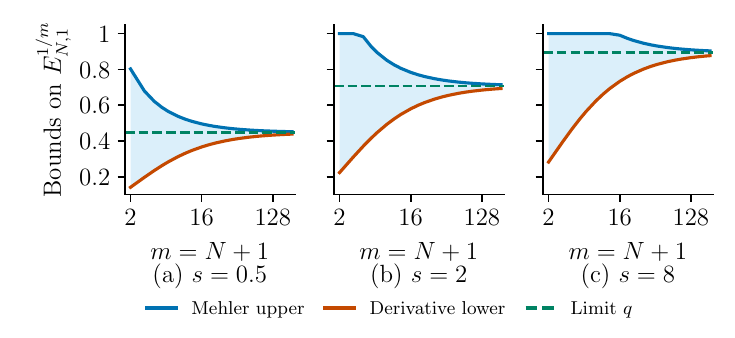}
\caption{Bounds on the \(m\)th root of the worst-case total variation error in one dimension for \(B=2\). Shading spans \(L_m^{1/m}\) to \(U_m^{1/m}\). Dashed lines give the limiting bases from \cref{thm:recovery}.}\label{fig:mehler}
\end{figure}

\begin{figure}
\centering
\includegraphics[width=0.67\textwidth]{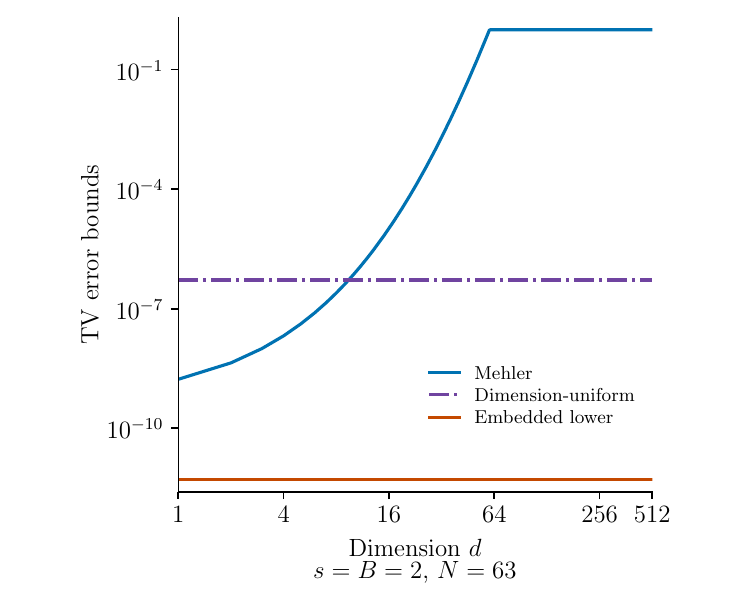}
\caption{Worst-case total variation error bounds as dimension \(d\) varies, with \(s=B=2\) and \(N=63\). The lower bound embeds the one-dimensional problem in each dimension. The upper bounds come from \cref{thm:recovery,prop:uniform}.}\label{fig:mehler-dimension}
\end{figure}

\begin{example}[the exponent at finite degree]\label{ex:mehler}
Take \(d=1\), \(B=2\) and \(m=N+1\). Half the TV distance of the two smoothed probability laws in the lower-bound proof gives \(L_m\). The Mehler reconstruction gives the upper bound \(U_m\). With \(b_m=\max(q^2,\sqrt{m/(m+1)})\), these quantities are

\[
\begin{aligned}
L_m&=\frac{q^m}{4}
\frac{\int_{\mathbb R}|\operatorname{He}_m(t)|e^{-t^2/2}\,dt}
     {\int_{\mathbb R}|\operatorname{He}_m(t)|e^{-t^2/4}\,dt},\\
U_m&=\min\{1,\,2q^mb_m^{-m/2}(1-b_m^2)^{-1/4}\}.
\end{aligned}
\]

Integrate over the intervals cut out by the roots of \(\operatorname{He}_m\). At \(s=2\), \(m=64\) this gives \(0.6661\le E_{63,1}(2,2)^{1/64}\le0.7294\), around the limiting base \(1/\sqrt2\). \Cref{fig:mehler} shows how the one-dimensional bounds depend on \(m\) and \(s\). \Cref{fig:mehler-dimension} compares the bounds across dimensions at degree \(63\).
\end{example}

\begin{corollary}[recovery from a noisier response]\label{cor:responses}
Let \(Y_1,\ldots,Y_n\) be independent draws from \(\nu*g_t\) in \(d\) dimensions, with \(\nu\in\mathcal C_{s,d}(B)\) and known \(t\ge1\). The target density is \(f=\nu*g_1\). Put \(r=\sqrt{s/(s+1)}\), \(a_t=\sqrt{(s+t)/(s+1)}\), \(M_t=B(1+s/t)^{d/2}\) and \(D_A=\binom{A+d}{d}\). There is a density estimate with

\[
\begin{aligned}
\mathbb{E}[d_{\mathrm{TV}}(\hat f_A,f)] \le a_t^A\sqrt{M_tD_A/n}
+B r^{A+1}\inf_{r^2\le b<1}b^{-(A+1)/2}(1-b^2)^{-d/4}.
\end{aligned}
\]

For fixed \(s,B,d,t\), take \(A=\lfloor\log n/\log(1+t/s)\rfloor\).

The risk is \(O_{s,B,d,t}((1+\log n)^{d/2}n^{-\gamma})\), where \(\gamma=\allowbreak\log(1+1/s)\allowbreak/(2\log(1+t/s))\). In a known linear signal space its dimension replaces \(d\) throughout.

\end{corollary}

\begin{proof} Estimate the coefficient of \(h_\alpha(x/\sqrt{s+1})\) in \(f/g_{s+1}\) by averaging \(a_t^{|\alpha|}h_\alpha(Y_i/\sqrt{s+t})\). Given \(U=u\), its expectation is \(r^{|\alpha|}h_\alpha(u/\sqrt s)\) by the Hermite generating function. Thus all retained coefficients are unbiased. Maximization over \(y\) gives

\[
\sup_y\frac{g_t(y-u)}{g_{s+t}(y)}=(1+s/t)^{d/2}e^{|u|^2/(2s)}.
\]

Gaussian orthonormality now bounds the sum of coordinate variances by \(a_t^{2A}M_tD_A/n\). Cauchy--Schwarz bounds the expected integrated absolute estimation error by its square root. The Mehler tail estimate of \cref{thm:recovery} with latent reference variance \(s\) gives the second term. Since the constant coefficient is one, taking the positive part and normalizing gives the TV bound as in that theorem. For the rate use \(b^2=(A+1)/(A+1+d)\) for all sufficiently large \(A\). The orthogonal complement of a known signal space contributes an independent standard Gaussian to the target. 
\end{proof}

The two terms are the estimation and approximation errors, and the stated \(A\) balances them. The first grows like \(a_t^A\) because an unbiased estimate of a degree-\(A\) coefficient from noisy responses has variance exponential in \(A\). For the conditional responses used in \cref{sec:trains}, \(t=2\) and \(\gamma>1/4\) for every finite \(s>0\). The estimator forms \(D_A\) coefficients, which is polynomial in \(d\) only for fixed \(A\). \Cref{sec:trains} replaces the dense array by a low-rank one.

\subsection{Geometric recovery in arbitrary dimension}\label{sec:uniform}

When \(d\) grows with \(N\), the number of Hermite coefficients can overwhelm the Mehler tail bound. A different expansion controls the integrated error of a translated kernel directly. The linear term in its exponent moves the location. An additional quadratic term keeps the growth caused by complex displacements within the latent tail budget.

\begin{proposition}[dimension-uniform moment recovery]\label{prop:uniform}
Let \(s>0\), \(B>1\), \(v=s/2\), and set

\[
R^2=\frac{\sqrt{(2+v)^2+4}-2}{v}>1 .
\]

For every \(d\ge1\) and \(N\ge1\), a single polynomial reconstruction from \(\mathcal I_N(\nu)\), followed by taking its positive part and normalizing, gives

\[
E_{N,d}(s,B)\le B R^{-(N+1)} .
\]

Thus \(O_s(\log(B/\varepsilon))\) moment degrees suffice in every dimension. The array of supplied moments can still be large.

\end{proposition}

\begin{proof} Write \(D(z)=2+v-vz^2\), \(a(z)=2z/D(z)\) and \(b(z)=a(z)^2/2+z^2/D(z)\). For \(|z|<R\) consider

\[
H(z,x)=\gamma(x)\int \exp\!\left(a(z)\langle x,u\rangle-\tfrac12 b(z)|u|^2\right)d\nu(u).
\]

Since \(a(1)=b(1)=1\), \(H(1,\cdot)\) is the target density. Integrating the modulus of the kernel in \(x\) gives \(\exp(((\operatorname{Re}a)^2-\operatorname{Re}b)|u|^2/2)\). The choice of \(b\) gives the identity

\[
1-s\big((\operatorname{Re}a)^2-\operatorname{Re}b\big)=
\frac{(2+v)^2-4v|z|^2-v^2|z|^4}{|D(z)|^2}>0 .
\]

The numerator first vanishes at \(|z|=R\), inside the nearest pole. Consequently the integral of \(|H(z,x)|\) is at most \(B\), independently of \(d\). After testing against any bounded function, integration along compact contours and Morera's theorem justify holomorphy. Cauchy's formula then identifies each coefficient \(h_j(x)=[z^j]H(z,x)\) by integration over \(\nu\). Since \(a\) starts at degree one and \(b\) at degree two, \(h_j\) uses moments only through degree \(j\) and is a polynomial in \(x\) times \(\gamma(x)\).

Put \(m=N+1\) and \(r=1/R\). Form the signed density

\[
p_N=\sum_{j=0}^{m-1}(1-r^{2(m-j)})h_j .
\]

For a real test \(|T|\le1\), write \(L(w)=\int T(x)H(Rw,x)\,dx=\sum_j c_jw^j\). The Poisson kernel \(P_r(\theta)=(1-r^2)/|1-re^{i\theta}|^2\) has Fourier coefficients \(r^{|j|}\). Therefore, with the boundary integral interpreted as a radial limit,

\[
L(r)-\sum_{j<m}(r^j-r^{2m-j})c_j
=r^m\lim_{h\uparrow1}\int_{-\pi}^{\pi}
 P_r(\theta)e^{-im\theta}L(he^{i\theta})\,\frac{d\theta}{2\pi}.
\]

The identity follows by integrating each coefficient at \(h<1\) and taking the limit. The right side has modulus at most \(Br^m\). Taking the supremum over tests gives \(\|H(1,\cdot)-p_N\|_1\le Br^m\). If \(p_N\) has positive-part mass at least one, normalization decreases its positive excess over the target. If that mass lies in \((0,1)\), normalization decreases its deficit. Either comparison bounds total variation by the signed \(L^1\) error. A zero positive part forces that error to be at least one, so any probability density suffices. 
\end{proof}

At \(s=B=2\) and \(N=63\) the bound is below \(5.27\times10^{-7}\) regardless of \(d\). The optimal base for \(d=o(N)\) is \(1/\sqrt2\), while the dimension-uniform construction has base \(1/R\simeq0.789\). Determining the sharp base for arbitrary dimension growth remains open.

\subsection{The one-dimensional prefactor and proportional dimension}\label{sec:line}

\begin{theorem}[sharp one-dimensional recovery]\label{thm:line}
For \(s>0\), \(B>1\), \(m=N+1\ge2\) and \(q=\sqrt{s/(s+2)}\),

\[
E_{N,1}(s,B)\asymp_{s,B}m^{-1/2}q^m.
\]

Consequently the least sufficient number \(m\) is \([\log(1/\varepsilon)-\tfrac12\log\log(1/\varepsilon)]/\log(1/q)+O_{s,B}(1)\). A linear map from the first \(N\) moments to a signed density, followed by positive-part normalization, attains the upper bound.

\end{theorem}

\begin{proof} The lower bound is \cref{thm:recovery}. For the upper bound we extrapolate the missing coefficients along the Hermite recurrence, which requires a uniform bound on the derivatives \(\psi_n'\) introduced next. Write \(h_n=\operatorname{He}_n/\sqrt{n!}\), \(\psi_n(y)=e^{-y^2/4}h_n(y)\), and \(v=s/2\). The Hermite recurrence gives

\[
\psi_n-\sqrt{(n-1)/n}\,\psi_{n-2}=-2\psi'_{n-1}/\sqrt n.
\]

We need \(\|\psi'_n\|_\infty\le C(n+1)^{1/4}\). Set \(\lambda=n+1/2\). The Hermite equation gives \(\psi_n''+(\lambda-y^2/4)\psi_n=0\). The energy \(J=(\psi_n')^2+(\lambda-y^2/4)\psi_n^2\) has derivative \(-y\psi_n^2/2\). The values \(h_{2j}(0)^2=4^{-j}\binom{2j}j\) and \(h'_{2j+1}(0)=\sqrt{2j+1}h_{2j}(0)\) imply \(J(0)\le C\sqrt{n+1}\). This bounds \(|\psi_n'|\) up to the turning point \(2\sqrt\lambda\). Beyond that point the decaying solution has constant sign, since integrating the differential equation against \(\psi_n\) from any zero to infinity would force \(\psi_n\) to vanish identically. Thus \(|\psi_n'|\) decreases there. Parity handles \(y<0\). Also \(\|h_n\|_{L^1(\gamma)}\le C(n+1)^{-1/4}\) by the derivative estimate for the Gaussian density in \cref{thm:criticality}(b).

Let \(a_n=\int h_n(u/\sqrt v)\,d\nu(u)\). The broadened Mehler expansion has coefficients \(q^n a_n\). For each \(n=m+j\), choose \(r\in\{m-2,m-1\}\) of the same parity as \(n\), and replace \(a_n\) by \(c_{n,r}a_r\), where \(c_{n,r}\) is the product of the recurrence multipliers from \(n\) down to \(r+2\). Each multiplier is at most one. Telescoping the recurrence gives

\[
\sup_y e^{-y^2/4}|h_n(y)-c_{n,r}h_r(y)|\le C(j+1)m^{-1/4}.
\]

The tail assumption bounds the integrated coefficient error by \(CB(j+1)m^{-1/4}\). The output basis function contributes another \(Cm^{-1/4}\) in \(L^1\). Summing \(\sum_{j\ge0}(j+1)q^{m+j}\) proves the upper bound. One may stop the replacement at \(n=2m-1\) and discard the subsequent terms. Indeed the ordinary tail from \(2m\) is \(O_{s,B}(m^{1/4}q^{2m})\) by \cref{thm:recovery}'s Mehler estimate with \(b=1-1/(2m)\) for all sufficiently large \(m\). The finitely many remaining \(m\) are absorbed in the constant. Positive-part normalization completes the proof. 
\end{proof}

The parity correction approximates each missing latent coefficient by the last observed coefficient of its parity, using the Hermite recurrence at the weight allowed by the tail budget.

\begin{proposition}[proportional-dimension envelope]\label{prop:proportional}
Fix \(s>0\), \(B>1\), \(m=N+1\), and suppose \(d/m\to\rho>0\) as \(N\to\infty\). Put \(q=\sqrt{s/(s+2)}\), \(b_0=\max\{0,(s-1)/(s+1)\}\), and take \(R\) from \cref{prop:uniform}. Then

\[
q\le\liminf E_{N,d}^{1/m}\le\limsup E_{N,d}^{1/m}\le
\min\left\{R^{-1},\inf_{b_0<b<1}\sqrt{\frac{s}{1+(s+1)b}}(1-b^2)^{-\rho/4}\right\}.
\]

The unconstrained minimizer is the unique positive solution of \(\rho b(1+(s+1)b)=(s+1)(1-b^2)\), clipped below at \(b_0\). The upper bound tends to \(q\) as \(\rho\downarrow0\).

\end{proposition}

\begin{proof} Repeat the Mehler calculation with latent reference variance \(v_b=sb/(1+b)\) and correlation \(q_b=\sqrt{v_b/(1+v_b)}\). The condition \(q_b^2<b\) holds on the specified interval. In the square root of the diagonal Mehler identity the latent exponent is \(b/[2v_b(1+b)]=1/(2s)\). Thus the finite-degree bound is \(B(q_b/\sqrt b)^m(1-b^2)^{-d/4}\). Take \(m\)th roots and combine with the embedded one-dimensional lower bound and \cref{prop:uniform}. Logarithmic differentiation gives the stated optimizer. 
\end{proof}

\begin{proposition}[radial witnesses in proportional dimension]\label{prop:radial-witness}
Let \(0<v<s\) and \(d/(2j)\to\rho>0\) as \(j\to\infty\). The signed measure \(d\sigma(x)=\Delta^j g_v(x)\,dx\) annihilates polynomials of degree below \(2j\). Put \(\tau=v/s\) and let \(L_j\) be the ratio of \(\|\sigma*\gamma\|_1\) to \(\int e^{|x|^2/(2s)}d|\sigma|(x)\). Then

\[
\lim L_j^{1/(2j)}=\sqrt{\frac{v}{1+v}}\sqrt{\frac{C_\rho(0)}{C_\rho(\tau)}},\qquad C_\rho(t)=\sup_{z\ge0}\frac{\sqrt{t^2+z^2}}{((1-t)^2+z^2)^{(1+\rho)/2}}.
\]

The best base over fixed \(v\) is \(\sqrt{s\tau_* /(1+s\tau_*)}(1-2\tau_*)^{\rho/4}<\sqrt{s/(s+2)}\), where \(\tau_*=2/(2+\rho+\sqrt{(2+\rho)^2+4\rho s})\).
\end{proposition}

\begin{proof} Integration by parts gives moment cancellation. With \(a=d/2\), the Gaussian heat equation gives \(\Delta^jg_v=(-2/v)^j j! L_j^{a-1}(|x|^2/(2v))g_v\). Hence \(L_j=(v/(1+v))^j\|p_0\|_1/\|p_\tau\|_1\), where \(p_t(x)=e^{tx}L_j^{a-1}(x)x^{a-1}e^{-x}/\Gamma(a)\) on \(x>0\), and zero elsewhere. Integrating the Laguerre generating function gives

\[
F_t(z)=\int e^{izx}p_t(x)\,dx=\binom{j+a-1}{j}\frac{(-t-iz)^j}{(1-t-iz)^{a+j}}.
\]

For fixed \(t<1\), \(\|F_t\|_\infty\le\|p_t\|_1\le C(1+j)\|F_t\|_\infty\) for all sufficiently large \(j\). For the upper estimate, weighted Cauchy--Schwarz and Plancherel give \(\|p_t\|_1\le C(\|F_t\|_2+\|F'_t\|_2)\). Write \(D(z)=(t^2+z^2)^{1/2}/((1-t)^2+z^2)^{(1+a/j)/2}\). Its supremum exceeds a constant \(c>0\) by taking \(z=1\). For sufficiently large fixed \(K\) and \(|z|>K\), \(D(z)\le c(K/|z|)^{\rho/3}\), uniformly for large \(j\). This bounds the \(L^2\) norm by a constant times the supremum. Logarithmic differentiation gives \(|F'_t|\le Cj|F_t|\) except possibly near zero for \(t=0\). There \(D(z)\le|z|\) and \(D(z)^j/|z|\le c^{j-1}\) for \(0<|z|<c/2\), supplying the same derivative norm bound. Taking roots gives the displayed limit. The maximizing \(z^2\) is \([((1-t)^2-(1+\rho)t^2)/\rho]_+\), with transition at \(t=(1+\sqrt{1+\rho})^{-1}\). Before this transition \(C_\rho(t)/C_\rho(0)=(1-2t)^{-\rho/2}\), and afterwards \(C_\rho(t)=t/(1-t)^{1+\rho}\). The latter branch gives a decreasing witness base. Differentiation on the first branch yields \(1-2\tau_*=\rho\tau_*(1+s\tau_*)\), with \(\tau_*\) inside the branch. Since \(\tau_*<1/2\), its base is strictly below \(q\). 
\end{proof}

Of the reconstructions in this section, the learner uses only the rational taper \eqref{eq:rational} on the original Hermite levels. Its error is pointwise on a projection event for a finite cell law with a Gaussian leakage profile, and each level is a contraction of a low-rank tensor. The Mehler reconstruction is sharp for exact moments and forms \(\binom{N+d}{d}\) coefficients. The quadratic correction of \cref{prop:uniform} uses radial contractions, and \cref{prop:variance} quantifies their ambient noise cost even at degree two.

\section{The real-number model and its primitives}\label{sec:machine}

\subsection{The machine and parameter choices}\label{sec:primitives}

\begin{definition}[real-number machine]\label{def:machine}
A register holds a real number. An operation is an arithmetic operation, a square root of a nonnegative number, a comparison and branch, a copy, or an independent standard Gaussian or uniform draw. Division requires a nonzero denominator. A draw from the mixture costs \(d\) operations.
\end{definition}

This extends the real-number model of \cite{BCSS98} by square roots and the stated random primitives. Square roots suffice for norms, thinning, radial clipping and the half-integer powers in the taper. The following bounds implement the remaining parameter choices.

\begin{lemma}[one-sided approximants]\label{lem:approximants}
Let \(\lg y\) be the least nonnegative integer \(j\) with \(2^j\ge y\), for \(y>0\). For \(x\ge0\) set \(M=\max(2,\lceil6x\rceil)\) and \(\exp^+(x)=2\sum_{j=0}^M x^j/j!\). Then

\begin{equation}\label{eq:arithmetic-bounds}
\ln y\le\lg y,\qquad
\lg y\le1+\log_2 y\quad(y\ge1),\qquad
e^x\le\exp^+(x)\le2e^x.
\end{equation}

Repeated doubling costs \(O(1+\lg y)\) operations. Counting up to \(M\) and forming successive terms costs \(O(1+x)\).

\end{lemma}

\begin{proof} The logarithmic bounds follow from the definition. For \(j>M\), the estimate \(j!\ge(j/e)^j\) gives \(x^j/j!\le2^{-j}\). The sum of the omitted terms is at most \(2^{-M}\), while the retained constant term equals one. This proves both exponential bounds. 
\end{proof}

\subsection{Capped Jacobi diagonalization}\label{sec:jacobi}

The tensor estimator needs a subspace with small discarded energy. To construct it we use the classical maximal-pivot Jacobi algorithm \cite{GVL13}. Write \(\operatorname{off}(A)^2=\sum_{i\ne j}a_{ij}^2\) for a symmetric \(m\times m\) matrix.

\begin{lemma}[one rotation]\label{lem:rotation}
Let \(a_{pq}\ne0\) and set

\[
u=\frac{a_{qq}-a_{pp}}{2a_{pq}},\qquad t=\frac{\operatorname{sgn}_+(u)}{|u|+\sqrt{1+u^2}},
\]

with \(\operatorname{sgn}_+(0)=1\). Put \(c=(1+t^2)^{-1/2}\) and \(s=tc\). Let \(J\) act on coordinates \(p,q\) by the matrix with rows \((c,s)\) and \((-s,c)\) and as the identity elsewhere. Then \(B=J^\top A J\) satisfies

\[
b_{pq}=0,\qquad \operatorname{off}(B)^2=\operatorname{off}(A)^2-2a_{pq}^2.
\]

\end{lemma}

\begin{proof} The rotation is orthogonal. Its transformed pivot is \(c^2a_{pq}(1-t^2-2ut)=0\). The other diagonal entries stay fixed and the Frobenius norm of the pivot block is invariant, giving the identity. 
\end{proof}

\begin{lemma}[capped Jacobi]\label{lem:jacobi}
Start with \(Q=I\) and repeatedly choose a largest absolute off-diagonal entry of \(Q^\top A Q\), breaking ties by index. Apply \cref{lem:rotation} and accumulate the rotation in \(Q\). Stop when the off-diagonal norm is at most \(\xi>0\), or report failure if a specified cap \(T\) on rotations has been reached with a larger residual. The returned diagonal \(D\) satisfies

\[
\|A-Q\operatorname{diag}(D)Q^\top\|_F\le\xi.
\]

If \(\|A\|_F\le\Gamma\) and \(\Gamma>0\), then \(T=m(m-1)\lg(\Gamma/\xi)\) suffices. Every run uses \(O(m^2(1+T))\) operations.

\end{lemma}

\begin{proof} For \(m\ge2\) the largest pivot has square at least \(\operatorname{off}(A)^2/(m(m-1))\). Each rotation thus contracts squared off-diagonal norm by a factor at most \(1-2/(m(m-1))\). The claimed cap follows from \(1-x\le e^{-x}\) and \eqref{eq:arithmetic-bounds}. For \(m=1\) the initial residual vanishes. Orthogonal invariance identifies the final off-diagonal norm with the displayed error. Scanning the matrix costs \(O(m^2)\) per rotation and updating the two affected rows and columns costs \(O(m)\). 
\end{proof}

\begin{proposition}[threshold and capture]\label{prop:capture}
Suppose \(G\succeq0\) has rank at most \(\kappa\), \(\|G-\hat G\|_{\mathrm{op}}\le\theta/4\) and the output of \cref{lem:jacobi} on the symmetric matrix \(\hat G\) has residual at most \(\theta/4\). Let \(U\) consist of the columns of \(Q\) with diagonal value greater than \(\theta\). It has at most \(\kappa\) columns. For every \(\varrho>0\) and \(b\) with \(G\succeq\varrho bb^\top\),

\[
\|(I-UU^\top)b\|^2\le3\theta/(2\varrho).
\]

\end{proposition}

\begin{proof} The reconstructed diagonal matrix in the original basis is within \(\theta/2\) of \(G\) in operator norm. If the retained subspace had dimension greater than \(\kappa\), it would contain a unit vector in \(\ker G\), contradicting the threshold. Every unit vector in the discarded subspace has \(G\)-energy at most \(3\theta/2\). Apply this to the unit vector along the discarded part of \(b\) whenever that part is nonzero. 
\end{proof}

\section{The anchored reduction}\label{sec:reduction}

Fix \(0<\eta\le1/8\), an anchor list and a cell \(m\) with profile \((K_m,R,v')\). Set

\[
\begin{aligned}
D_1=R+\sqrt{2v'\lg(kK_m/\eta)}, \qquad
\tau=2D_1+\sqrt{2\lg(4k/\eta)},\qquad c_m=\exp^+(\tau D_1).
\end{aligned}
\]

Let \(p\) be the odd integer in \eqref{eq:rational}. Choose the least \(A\ge0\) satisfying

\[
A+1\ge 2^{p+1}\left[\frac{R^2}{2v'}+
 \frac{p\tau^2}{2}+\lg\frac{2^{p+1}kK_m}{\eta}\right].
\]

Use the rational taper in \eqref{eq:rational} with this degree and write \(q_m=q_{\nu_m,p}\). Degrees and caps may vary between cells. Write \([z]_c=\min(c,\max(0,z))\).

\begin{theorem}[anchored reduction]\label{thm:reduction}
Let \(\mathcal M\) be a set of kept cells of positive mass. Take nonnegative estimates \(\hat P(m)\) there and zero elsewhere. Suppose that

\[
\sum_{m\in\mathcal M}|\hat P(m)-P(m)|+\sum_{m\notin\mathcal M}P(m)\le\eta.
\]

For each kept cell let \(h_1,\ldots,h_L\) be independent standard Gaussian vectors and \(\bar h=L^{-1/2}\sum_j h_j\). Suppose \(\mathbb{E}[|\tilde r_m-q_m(\bar h)|]\le\eta\) for an evaluator \(\tilde r_m\) of these vectors. Propose \(m\) proportionally to \(\hat P(m)\), propose \(x=a_m+\bar h\), and accept with probability \([\tilde r_m]_{c_m}/c\), where \(c=\max_{m\in\mathcal M}c_m\). The accepted law has total variation distance at most \(8\eta\) from \(F\) and acceptance probability at least

\[
\frac{1-7\eta-6\eta^2}{(1+\eta)c}>0 .
\]

Both \(A\) and \(\log c_m\) are \(O_{v'}(R^2+\log(kK_m/\eta))\).

\end{theorem}

\begin{lemma}[one cell]\label{lem:one-cell}
Under the hypotheses of \cref{thm:reduction}, for each kept cell put \(r=r_{\nu_m}\). Then \(\mathbb{E}[|r(\bar h)-[\tilde r_m]_{c_m}|]\le6\eta\).

\end{lemma}

\begin{proof} Separate atoms with \(|\delta_l|\le D_1\) from the remaining ones, and write \(r=r_{\mathrm{near}}+r_{\mathrm{far}}\). Integrating each kernel gives one, hence \(\mathbb{E}[r_{\mathrm{far}}(\bar h)]\le kK_m e^{-(D_1-R)^2/(2v')}\le\eta\). On the event \(G\) that \(|\langle\delta_l,\bar h\rangle|\le\tau|\delta_l|\) for every \(l\), the near contribution lies in \([0,c_m]\) and \eqref{eq:rational} gives \(|r-q_m|\le\eta\). Projection onto \([0,c_m]\) is 1-Lipschitz. The contribution of \(G\) to \(\mathbb{E}[|r_{\mathrm{near}}-[\tilde r_m]_{c_m}|]\) is therefore at most \(3\eta\).

Under a near shifted Gaussian, each nonzero normalized projection in the definition of \(G\) has a mean of absolute value at most \(D_1\) and variance one. A union bound yields \(\mathbb{E}[r_{\mathrm{near}}(\bar h)1_{G^c}]\le2k e^{-(\tau-D_1)^2/2}\le\eta\). Under the centered Gaussian, \(c_m\Pr[G^c]\le4k e^{-\tau(\tau-2D_1)/2}\le\eta\). Adding the far mass and the two event contributions gives \(6\eta\). 
\end{proof}

\begin{proof}[Proof of \cref{thm:reduction}] The unnormalized accepted density is proportional to

\[
H=\sum_{m\in\mathcal M}\hat P(m)H_m,\qquad
H_m(x)=\gamma_{a_m}(x)\mathbb{E}\big[[\tilde r_m]_{c_m}\mid\bar h=x-a_m\big].
\]

Jensen's inequality and \cref{lem:one-cell} give \(\|H_m-F_m\|_1\le6\eta\) and \(\|H_m\|_1\le1+6\eta\). \Cref{prop:cells} now gives \(\|H-F\|_1\le(1+6\eta)\eta+6\eta\). The right side is less than one even at \(\eta=1/8\). Normalization costs at most \(|1-\|H\|_1|\) in \(L^1\), proving the total variation estimate. The acceptance probability is \(\|H\|_1/(c\sum_m\hat P(m))\). Finally \(D_1^2\) and \(\tau^2\) have the asserted order by \eqref{eq:arithmetic-bounds}. 
\end{proof}

Rejection sampling cannot assign mass faster than the proposal supplies it. The following bound quantifies the loss when the anchors miss a region of component means, and shows why the cover must reach every component that matters.

\begin{proposition}[rejection from fixed anchors]\label{prop:rejection}
Let \(f\) be the finite unit-covariance mixture of \cref{sec:model}. Fix a probability density \(q=\sum_mp_m\gamma_{a_m}\) with positive weights, \(C=\operatorname{conv}\{a_m\}\) and \(c\ge1\). Over measurable acceptance rules of rate at least \(1/c\), the smallest total variation loss for the accepted law is

\[
E_c(f,q)=\sup_S\{F(S)-cQ(S)\}=\int(f-cq)_+\ \ge\sum_{i:w_i>0}w_iG_{c/w_i}(\operatorname{dist}(\mu_i,C)),
\]

where \(G_b(0)=0\) and \(G_b(r)=\Phi(r/2-\log(b)/r)-b\Phi(-r/2-\log(b)/r)\). Exact sampling at positive acceptance is possible precisely when \(C\) contains every positive-weight mean. Mixing in a fallback with probability \(\beta\) can reduce the lower bound by at most \(\beta\).

\end{proposition}

\begin{proof} Every accepted density \(g\) satisfies \(g\le cq\). If \(f\le cq\), take \(g=f\). Otherwise fill the missing mass of \(\min(f,cq)\) proportionally to \((cq-f)_+\). The resulting density attains \(E_c\) by accepting with probability \(g(x)/(cq(x))\), as in Lemma 3.17(1) of \cite{DR14}. Pointwise \(\sum_i(w_i\gamma_{\mu_i}-cq)_+\le(f-cq)_+\). For \(r=\operatorname{dist}(\mu_i,C)>0\), project along the unit normal from the nearest point of \(C\) to \(\mu_i\). Relative to that point the component projects to \(N(r,1)\) and every proposal mean projects to a nonpositive number. Testing the half-line above \(r/2+\log(c/w_i)/r\) gives \(G_{c/w_i}(r)>0\). Conversely, \(\mu_i\in C\) implies \(e^{\langle\mu_i,x\rangle}\le\max_me^{\langle a_m,x\rangle}\). Cancelling the common Gaussian in \(f/q\) therefore bounds the ratio whenever all means belong to \(C\). 
\end{proof}

\section{Thinned tensor trains}\label{sec:trains}

The degree-\(n\) moment \(M_n=\int v^{\otimes n}d\nu(v)\) has \(d^n\) entries. Its summands span at most \(k\) directions. The nested subspaces developed for mixture estimation in \cite{LL22,DK24} allow the learner to access this structure recursively. Tensor trains give coordinates for nested tensor-product subspaces \cite{Ose11,BSU16}. A moment of a \(k\)-atomic law has canonical rank at most \(k\), and classes of such tensors have metric entropy polynomial in \(d\) \cite{ZK26}, so that estimating them from samples need not cost \(d^n\). Here each slot indexes one factor of a moment. The subspace at that degree is learned from independent noisy views of a common latent mean. In this section \(N\) and \(N'\) are batch sizes, \(\kappa\) is a rank cap, \(n\) is a tensor order and \(L\) is a number of auxiliary Gaussian vectors.

\subsection{Product estimates and subspace capture}\label{sec:products}

\begin{lemma}[independent parts]\label{lem:parts}
For independent \(Z^{(i)}\sim N(v,\sigma^2I)\) and any order-\(j\) tensor \(T\),

\[
\begin{aligned}
\mathbb{E}[\langle T,\otimes_i Z^{(i)}\rangle]&=\langle T,v^{\otimes j}\rangle,\\
\mathbb{E}[\langle T,\otimes_i Z^{(i)}\rangle^2]&\le\|T\|^2(|v|^2+\sigma^2)^j .
\end{aligned}
\]

Here both tensor products range over \(1\le i\le j\).

\end{lemma}

\begin{proof} Independence gives the first moment and makes the second-moment operator the tensor product of \(j\) copies of \(vv^\top+\sigma^2 I\). Its operator norm is \((|v|^2+\sigma^2)^j\).
\end{proof}

The lemma is the source of the estimation error. A product of \(j\) independent parts is an unbiased estimate of a degree-\(j\) moment, and its variance grows geometrically in \(j\). The sample cost of a degree-\(A\) fit is therefore exponential in \(A\), and the rank bound \(k\) keeps its dependence on \(d\) polynomial.

A tensor train consists of column-orthonormal \(U_j\in\mathbb R^{k_{j-1}d\times k_j}\), starting with \(k_0=1\). Its basis \(B^{(j)}\) spans \(\hat S_j\subseteq\hat S_{j-1}\otimes\mathbb R^d\). The \(\beta\)th basis vector is the combination of \(B_\alpha^{(j-1)}\otimes e_i\) with coefficients \((U_j)_{(\alpha,i),\beta}\). For inputs \(z_1,\ldots,z_j\), start at \(c_0=1\) and compute projected coordinates by

\[
c_j(z_1,\ldots,z_j)=U_j^\top\big(c_{j-1}(z_1,\ldots,z_{j-1})\otimes z_j\big).
\]

A draw selects \(v\) from \(\lambda=\sum_{a\le\kappa}\varrho_a\delta_{v_a}\) and provides \(2n\) conditionally independent \(N(v,\sigma^2I)\) parts. \Cref{alg:tt}, called TT below, trains one level from each fresh batch. In the variant of \cref{sec:cover} it first applies the clipping map \(C_H(x)=x\min(1,H/|x|)\), with \(C_H(0)=0\), separately to the two features. An empty retained subspace has empty coordinates at all succeeding levels, which can be handled without matrix operations.

\begin{algorithm}
\caption{\textsc{TT}: thinned tensor train}\label{alg:tt}
\begin{algorithmic}[1]
\Require Order \(n\), rank cap \(\kappa\), batch size \(N\), thresholds \(\theta_j\), norm bounds \(\Gamma_j\), optional clipping level \(H\)
\For{\(j=1,\ldots,n\)}
  \State Take a fresh batch of \(N\) draws, each with parts \(Z^{(1)},\ldots,Z^{(2j)}\).
  \State \(p\gets c_{j-1}(Z^{(1)},\ldots,Z^{(j-1)})\otimes Z^{(j)}\) and \(q\gets c_{j-1}(Z^{(j+1)},\ldots,Z^{(2j-1)})\otimes Z^{(2j)}\)
  \State In the clipped variant, \(p\gets C_H(p)\) and \(q\gets C_H(q)\).
  \State \(\hat G_j\gets\) the batch average of \((pq^\top+qp^\top)/2\), a matrix of dimension \(m_j=k_{j-1}d\)
  \State Run \cref{lem:jacobi} on \(\hat G_j\) with tolerance \(\theta_j/4\) and cap \(m_j(m_j-1)\lg(4\Gamma_j/\theta_j)\).
  \State \(U_j\gets\) the columns of \(Q\) with diagonal value above \(\theta_j\)
  \If{the cap was reached or \(U_j\) has more than \(\kappa\) columns}
    \State \Return failure
  \EndIf
\EndFor
\State \Return \(U_1,\ldots,U_n\)
\end{algorithmic}
\end{algorithm}

\begin{lemma}[rank and capture]\label{lem:rank}
Let \(G_j\) be the expectation of \(\hat G_j\) conditional on the previous batches. Suppose these matrices are positive semidefinite of rank at most \(\kappa\). Whenever \(\|\hat G_j-G_j\|_F\le\theta_j/4\) and \(\|\hat G_j\|_F\le\Gamma_j\) hold at all reached levels, TT succeeds, every rank is at most \(\kappa\), and

\[
\|(I-U_jU_j^\top)b\|^2\le3\theta_j/(2\varrho)
\quad\text{if }G_j\succeq\varrho bb^\top .
\]

For the unclipped construction put \(W_j=\int(|v|^2+\sigma^2)^j d\lambda\) and take \(\Gamma_j\ge W_j+\theta_j/4\). Except with probability at most \(\sum_{j=1}^n16\kappa^2d^2 W_{2j}/(N\theta_j^2)\) the algorithm succeeds and, for each positive-mass atom,

\[
\|(I-\Pi_{\hat S_j})v_a^{\otimes j}\|^2
\le\frac{3}{2\varrho_a}\sum_{i=1}^j\theta_i|v_a|^{2(j-i)} .
\]

The cost of all diagonalizations on any run is \(O(\sum_{j\le n}(\kappa d)^4[2+\lg(4\Gamma_j/\theta_j)])\).

\end{lemma}

\begin{proof} Induction and \cref{prop:capture} prove the first statement, including the rank cap. For the plain construction, \(p\) and \(q\) have the same conditional mean \(b(v)=c_{j-1}(v,\ldots,v)\otimes v\) and are independent given \(v\) and earlier batches. Thus \(G_j=\int b(v)b(v)^\top d\lambda\), with \(\|G_j\|_F\le\int |v|^{2j}d\lambda\le W_j\). By \cref{lem:parts} in the at most \(\kappa d\) orthonormal directions of the product subspace, \(\mathbb{E}[|p|^2\mid v]\le\kappa d(|v|^2+\sigma^2)^j\). Conditional independence bounds the mean squared Frobenius error of the average by \(\kappa^2d^2W_{2j}/N\). Markov's inequality, applied to the fresh batch at each reached level, gives the stated failure probability. For an atom, \(G_j\succeq\varrho_ab(v_a)b(v_a)^\top\). The projection residual at level \(j\) splits orthogonally into the inherited residual tensorized with \(v_a\) and the newly discarded part. Iteration gives the displayed sum, also when a subspace becomes empty. The cost follows by substituting \(m_j\le\kappa d\) in \cref{lem:jacobi}. 
\end{proof}

\subsection{Estimating the cell moments}\label{sec:moments}

In a cell, the centered responses have law \(\nu*N(0,2I_d)\), where \(\nu\) has profile \((K,R,v')\). For an integer \(A\ge1\), use \cref{lem:multi-thinning} to form \(2A\) parts of conditional variance \(4A\) from each response. Write \(S_A(x)=\sum_{i=0}^A x^i/i!\). For \(0<\eta_0,\delta<1\), set

\[
\begin{aligned}
\Xi&=(R+2\sqrt{v'A})^2, & \Omega&=2\Xi+8A,\\
\theta_0&=\frac{\eta_0^2}{6k^2AKS_A(\Xi)}, &
\theta_j&=\theta_0\Xi^j,\qquad \Gamma_j=kK\Omega^j+\theta_j/4.
\end{aligned}
\]

Run the unclipped train with \(n=A\) and \(\kappa=k\). At each order \(j\), average \(c_j\) on an independent batch of \(N'\) samples to obtain \(\hat m_j\). Let \(\hat M_j\) denote the tensor with these coordinates in \(\hat S_j\), and set \(\hat M_0=M_0=1\). Storing the train and \(\hat m_j\) suffices.

\begin{theorem}[Hermite levels in polynomial time]\label{thm:levels}
Take batch sizes satisfying

\[
N\ge\frac{32Ak^3d^2K(25/4)^A}{\theta_0^2\delta},
 \qquad N'\ge\frac{8Ak^2KS_A(\Omega)}{\eta_0^2\delta}.
\]

With probability at least \(1-\delta\) the algorithm succeeds and \(\|\hat M_n-M_n\|\le\eta_0\sqrt{n!}\) simultaneously for \(1\le n\le A\). With batch sizes given by these bounds rounded up, the cost is \(\operatorname{poly}(d,k,K,1/\eta_0,1/\delta)e^{O(R^2+v'A)}\) on every run. The constant \(25/4\) uses \(v'\ge4\), which the definition of a profile guarantees (\(v'\ge8\)). \Cref{thm:taper} alone allows \(v'\ge1\).

\end{theorem}

\begin{proof} We bound the moments of the cell law, apply \cref{lem:rank} to the train, and then control the coordinate averages. For \(0\le i\le4A\) each atom satisfies \(\rho_l|\delta_l|^i\le K\Xi^{i/2}\). Indeed, above \(R\) the maximizer of \(x^i e^{-(x-R)^2/(2v')}\) satisfies \(x(x-R)=v'i\) for \(i>0\), and hence \(x\le R+\sqrt{v'i}\). For \(1\le j\le2A\) this yields \(W_j\le2^{j-1}(kK\Xi^j+(4A)^j)\le kK\Omega^j\). Since \(\Xi\ge4v'A\ge16A\), we have \(\Omega/\Xi=2+8A/\Xi\le 5/2\), and \cref{lem:rank} gives failure probability at most \(\delta/2\).

On its success event, summing the atomwise residual bound with weights \(\rho_l\) gives

\[
\|(I-\Pi_{\hat S_n})M_n\|
 \le k\sqrt{3nK\theta_0\Xi^n/2}\le\eta_0\sqrt{n!}/2 .
\]

Here \(\Xi^n/n!\le S_A(\Xi)\) for every retained degree. Conditional on the train, the independent coordinate average has squared error in expectation at most \(kW_n/N'\le n!k^2K S_A(\Omega)/N'\). By Markov's inequality and a union bound, all coordinate errors are at most \(\eta_0\sqrt{n!}/2\) except with probability \(\delta/2\). The cost follows from the uniform rank cap, \(\lg(4\Gamma_j/\theta_j)=O(1+\log(kK/\theta_0)+A)\), and the coordinate recursion. For \(x\ge A\ge1\) the summands of \(S_A(x)\) are increasing, so

\[
S_A(x)\le(A+1)x^A/A!\le (2ex/A)^A .
\]

For \(R=0\), \(\Xi=4v'A\) and \(\Omega=(8v'+8)A\). The exponential dependence of the fitting cost is then \((C(v'+1))^{O(A)}\). 
\end{proof}

\subsection{An exact error identity for the evaluator}\label{sec:evaluator}

For \(L\ge1\) independent standard Gaussian vectors \(h_1,\ldots,h_L\) and any taper weights \((\Lambda_n)_{n=0}^A\), write \(q_\nu=\sum_{n=0}^A\Lambda_nY_n\) and \(\bar h=L^{-1/2}\sum_jh_j\). Define

\[
\hat Q_n=L^{-n/2}\sum_{j_1<\cdots<j_n}
    \langle\hat M_n,h_{j_1}\otimes\cdots\otimes h_{j_n}\rangle,
 \qquad \tilde r=1+\sum_{n=1}^A\Lambda_n\hat Q_n .
\]

The sum is zero when \(n>L\). Start \(V_0=1\) and the remaining \(V_i\) at zero. On reading \(h_j\), update \(V_i\gets V_i+U_i^\top(V_{i-1}\otimes h_j)\) in descending order of \(i\) from \(\min(j,A)\) to \(1\). Then \(\hat Q_n=L^{-n/2}\langle\hat m_n,V_n\rangle\). This evaluates the entire score in \(O(LAk^2d)\) operations.

\begin{proposition}[orthogonal evaluator error]\label{prop:evaluator}
For any fixed fitted train let \(b_{L,n}=\binom{L}{n}n!/L^n\), with \(\binom{L}{n}=0\) for \(n>L\). Then

\[
\begin{aligned}
\mathbb{E}[|\tilde r-q_\nu(\bar h)|^2]
 =\sum_{n=1}^A\frac{\Lambda_n^2}{n!}\big(
   b_{L,n}\|\hat M_n-M_n\|^2
      +(1-b_{L,n})\|M_n\|^2\big).
\end{aligned}
\]

The formula separates error in the fitted coordinates from the error caused by suppressing repeated auxiliary indices. It allows nonsymmetric fitted tensors.

\end{proposition}

\begin{proof} Let \(Q_n\) denote \(\hat Q_n\) with \(M_n\) in place of \(\hat M_n\). The Hermite generating function at \(\bar h\) is a product of generating functions at the \(h_j\) with argument scaled by \(L^{-1/2}\). Its degree-\(n\) coefficient splits by the \(L\) slot degrees, nonnegative integers summing to \(n\). Distinct vectors of slot degrees are orthogonal in product Gaussian space, since in at least one slot the homogeneous Hermite degrees differ. The terms with every slot degree at most one give \(Q_n\) after integration over \(\nu\). Their sum is the orthogonal projection of \(Y_n(\bar h)\) onto the sum of those slot-degree spaces.

For any tensor \(T\), two different increasing \(n\)-tuples of auxiliary indices give orthogonal contractions. The contraction on one subset has second moment \(\|T\|^2\) by independence and isotropy. Consequently \(\hat Q_n-Q_n\) has squared norm \(b_{L,n}\|\hat M_n-M_n\|^2/n!\) and belongs to the retained slot-degree spaces. The identity \(\mathbb{E}[Y_n(\bar h)^2]=\|M_n\|^2/n!\) follows from the Gaussian generating function by extracting equal coefficients for two shifts in \(\exp(s t\langle v,v'\rangle)\) and integrating both shifts. Subtracting the squared norm of \(Q_n\) gives the second term in the formula. Different total degrees are orthogonal as well. 
\end{proof}

To choose \(L\) from the fit, let \(t_n=\|\hat m_n\|/\sqrt{n!}\) and put

\[
V_{\mathrm{fit}}=\frac12\sum_{n=1}^A\Lambda_n^2n(n-1)(t_n+\eta_0)^2,
 \qquad V_{\mathrm{env}}=\tfrac12 A(A-1)(kK)^2S_A(\Xi).
\]

On the event of \cref{thm:levels}, \cref{prop:evaluator} implies

\begin{equation}\label{eq:evaluator-bound}
\mathbb{E}[|\tilde r-q_\nu(\bar h)|^2]\le
   \eta_0^2\sum_{n=1}^A\Lambda_n^2+
   \min(V_{\mathrm{fit}},V_{\mathrm{env}})/L .
\end{equation}

Indeed \(1-b_{L,n}\le n(n-1)/(2L)\) by the union bound for a collision in \(n\) independent draws from \(L\) slots. The normalized moment norm is at most \(t_n+\eta_0\) by \cref{thm:levels}, and \(\sum_{n=1}^A\|M_n\|^2/n!\le(kK)^2 S_A(\Xi)\) by its moment envelope. 
\begin{corollary}[density estimation from anchors]\label{cor:anchors}
Suppose a list of \(M\) anchors independent of the fit sample is given and all positive-mass cells have profile \((K_0/P(m),R,v')\). For \(K_0\ge1\) and \(0<\varepsilon,\delta<1\), a sampler within \(\varepsilon\) total variation is obtained with probability at least \(1-\delta\). The number of training observations and operations is at most \((CdkMK_0/(\varepsilon\delta))^C e^{CR^2}\), with \(C\) depending only on \(v'\). An \(R\)-cover admits \(K_0=1\), \(v'=8\).

\end{corollary}

\begin{proof} Set \(\eta=\varepsilon/8\) and take a frequency batch of size \(C_1 M\lg(2M/\delta)/\varepsilon^2\) rounded up, for a sufficiently large universal \(C_1\). Keep empirical frequencies at least \(\varepsilon/(32M)\). Their full vector is within \(\varepsilon/16\) of \(P\) in \(\ell_1\) with probability at least \(1-\delta/6\). Indeed its expected error is at most the square root of \(M\) divided by the square root of the batch size, and changing one sample changes the error by at most two divided by the batch size, so bounded differences gives the assertion. Chernoff's inequality and a union bound also ensure that all kept cells have \(P(m)\ge p_*=\varepsilon/(64M)\) and all dropped cells have \(P(m)\le\varepsilon/(16M)\) except with probability \(\delta/6\). The first input of \cref{thm:reduction} follows. Use the common profile bound \(K=K_0/p_*\) in all kept cells.

In each kept cell use the rational taper from \eqref{eq:rational}, take \(\eta_0=\eta/(2\sqrt A)\), and choose

\[
L=\max\{1,\lceil 2\min(V_{\mathrm{fit}},V_{\mathrm{env}})/\eta^2\rceil\}.
\]

Equation \eqref{eq:evaluator-bound} and Cauchy--Schwarz give the second input of \cref{thm:reduction}. The same guarantee holds if \(L\) is increased to a perfect square, or a larger predetermined envelope is used. Let \(s\) be the number of conditional samples that \cref{thm:levels} requires at failure probability \(\delta/(3M)\), with its deterministic training schedule. Draw a single fresh batch of size \(\lceil8(s+\lg(3M/\delta))/p_*\rceil\). Each kept cell supplies \(s\) points except with probability at most \(\delta/(3M)\) by the binomial lower-tail bound. Conditional on the cell assignments, its fitting views have the required independent conditional law. A count shortfall or a training abort returns a failure flag. Union bounds now give the stated confidence, and \cref{thm:reduction} gives the total variation guarantee. The deterministic degree is \(O_{v'}(R^2+\log(kMK_0/\varepsilon))\) and \(\Xi\le2R^2+8v'A\), proving the cost bound. 
\end{proof}

We next ask how far the fitting noise can be reduced. Even exact tensor arithmetic leaves randomness in the observed responses. For a fixed polynomial score, the following calculation identifies its least possible variance under distribution-wide unbiasedness. It extends the usual unbiased Hermite estimators of individual moments \cite{WY20} to the full score, including correlations between levels.

\begin{proposition}[minimum variance for an unbiased response score]\label{prop:variance}
Let \(P\) be a real polynomial of degree \(A\) on \(\mathbb R^d\). Observe \(n\ge1\) independent responses \(Y=U+\varsigma Z\) with known \(\varsigma>0\) and \(Z\sim N(0,I_d)\) independent of \(U\). Put \(Q=\exp(-\varsigma^2\Delta/2)P\), where \(\Delta\) is the Laplacian and the series terminates. Among estimators unbiased for \(\nu P=\int P\,d\nu\) for every finitely supported mixing law, the sample average of \(Q(Y)\) has least variance at each such law. The minimum multiplied by \(n\) is

\[
\operatorname{Var}[P(U)]+
\sum_{j=1}^A\frac{\varsigma^{2j}}{j!}\mathbb{E}[\|\nabla^jP(U)\|_F^2].
\]

For unbiasedness restricted to laws with at most \(k\) atoms the same conclusion holds at laws with fewer than \(k\) atoms.

\end{proposition}

\begin{proof} Condition on \(U=u\). The Hermite expansion of \(Q(u+\varsigma Z)\) then has the coefficient \(\varsigma^{|\beta|}\partial^\beta P(u)/\beta!\) at the unnormalized product polynomial \(\operatorname{He}_\beta(Z)\). This follows by applying the generating function to exponentials and extracting polynomial coefficients. The constant is \(P(u)\). Orthogonality and total variance give the display. For any competing estimator \(T\) of finite variance, symmetrize over the data and average out internal randomness. Put \(g(y)=\mathbb{E}[T\mid Y_1=y]\) and \(b=\nu P\). Along the path \((1-h)\nu+h\delta_u\) the expectation of \(T\) is a polynomial in \(h\). Differentiating at zero gives

\[
\mathbb{E}[n g(u+\varsigma Z)-(n-1)b]=P(u).
\]

Choose an atom \(u_0\) of \(\nu\). The function \(ng-(n-1)b-Q\) belongs to \(L^2(N(u_0,\varsigma^2I_d))\) and has zero integral against every Gaussian translate. Differentiating these integrals at \(u_0\) gives zero against each Hermite polynomial, hence the function vanishes by completeness. The orthogonal first-order projection of \(T-b\) onto sums of centered functions of individual responses is therefore \(n^{-1}\sum_i(Q(Y_i)-b)\). Projection decreases variance. The path adds at most one atom, proving the restricted claim. 
\end{proof}

\Cref{sec:num-taper} evaluates this variance floor numerically.

Radial polynomials show the effect of the ambient dimension. For \(P(u)=|u|^{2a}\) the Laguerre generating function gives the polynomial \(Q\) of \cref{prop:variance} as

\[
Q(y)=(-2\varsigma^2)^a\, a!\, L_a^{d/2-1}\big(|y|^2/(2\varsigma^2)\big).
\]

At \(U=0\), Laguerre orthogonality gives minimum variance \((4\varsigma^4)^a a!(d/2)_a/n\), where \((x)_a=x(x+1)\cdots(x+a-1)\). Already \(a=1\) costs \(2d\varsigma^4/n\). Estimating a radial moment from noisy responses therefore has variance growing with the ambient dimension. A known signal space replaces \(d\) by its rank.

\section{Weight-adaptive covers without separation}\label{sec:cover}

It remains to produce the anchors. A component of weight \(w\) supplies a fraction \(w\) of the sample, so heavy components can be located precisely and light ones only coarsely. \Cref{prop:cells}(3) shows that coarse location suffices when the weight is exponentially small in the squared distance to the nearest anchor. A pairwise test trained on the whole mixture gives \(r_i^2=O(\log(dk/(\delta_c w_i)))\) simultaneously for all components above a cutoff. In this section \(n\) is the degree of the pair test, \(S\) a scale, \(\varrho\) a lower bound on the weight of a pair of labels, \(\sigma^2=257/128\) the variance of a difference of test views, and \(\delta_c\) the failure budget of the cover.

\subsection{A clipped pair test on the whole mixture}\label{sec:pair-test}

We first describe the training data and the fresh observations. For two independent mixture draws with labels \(i,i'\), write \(\Delta=\mu_i-\mu_{i'}\). This difference has at most \(\kappa=k^2\) possible values. For training only, thin each raw draw into \(2n\) independent parts conditional on its label. A part difference has covariance \(4nI\). Set \(S_0=\sqrt{17n}\). At scale \(S=2^eS_0\) the normalized training mean is \(u=\Delta/S\) and the noise variance is \(v^2=4n/S^2\le4/17\). Use the clipped tensor train of \cref{sec:products} on disjoint training pairs with

\[
a=\frac{200^{-n/2}}{4\sqrt n},\qquad H=\frac{\kappa d5^n}{a},\qquad
\theta=\frac{\varrho 200^{-n}}{24n},\qquad \Gamma_j=H^2 .
\]

For each fresh panel observation \(X\) draw \(G_0\sim N(0,I_d)\) and form \(X^a=X+G_0/16\) and \(X^b=X-16G_0\). The views are conditionally independent given the label. Differences \(Y=X^a-X'^a\) have covariance \(\sigma^2 I\) with \(\sigma^2=257/128\). Keep \(X^b\) for the anchor average.

The test itself adds an imaginary Gaussian to the observed difference. Put \(\rho_g=\sigma(\sqrt d+\sqrt{2n})\), the \emph{norm gate}. Declare a pair far if \(|Y|>\rho_g\), or if at some scale \(S=2^eS_0\le2\rho_g\) with integer \(e\ge0\),

\begin{equation}\label{eq:pair-statistic}
W=(Y+\mathrm i\sigma G)/S,\qquad
T_S=\big|\Pi_{\hat S_n}W^{\otimes n}\big|^2\ \ge\ 4^{-n}/4 .
\end{equation}

Here \(G\) is an independent standard Gaussian. The Euclidean norm on complex coordinates uses squared moduli. Because the basis of the trained space is real, the same multilinear recursion is evaluated on \(W\) in every slot with two real arrays, and only the training features require independent slots. The number \(E_0\) of tested scales satisfies \(E_0\le E:=2+\lg(1+d)\) by doubling \(S_0\) until it exceeds \(2\rho_g\).

\begin{lemma}[pair test]\label{lem:pair-test}
Let \(0<\zeta<1\) and train each level at each scale on \(N_p\) pairs, where

\[
N_p\ge2.4\cdot10^6\, n^5E_0k^8d^4\, 10^{12n}/(\varrho^2\zeta).
\]

Let \(C_J\) be the absolute constant in the cost bound of \cref{lem:rank}. With probability at least \(1-\zeta\) over training no failure is reported. On this event a fresh pair with \(\Delta=0\) is declared far with probability at most \((32k^2E_0+1)e^{-n}\). If \(|\Delta|\ge S_0\) and \(w_iw_{i'}\ge\varrho\), it is declared near with probability at most \(5e^{-n}\). The diagonalizations cost at most \(C_JnE_0(k^2d)^4(3+20n+\log_2(1536n^2k^4d^2/\varrho))\) operations.

\end{lemma}

\begin{proof} The proof has three parts. We first show that the trained subspaces capture the tensor powers of matched atoms, then that clipping moves the captured mean by little, and finally that a fresh pair is classified correctly, by a second-moment computation.

\emph{Construction.} Fix a scale and a level \(j\) that is reached, so that \(k_{j-1}\le\kappa\). Write \(\lambda\) for the law of the normalized difference \(u\), \(\varrho_u\) for its weights and \(\mathcal F_{j-1}\) for the training batches of the earlier levels. Given \(u\), the features \(p\) and \(q\) are independent and identically distributed, so with \(f_u=\mathbb{E}[C_H(p)\mid u,\mathcal F_{j-1}]\) the clipped Gram matrix has conditional mean \(G_j=\int f_uf_u^\top d\lambda(u)\). Separate clipping preserves this positive semidefinite factorization and its rank bound \(\kappa\) for every configuration of atoms. Each summand has Frobenius norm at most \(|C_H(p)|\,|C_H(q)|\le H^2\). Hence \(\|\hat G_j\|_F\le H^2=\Gamma_j\) on every run, so the rotation cap never causes an abort (\cref{lem:jacobi}). Also \(\mathbb{E}[\|\hat G_j-G_j\|_F^2\mid\mathcal F_{j-1}]\le H^4/N_p\), so \(\|\hat G_j-G_j\|_F\le\theta/4\) except with probability \(16H^4/(N_p\theta^2)\). Since \(H^4/\theta^2=147456\,n^4k^8d^4\,10^{12n}/\varrho^2\), summing over \(n\) levels and \(E_0\) scales gives at most \(2359296\,n^5E_0k^8d^4\,10^{12n}/(N_p\varrho^2)\le\zeta\). On this event \cref{lem:rank} gives, at every scale, \(k_j\le\kappa\) and, because \(G_j\succeq\varrho_uf_uf_u^\top\), \(|(I-U_jU_j^\top)f_u|^2\le3\theta/(2\varrho_u)\le a^2\) for every atom of weight \(\varrho_u\ge\varrho\). For the cost, \(H^2=16n\kappa^2d^2\,25^n200^n\) and \(1/\theta=24n\,200^n/\varrho\), so \(4\Gamma_j/\theta=1536\,n^2k^4d^2\,10^{6n}/\varrho\). \Cref{lem:rank}, \eqref{eq:arithmetic-bounds} and \(\log_210^6\le20\) give the stated bound.

\emph{Clipping bias.} Call an atom \emph{matched} if \(1\le|u|\le2\), and let \(\pi(u)=\mathbb{E}[p\mid u,\mathcal F_{j-1}]\), the coordinates of the projection of \(u^{\otimes j}\) onto \(\hat S_{j-1}\otimes\mathbb R^d\). By \cref{lem:parts} applied to the at most \(\kappa d\) basis tensors of that space, \(\mathbb{E}[|p|^2\mid u]\le\kappa d\,(|u|^2+v^2)^j\le\kappa d\,5^j\) for a matched atom, so

\[
|\pi(u)-f_u|\le\mathbb{E}\big[|p|\,1_{|p|>H}\mid u\big]\le\frac{\kappa d\,5^j}{H}\le a .
\]

Hence \(|(I-U_jU_j^\top)\pi(u)|\le2a\) for a matched atom of weight at least \(\varrho\), and the recursion in the proof of \cref{lem:rank} gives

\[
\varepsilon_u^2:=\big|(I-\Pi_{\hat S_n})u^{\otimes n}\big|^2\le4a^2\sum_{i<n}|u|^{2i}\le4na^2|u|^{2n}\le200^{-n}|u|^{2n}.
\]

\emph{Fresh-pair analysis.} For \(Z=G_1+\mathrm iG_2\) with independent standard real Gaussian vectors, \(\mathbb{E}[\exp(t\cdot Z+s\cdot\bar Z)]=\exp(2t\cdot s)\). Comparing coefficients gives orthogonality of different homogeneous degrees and

\[
\mathbb{E}[|\langle D,Z^{\otimes j}\rangle|^2]=2^j j!\,|\operatorname{Sym}D|^2
\]

for a real tensor \(D\), where \(\operatorname{Sym}\) averages all slot permutations. At \(u=0\), summing over at most \(\kappa\) basis tensors gives \(\mathbb{E}[T_S]\le\kappa n!(2\sigma^2/S^2)^n\). Markov's inequality and \(n!\le e\sqrt n(n/e)^n\) give false-positive probability at most \(32\kappa e^{-n}\) per scale, since \(4e\sqrt n(257/272)^n\le32\). Gaussian norm concentration adds \(e^{-n}\) for the gate.

If \(|\Delta|>2\rho_g\), the norm gate fails with probability at most \(e^{-n}\). Otherwise choose a matched scale and write \(D=(I-\Pi_{\hat S_n})u^{\otimes n}\). Except with probability \(\exp(-S^2|u|^2/(8\sigma^2))\le e^{-n}\), the real projection \(\Re\langle u,W\rangle\) is at least \(|u|^2/2\). For the capture error, contract \(\operatorname{Sym}D\) against \(u\) in \(n-j\) slots. Its remaining tensor has norm at most \(|D||u|^{n-j}\). Homogeneous orthogonality and \(\binom nj\le n^j/j!\) yield

\[
\begin{aligned}
\mathbb{E}[|\langle D,W^{\otimes n}\rangle|^2]
&\le |D|^2|u|^{2n}\sum_{j=0}^n\binom nj^2j!
 \left(\frac{2\sigma^2}{S^2|u|^2}\right)^j\\
&\le 200^{-n}|u|^{4n}\exp(n/4).
\end{aligned}
\]

Thus the residual exceeds \(2^{-n-1}|u|^{2n}\) with probability at most \(4(4e^{1/4}/200)^n\le4e^{-n}\). Outside these events, the projection onto \(\Pi_{\hat S_n}u^{\otimes n}\) alone gives \(T_S\ge4^{-n}|u|^{2n}/4\). The auxiliary imaginary Gaussian cancels all Gaussian pairings within a single tensor. This is what permits reuse of one low-noise difference across all slots. 
\end{proof}

\subsection{Anchors at one weight threshold}\label{sec:tier}

Fix \(0<w\le1\) and \(0<\alpha\le1/2\). Draw \(N_r\) references and \(N_g\) neighbours, where

\[
\begin{aligned}
N_r&=\lceil\lg(8k/\alpha)/w\rceil, &
m_0&=2^{15}(d+2\lg(8N_r/\alpha)),\\
N_g&=\lceil8(m_0+\lg(8k/\alpha))/w\rceil, &
\xi&=\alpha/[8k(N_r+N_g)] .
\end{aligned}
\]

Use \(\varrho=\xi^2\), \(\zeta=\alpha/8\) and the least integer \(n\ge1\) with \((8/3)^n\ge512 k^2 E P/\alpha\), where \(P=N_r(N_g+N_r)\). \Cref{alg:tier} states the procedure.

\begin{algorithm}
\caption{\textsc{Tier}: anchors at one weight threshold}\label{alg:tier}
\begin{algorithmic}[1]
\Require Threshold \(w\), failure budget \(\alpha\), and a sample oracle
\State Train the pair test of \cref{sec:pair-test} on fresh pairs with \(\varrho=\xi^2\), \(\zeta=\alpha/8\) and degree \(n\).
\State Draw \(N_r\) references and \(N_g\) neighbours, and form the views \(X^a,X^b\) of each.
\State Test every reference against every neighbour and every other reference.
\State Call a reference \emph{eligible} if at least \(m_0\) neighbours test near it.
\State In sample order, keep an eligible reference if it tests far from every previously kept reference.
\If{training failed or more than \(k\) references are kept}
  \State \Return failure
\EndIf
\State \Return for each kept reference the average of the held-out views \(X^b\) of its near neighbours
\end{algorithmic}
\end{algorithm}

\begin{lemma}[one tier]\label{lem:tier}
With probability at least \(1-\alpha\) the tier outputs at most \(k\) anchors, and every component of weight at least \(w\) is within \((2+1/32)S_0\) of the list. Its cost is polynomial in \(d,k,1/w,1/\alpha\).

\end{lemma}

\begin{proof} Except with probability \(\alpha/8\), all panel labels have weight at least \(\xi\). Conditional on these labels and successful training, a union bound over the predetermined set of at most \(P\) comparisons costs \(\alpha/8\) by \cref{lem:pair-test}. Thus same-component pairs are near and every near pair has mean distance less than \(S_0\). Costs of training failure, missing a heavy reference and having fewer than \(m_0\) neighbours from a heavy component are each at most \(\alpha/8\), the last two by the binomial tails \(ke^{-wN_r}\) and \(ke^{-wN_g/8}\).

Given the labels and all selection variables, the noise in an average of \(q\ge m_0\) held-out observations has law \(N(0,(257/q)I)\). Its norm is at most \(\sqrt{257/q}(\sqrt d+\sqrt{2\ln(8N_r/\alpha)})\le\sqrt{514/2^{15}}\le S_0/32\) simultaneously with additional failure probability \(\alpha/8\). The neighbour means average to within \(S_0\) of the reference mean. At most one reference is kept per component. Every heavy component supplies an eligible reference, and if it was discarded, its mean is within \(S_0\) of the mean of a kept reference. Applying the preceding anchor-average bound proves the claimed radius. The degree and cost follow from the panel sizes and \cref{lem:pair-test}. 
\end{proof}

\subsection{The weight-adaptive cover}\label{sec:cover-main}

\begin{algorithm}
\caption{\textsc{Cover}}\label{alg:cover}
\begin{algorithmic}[1]
\Require Maximum tier \(U\in\mathbb N_0\), failure budget \(\delta_c\), and a sample oracle
\State \(\mathcal A\gets\varnothing\)
\For{\(u=0,\ldots,U\)}
  \State Run \cref{alg:tier} on fresh data with \(w=2^{-u}\) and \(\alpha=\delta_c/[2(u+1)^2]\).
  \If{the tier reports failure}
    \State \Return failure
  \EndIf
  \State Add the tier's anchors to \(\mathcal A\).
\EndFor
\State \Return \(\mathcal A\)
\end{algorithmic}
\end{algorithm}

\begin{theorem}[weight-adaptive cover]\label{thm:cover}
Let \(0<\delta_c\le1/4\) and \(Q_c=2dk/\delta_c\). With probability at least \(1-\delta_c\), \cref{alg:cover} returns at most \(k(U+1)\) anchors, and every component with \(w_i\ge2^{-U}\) satisfies

\[
r_i^2=\min_{a\in\mathcal A}|\mu_i-a|^2\le164\log(Q_c^{12}/w_i).
\]

The sample and operation budgets are polynomial in \(d,k,2^U,1/\delta_c\).

\end{theorem}

\begin{proof} Sum the tier failure budgets. At the first tier with \(2^{-u}\le w_i\) apply \cref{lem:tier} and the explicit degree estimate of \cref{lem:degree}. It gives \(r_i^2\le17(65/32)^2 n_u\le164\log(Q_c^{12}/w_i)\). Tier \(u\) costs a polynomial in \(d,k,2^u,(u+1)^2/\delta_c\).
\end{proof}

The radius bound has the form \(r_i^2\le a_1\log Q_c+a_2\log(1/w_i)\) with separate coefficients. Equivalently \(w_i\le Q_c^{12}e^{-r_i^2/164}\): a component far from every anchor is light, at a Gaussian rate in its radius. Beyond that radius \cref{prop:cells}(3) supplies a second Gaussian factor. The two combine into one Gaussian in the distance \(D\) from the anchor, with variance determined by \(a_2=164\) and the thinning constant \(16\). The cutoff \(U\), the accuracy and the confidence enter the prefactor alone.

\begin{proposition}[every cell has a Gaussian profile with constant leak variance]\label{prop:profile}
Set \(I_*=\{i:w_i\ge2^{-U}\}\) and \(Z=\sum_{i\in I_*}w_i\). On the event of \cref{thm:cover}, for every \(i\in I_*\) and every anchor \(a_m\), with \(D=|\mu_i-a_m|\),

\begin{equation}\label{eq:constant-leak}
w_i\pi_i(m)\le Q_c^{12}\exp(-D^2/180).
\end{equation}

Consequently, if \(Z>0\), each positive-mass cell of \(F_*=\sum_{i\in I_*}(w_i/Z)N(\mu_i,I)\) has profile \((Q_c^{12}/(ZP(m)),0,v_*)\) with \(v_*=(164+16)/2=90\), where \(P(m)\) is the cell mass under \(F_*\). The leak variance \(v_*\) does not depend on \(U\), \(d\), \(k\) or \(\delta_c\).

\end{proposition}

\begin{proof} By \cref{thm:cover}, \(w_i\le Q_c^{12}e^{-r_i^2/164}\). By \cref{prop:cells}(3), \(\pi_i(m)\le e^{-(D-r_i)_+^2/16}\). Since \(D\le r_i+(D-r_i)_+\), the Cauchy--Schwarz inequality gives

\[
D^2\le\big(\sqrt{164}\cdot\tfrac{r_i}{\sqrt{164}}+\sqrt{16}\cdot\tfrac{(D-r_i)_+}{\sqrt{16}}\big)^2
\le(164+16)\Big(\frac{r_i^2}{164}+\frac{(D-r_i)_+^2}{16}\Big).
\]

Multiplying the two bounds proves \eqref{eq:constant-leak}. For the profile, the conditional weight of atom \(i\) in cell \(m\) of \(F_*\) is \((w_i/Z)\pi_i(m)/P(m)\). Its displacement has norm \(D\), and \(D^2/180=D^2/(2v_*)\).
\end{proof}

\section{Density estimation with a common spherical covariance}\label{sec:main}

\begin{theorem}[density estimation without separation]\label{thm:main}
Let \(F=\sum_{i\le k}w_iN(\mu_i,I_d)\) with arbitrary means and arbitrary weights. There is an algorithm on the machine of \cref{def:machine} that, given \(d\), \(k\) and \(\varepsilon,\delta\in(0,1/2)\), uses \(\mathrm{poly}(d,k,1/\varepsilon)\log(1/\delta)\) samples and operations on every run and outputs, with probability at least \(1-\delta\), a sampler for a distribution \(\bar H\) with \(d_{\mathrm{TV}}(\bar H,F)\le\varepsilon\). Each draw from the returned sampler costs a number of operations polynomial in \(d\), \(k\), \(1/\varepsilon\) and \(\log(1/\delta)\) on every run.

\end{theorem}

We first construct a \emph{base learner}, which has the same guarantee with \(\mathrm{poly}(d,k,1/\varepsilon,1/\delta)\) in place of both cost bounds and is valid for all \(\varepsilon,\delta\in(0,1)\). The logarithmic dependence on \(\delta\) then follows from \cref{thm:confidence}, whose proof uses only the base learner at a constant failure probability.

\begin{proof} The difficulty is that the cover reaches only components of weight at least \(2^{-U}\), whereas the fit is analysed for a mixture without the lighter ones. We resolve it by coupling the fitting batch with a batch from the truncated mixture.

\emph{Base learner.} Let \(\delta_c=\delta/4\), \(Q_c=2dk/\delta_c\) and \(Q=2dk/(\varepsilon\delta)\). For a cutoff \(U\) let \(T(U)\) be a deterministic integer upper bound on the fitting sample budget of \cref{cor:anchors} with \(M=k(U+1)\) anchors, \(K_0=2Q_c^{12}\), \(R=0\), \(v'=v_\ast\), accuracy \(\varepsilon/2\) and failure probability \(\delta/4\). Use the worst-case profile and cell-count bounds in every batch and discard unused draws. For the explicit integer \(C\) below take \(T(U)=\lceil(Q(U+1))^C\rceil\). Choose the least \(U\ge1\) with

\[
k2^{-U}\le\varepsilon/4\qquad\text{and}\qquad k2^{-U}\,T(U)\le\delta/4 .
\]

For fixed \(C\), \((U+1)^C\le2^{U/2}\) for all \(U\) above a constant depending only on \(C\). Thus an integer \(U=O(\log Q)\) meets both inequalities, and the search costs \(\mathrm{poly}(Q)\).

Run \cref{alg:cover} with \(U,\delta_c\), at cost \(\mathrm{poly}(Q)\) by \cref{thm:cover}. On its success event \(Z\ge1-k2^{-U}\ge1/2\), and by \cref{prop:profile} every positive-mass cell of \(F_\ast\) has profile \((2Q_c^{12}/P(m),0,v_\ast)\). Draw \(T(U)\) fresh samples from \(F\) and fit by \cref{cor:anchors}. One draw from \(F\) can be coupled with one draw from \(F_\ast\) so that they differ with probability \(1-Z\le k2^{-U}\). The batches therefore coincide except with probability \(k2^{-U}T(U)\le\delta/4\). With common internal randomness they then give the same fit. By \cref{cor:anchors} applied to \(F_\ast\), the uncapped accepted law is within \(\varepsilon/2\) of \(F_\ast\) except with probability \(\delta/4\), and \(d_{\mathrm{TV}}(F_\ast,F)\le1-Z\le\varepsilon/4\). The failure probabilities sum to \(3\delta/4\). Let \(c\) be a deterministic upper bound for all cell envelopes. On a successful fit the acceptance probability in \cref{thm:reduction} is at least \(1/(4c)\). Stop after \(\lceil4c\lg(8/\varepsilon)\rceil\) proposals and return an independent standard Gaussian on exhaustion or on a reported failure. On a successful fit the cap adds at most \(\varepsilon/8\) to the TV error of \(\bar H\).

\emph{Operations and cost.} All steps use the primitives of \cref{def:machine} and parameters computed by \cref{lem:approximants}. Algorithm TT caps the matrix dimension at \(\kappa d\) and applies \cref{lem:jacobi}. The stated batch sizes are deterministic. The costs in \cref{thm:levels} and \cref{lem:pair-test} are polynomial because \(A,\Xi,n,\log(1/\varrho)=O(\log Q)\). A reported failure returns the Gaussian fallback. The event bounds above include these failures, and identical coupled batches give identical reports.

\emph{Logarithmic confidence.} Freeze the evaluators of the base learner as in \cref{sec:frozen}. \Cref{thm:confidence}, applied with \(\varepsilon/2\) in place of \(\varepsilon\), combines \(N_{\mathrm{rep}}=O(\log(1/\delta))\) independent base runs, each at failure probability \(1/64\) and accuracy proportional to \(\varepsilon\), into a density within \(\varepsilon/2\) of \(F\) with probability at least \(1-\delta\). Training costs \(N_{\mathrm{rep}}\,\mathrm{poly}(d,k,1/\varepsilon)\). A proposal from \(\bar q\) is accepted with probability \(\alpha_{\mathrm{acc}}(x)\ge\lambda/(2c+\lambda)\), where \(c\) is polynomial. Evaluating \(\alpha_{\mathrm{acc}}(x)\) takes the median of \(N_{\mathrm{rep}}\) frozen scores, each a clipped polynomial times Gaussian ratios \(e^{-t}\) with \(t\ge0\). Compute these ratios by the fixed-degree Taylor rule in the proof of \cref{thm:bit}, with a number of terms logarithmic in \(N_{\mathrm{rep}}/\varepsilon\) and in the training sample size, since every weight of \(\bar q\) is a multiple of the reciprocal of \(N_{\mathrm{rep}}\) times a batch size. The number of terms is chosen so that the accepted law moves by at most \(\varepsilon/4\) in total variation and the acceptance probability stays above \(\lambda/(4c+2\lambda)\). Stop after \(\lceil(4c+2\lambda)\lambda^{-1}\lg(4/\varepsilon)\rceil\) proposals and return a standard Gaussian on exhaustion. By \eqref{eq:arithmetic-bounds} this happens with probability at most \(\varepsilon/4\).
\end{proof}

We comment on the role of the cutoff in the last fit. The coupling is an indistinguishability argument. Components lighter than \(2^{-U}\) have total mass at most \(k2^{-U}\), so a batch of \(T(U)\) samples contains none of them with probability \(1-k2^{-U}T(U)\), and on that event the fit sees a sample from \(F_\ast\). The cutoff-independent profile of \cref{prop:profile} makes \(T(U)\) polynomial in \(U\). A uniform radius reaching the lightest retained component would require \(R^2\gtrsim\log(2^U)\), giving only \(T(U)\le\mathrm{poly}(2^U)\). Such a bound does not ensure \(k2^{-U}T(U)\le\delta/4\).

We now make the fitting budget of the base learner explicit. Write \(X=Q(U+1)\) and \(H=2^{16}(v_\ast+1)X^{12}\), and let \(p\) be the odd integer of the rational taper, the least with \(2^p\ge8v_\ast+1\). Put

\[
J=2^{p+1}(48pv_\ast+7),\qquad G=J\lg(48(v_\ast+1)),\qquad C=128G .
\]

The fitting call has accuracy \(e=\varepsilon/2\), failure budget \(\delta_f=\delta/4\) and \(M=k(U+1)\). With \(\eta=e/8\), its retained-cell floor, profile ceiling and per-cell failure budget are

\[
p_*=\frac{e}{64M},\qquad K=\frac{2Q_c^{12}}{p_*}=1024QM Q_c^{11},\qquad
\delta_m=\frac{\delta_f}{3M}.
\]

Thus \(1/p_*,1/\eta\le H\) and \(K,1/\delta_m\le H^2\). The frequency batch, of size \(N_f\), can use \(C_1=2^{15}\) in \cref{cor:anchors}. For \(R=0\) the prescriptions in \cref{sec:reduction,sec:trains} and \(S_A(\Omega)\le(2e\Omega/A)^A\le(48(v_\ast+1))^A\) give

\[
\begin{aligned}
&D_1^2\le10v_\ast\log_2H, \qquad \tau^2\le96v_\ast\log_2H, \qquad A\le J\log_2H,\\
&\Xi=4v_\ast A, \qquad \Omega=(8v_\ast+8)A, \qquad
\max\{S_A(\Omega),(25/4)^A\}\le H^G,\\
&\eta_0^{-2}\le H^7, \qquad \theta_0^{-1}\le H^{G+16}, \qquad N,N'\le H^{3G+50}.
\end{aligned}
\]

Each cell needs \(A(N+N')\) conditional observations. The single raw fitting budget, including the frequency batch, is at most

\[
N_f+\left\lceil8\big(A(N+N')+\lg(3M/\delta_f)\big)/p_*\right\rceil
\le H^{4G}\le X^{128G}.
\]

Here \(X\ge4\), and \(v_\ast=90\) gives \(H\le X^{24}\). With \(p=11\), the fitting-batch exponent is \(C=323931865088<3.24\times10^{11}\). An explicit feasible cutoff is \(U=8Cj\) with \(j=\lg(2Q)+\lg(8C)\). Indeed \(\log_2(U+1)\le4j\), so \((U+1)^C\le2^{U/2}\) and \(U/2\ge C\log_2Q+\log_2(8k/\delta)\). Also \(U\ge\log_2(4k/\varepsilon)\).

\begin{remark}[size of the exponent]\label{rem:exponent}
The fitting exponent is \(C=128\,J\lg(48(v_\ast+1))\) with \(J=2^{p+1}(48pv_\ast+7)\). At \(v_\ast=90\) and \(p=11\) this is \(3.24\times10^{11}\), and the training exponent below is less than \(70C\). Evaluating the same formula at the smallest leak variance allowed by \cref{prop:cells}, \(v'=8\) with \(p=7\), still gives \(7.9\times10^{8}\). The size therefore comes from the product of the taper degree \(A\le J\log_2H\) and the per-degree sampling cost \((48(v_\ast+1))^A\) of \cref{thm:levels}. The cover constants \(164\) and \(17\) enter through \(v_\ast\), and the second figure is the floor of this formula. Forcing \(p\) odd costs about a factor two in \(J\), and the factor \(10^{12n}\) of \cref{lem:pair-test} accounts for the multiple \(70\). The constant \(C\) is explicit and unoptimized. \Cref{thm:main} establishes polynomial complexity for arbitrary means and weights, and lowering \(C\) is an open problem.
\end{remark}

The cover has separate costs. A tier trains \(nE_0\) levels on pairs, with \(10^{12n}/\varrho^2\) in the pair count, and makes at most \(P=N_r(N_g+N_r)\) comparisons, with \(n=O(\log(Q2^U(U+1)^2))\). At inference each proposal costs \(O(LAk^2d+M)\), and the trial limit is \(\lceil4c\lg(8/\varepsilon)\rceil\).

For the full unit-variance learner let \(N_{\mathrm{all}}\) count raw training samples. The least cutoff satisfies \(2^U\le Q^{C+1+o(1)}\) as \(Q\to\infty\). Indeed any \(U=\lceil(C+1+b)\log_2 Q\rceil\) with fixed \(b>0\) eventually meets both cutoff constraints. Uniformly over tiers the panel size is at most \(Q^{C+2+o(1)}\), so \(1/\xi\le Q^{C+4+o(1)}\). The proof of \cref{lem:degree} before substituting the weight gives \(n\le 1+(25\log Q_c+(9/4)U\log 2)/\log(8/3)\). With \(\varrho=\xi^2\), \cref{lem:pair-test} therefore gives

\[
\limsup_{Q\to\infty}\frac{\log N_{\mathrm{all}}}{\log Q}
\le 9+4(C+4)+\frac{12\log 10}{\log(8/3)}\left(25+\tfrac94(C+1)\right)<70C.
\]

The tier and scale counts grow only polylogarithmically. This last bound concerns samples through training. Inference and arithmetic costs follow the separate counts above.

\subsection{Freezing the evaluators}\label{sec:frozen}

Conditional on \(L^{-1/2}\sum_jh_j=y\), the Gaussian vectors used in \cref{sec:trains} have the representation \(h_j=y/\sqrt L+g_j-\bar g\), where \(g_j\) are independent standard Gaussians and \(\bar g=L^{-1}\sum_jg_j\). Draw the \emph{residual family} \((g_j-\bar g)_{j\le L}\) once for each kept cell and keep it fixed. We call these the \emph{frozen residuals}. The evaluator in that cell is now a deterministic clipped polynomial \(P_m(y)\) taking values in \([0,c_m]\). Write \(p_m\) for its empirical cell frequency and use the score and proposal

\begin{equation}\label{eq:frozen-score}
\begin{aligned}
g(x)&=\sum_m p_m\gamma_{a_m}(x)P_m(x-a_m),\\
q(x)&=\sum_m p_m\gamma_{a_m}(x)+(1-\sum_m p_m)\gamma(x).
\end{aligned}
\end{equation}

The sums run over the kept cells. The probabilities are empirical frequencies before renormalization, so \(q\) integrates to one. Set \(g=0\) and \(q=\gamma\) on a reported failure. There is a deterministic polynomial cap \(c\ge1\) with \(0\le g\le cq\) on every run, by the threshold for keeping cells in \cref{cor:anchors}. Impose this cap and the stated limits on the number of anchors and retained ranks explicitly. Both the unknown-variance learner below and the amplification of \cref{sec:bit} compare such scores.

\subsection{An unknown common variance}\label{sec:variance}

The thinning construction needs a scale at which to add noise. Density estimation permits a candidate scale to be assessed by the density it produces. This distinction yields the following extension.

\begin{corollary}[unknown common spherical variance]\label{cor:variance}
Let \(F=\sum_{i\le k}w_iN(\mu_i,\sigma^2 I_d)\) with \(0<\sigma<\infty\) unknown. For \(0<\varepsilon,\delta<1/2\), a real-number learner produces a sampler within total variation \(\varepsilon\) of \(F\) with probability at least \(1-\delta\). The sample and operation counts are polynomial in \(d,k,1/\varepsilon,\log(1/\delta)\), uniformly over \(\sigma\), the weights and the means.

\end{corollary}

\begin{proof} Write \(f\) for the density of \(F\) and set \(t=\varepsilon/100\). We first construct a random interval \([a,b]\) containing \(\sigma\). Take \(m=\lceil 2k\lg 64\rceil\) independent pairs of observations and let \(D\) be the smallest norm of a pair difference. Conditional on any pair of labels, the first coordinate of the difference has density bounded by \(1/(2\sqrt\pi\sigma)\). Hence \(\Pr[D<\sigma/(32m)]\le1/(32\sqrt\pi)\). A pair has identical labels with probability \(\sum_iw_i^2\ge1/k\). Given this event its expected squared distance is \(2d\sigma^2\), so it lies within \(2\sigma\sqrt d\) with probability at least \(1/2\). Consequently \(\Pr[D>2\sigma\sqrt d]\le e^{-m/(2k)}\). Each pilot block therefore fails with probability less than one eighth. Take the median \(z\) of the minima of an odd number of independent blocks, at least \(8\lg(8/\delta)\). Hoeffding's inequality shows that, except with probability \(\delta/8\),

\[
a=\frac{z}{2\sqrt d}\le\sigma\le32mz=b,
\qquad b/a=64m\sqrt d.
\]

If \(z=0\) return the Gaussian fallback. This event has probability zero.

At known scale \(s\), standardize by \(s\), fit the unit-scale model with the frozen residuals of \cref{sec:frozen}, and transform the output back. Denote the frozen evaluator by \(P_j(x)\) in original coordinates for anchor \(a_j\). A successful fit returns the score

\[
g(x)=\sum_j p_j\gamma_{a_j,s}(x)P_j(x),
\qquad 0\le P_j\le c,\quad \sum_jp_j\le1,
\]

where \(\gamma_{a_j,s}\) is the density of \(N(a_j,s^2I_d)\), the \(p_j\) are empirical frequencies and \(c\) is a deterministic polynomial envelope. A reported failure returns \(g=0\). The integrated error bound in \cref{lem:one-cell} holds in expectation over the frozen residuals. At fit accuracy \(t/32\) and failure budget \(1/32\), the proof of \cref{thm:main} bounds this expectation on the successful coupled fit by \(8\eta+2(1-Z)\le t/16\). Markov's inequality therefore gives \(\|g-f_s\|_1\le t\) with probability at least \(3/4\) under the mixture density \(f_s\) of the same means and weights at scale \(s\). Let \(T\) be the deterministic total sample budget for this run, including its cover. Both \(T\) and \(c\) are polynomial in \(d,k,1/t\).

Use a multiplicative grid starting at \(a\), stopping at the last point no greater than \(b\), with ratio \(1+\vartheta\) where \(\vartheta=\min(t,1/(100T))/(4\sqrt d)\). Conditional on a successful pilot there is a grid point \(s\) with \(1\le\sigma/s\le1+\vartheta\). The Gaussian relative entropy formula and Pinsker's inequality give \(d_{\mathrm{TV}}(f,f_s)\le2\sqrt d\,\vartheta\) by convexity over labels. Coupling the \(T\) training observations changes the probability of the preceding score guarantee by at most \(2T\sqrt d\,\vartheta\). Also \(\|f-f_s\|_1\le t\). Independent repetitions at every grid point, using \(\lceil4\lg(8/\delta)\rceil\) runs per point, yield at least one score \(g_*\) with \(\|g_*-f\|_1\le2t\), except with conditional probability \(\delta/8\).

We give the selection details because neither the score masses nor the scale used by the successful fit are known. Index the scores by \(i\) and all their Gaussian terms by \(j\), counting at least one dummy term. Let \(S\) be the total number of terms, \(s_j\) their scales, \(u_j=|x-a_j|^2/(2s_j^2)\) and \(u_0=\min_j u_j\). Put \(\kappa(x)=(2\pi)^{-d/2}a^{-d}e^{-u_0}\). Then \(\int\kappa\le S(b/a)^d\) by bounding the maximum by the sum of the Gaussian terms multiplied by \((s_j/a)^d\). To evaluate each \(g_i/\kappa\), replace every \(e^{-(u_j-u_0)}\) by an approximation in \([0,1]\) of uniform absolute accuracy \(\rho=t/(cS(b/a)^d)\). The remaining scale factors \((a/s_j)^d\) are at most one. The approximate scores \(\tilde g_i\) satisfy \(\|g_i-\tilde g_i\|_1\le t\). Comparisons of these scores use only their computed ratios to the common \(\kappa\). For an exponent \(y>\lg(1/\rho)\) set the approximation of \(e^{-y}\) to zero. Otherwise the Taylor polynomial of \(e^{-y}\) at zero through degree \(\lceil6\lg(1/\rho)\rceil\) suffices by the real Taylor remainder, followed by clipping to \([0,1]\). The operation count is polynomial because \(\lg(1/\rho)=O(\log(cS/t)+d\log(b/a))\).

Let \(\mathcal Y\) contain each set \(\{\tilde g_i>\tilde g_j\}\) and its complement. This is the finite minimum-distance selection principle used for candidate densities \cite{SOAJ14}. Here the finite measures \(G_i(dx)=g_i(x)dx\) can have different total masses. Put \(D_i=\max_{A\in\mathcal Y}|G_i(A)-F(A)|\). Use a fresh holdout to estimate all \(F(A)\) to accuracy \(t\). For each \(G_i\), select a term with probability \(p_j\), assigning the unused mass the value zero. From the selected term draw \(x\sim N(a_j,s_j^2I_d)\) and record \(P_j(x)1_A(x)\) for each set. Its expectation is \(G_i(A)\) and its range is \([0,c]\). For \(J_0\) candidate scores, Hoeffding's inequality and a union bound show that \(O((1+c)^2t^{-2}\lg(8(J_0+1)^3/\delta))\) draws per procedure suffice to estimate all probabilities and integrals to accuracy \(t\) except with probability \(\delta/4\).

Select \(i\) minimizing the estimated discrepancy. Then \(D_i\le D_*+4t\le6t\). For the sign set of \(\tilde g_i-\tilde g_*\), integrate separately over the set and its complement. Replacing \(g_i,g_*\) by their approximants and back costs at most \(4t\) in total. This gives

\[
\|g_i-g_*\|_1\le2(D_i+D_*)+4t,
\qquad \|g_i-f\|_1\le22t .
\]

Normalizing the selected score therefore costs at most \(22t\) in total variation. Its mass exceeds \(1-22t>1/2\). Propose from the component terms with probabilities \(p_j\), treating unused mass as immediate rejection, and accept with probability \(P_j(x)/c\). A cap of \(\lceil2c\lg(1/t)\rceil\) trials and a Gaussian fallback add at most \(t\) to total variation on the stated event. Finally \(\lg(b/a)=O(\log(kd+2))\), so the grid, candidate collection, validation and sampling have the asserted cost. 
\end{proof}

The proof is model selection by holdout: a pilot brackets the scale, each candidate scale is fitted, and fresh data select among the fitted densities.

\begin{proposition}[an absolute input mesh cannot cover all scales]\label{prop:mesh}
Fix a mesh width \(h>0\), a finite observation budget \(n\) and an accuracy \(0<\varepsilon<1/2\). In the one-dimensional unknown-scale model with \(k=1\), any learner observing only nearest-lattice roundings has worst-case failure probability at least \(1/2\) in the supremum over parameters.

\end{proposition}

\begin{proof} Consider means \(\pm h/4\) and common standard deviation \(\sigma\). Under either law a single observation rounds to zero except with probability at most \(2\exp(-h^2/(32\sigma^2))\). The probability that all \(n\) rounded observations vanish therefore tends to one as \(\sigma\downarrow0\). Conditional on this record the distribution of the learner's output is the same for both targets. Their total variation distance is \(2\Phi(h/(4\sigma))-1\), which exceeds \(2\varepsilon\) for small enough \(\sigma\). The two success events on the common record are disjoint. At least one failure probability is thus at least half the smaller probability of that record. Taking \(\sigma\downarrow0\) proves the claim. 
\end{proof}

For this reason \cref{sec:bit} measures the rounding mesh in units of the noise standard deviation.

\section{Confidence amplification and the bit model}\label{sec:bit}

Fractional precision is measured in noise units. Reading a large location also costs its integer bit length. A common proposal for the fitted scores controls finite-precision sampling. In this section \(N_{\mathrm{rep}}\) is the number of independent base runs, \(N_{\mathrm v}\) the number of scalar views in one run, \(h\) and \(h_0\) are rounding meshes and \(\beta\) is the largest encoding length of a reported coordinate.

\subsection{Amplification by a pointwise median}\label{sec:median}

Use the frozen score \(g\) and proposal \(q\) of \eqref{eq:frozen-score} for each base run. Let \(f\) be the density of \(F\) and \(f_*\) that of \(F_*\). On a successful fit the proof of \cref{lem:one-cell} controls the joint absolute error before averaging the residuals. Integrating first in \(y\), then in the residuals, gives \(\mathbb{E}[\|g-f_*\|_1]\le8\eta\) by the same mixture calculation as in \cref{thm:reduction}. Also \(\|f_*-f\|_1\le2(1-Z)\). Set the failure budget of the base learner of \cref{thm:main} to \(1/64\) and choose its accuracy sufficiently small that \(8\eta+2(1-Z)\le\varepsilon/1024\). Conditional on a successful fit, Markov's inequality gives \(\|g-f\|_1\le\varepsilon/16\) except with probability \(1/64\). These constants leave room for finite-precision simulation.

\begin{theorem}[confidence amplification]\label{thm:confidence}
Suppose independent base runs return \((g_j,q_j)\) as above and each satisfies \(\Pr[\|g_j-f\|_1\le\varepsilon/16]\ge13/16\). Let \(N_{\mathrm{rep}}\) be the least odd integer at least \(128\lg(1/\delta)\), \(g_{\mathrm{med}}\) the pointwise median of the \(g_j\), \(\bar q=N_{\mathrm{rep}}^{-1}\sum_jq_j\), and \(\lambda=\varepsilon/8\). With probability at least \(1-\delta\) the density proportional to \(g_{\mathrm{med}}+\lambda\bar q\) is within \(\varepsilon\) of \(F\) in total variation. Its rejection sampler has success probability at least \(\lambda/(2c+\lambda)\) on every training outcome. The number of observations and the real-arithmetic cost of training are \(N_{\mathrm{rep}}\,\mathrm{poly}(d,k,1/\varepsilon)\).

\end{theorem}

\begin{proof} Hoeffding's inequality gives at least \(3N_{\mathrm{rep}}/4\) good scores except with probability \(\exp(-N_{\mathrm{rep}}/128)\). For each \(x\), at least half the scores lie on the side of the median away from \(f(x)\), counting equality. At least \(N_{\mathrm{rep}}/4\) of these are good. Summing their absolute errors pointwise and integrating gives \(\|g_{\mathrm{med}}-f\|_1\le4\varepsilon/16\). For any nonnegative integrable \(a\) of positive mass, \(d_{\mathrm{TV}}(a/\!\int a,f)\le\|a-f\|_1\). Apply this to \(a=g_{\mathrm{med}}+\lambda\bar q\). At least half the scores are at least \(g_{\mathrm{med}}\), so \(g_{\mathrm{med}}\le2c\bar q\). A proposal from \(\bar q\) is accepted with probability

\[
\alpha_{\mathrm{acc}}(x)=\frac{g_{\mathrm{med}}(x)+\lambda\bar q(x)}{(2c+\lambda)\bar q(x)}.
\]

This proves both the law and the acceptance bound. The evaluation procedure below implements the comparison with an independent uniform variate. 
\end{proof}

\subsection{Training on rounded observations}\label{sec:rounding}

Put \(Q_0=2dk/\varepsilon\) and take \(0<\varepsilon,\delta<1/2\). Integer accuracy and confidence parameters can equivalently be supplied as \(\lceil\log_2(1/\varepsilon)\rceil\) and \(\lceil\log_2(1/\delta)\rceil\), rounding the tolerances down to powers of two. We first describe one run with a constant failure budget.

We first make the Jacobi rotations rational. If \(a_{pq}\) is the largest off-diagonal pivot, choose \(c_t=(1-t^2)/(1+t^2)\) and \(s_t=2t/(1+t^2)\) with \(t\) dyadic. The transformed pivot has numerator

\[
a_{pq}(1-6t^2+t^4)+2t(1-t^2)(a_{pp}-a_{qq}).
\]

Its values at zero and at one of \(\pm1/2\) bracket zero. If \(\|A\|_F\le\Gamma\), the pivot as a function of \(t\) has derivative bounded by \(4\Gamma\), by differentiating the sine and cosine representation with angle \(2\arctan t\). Indeed \(4a_{pq}^2+(a_{pp}-a_{qq})^2\le2\Gamma^2\). When \(\mathrm{off}(A)>\xi\) and \(N=m(m-1)/2>0\), \(|a_{pq}|\ge\xi/\sqrt{2N}\). Bisection to interval width \(\xi/(8\Gamma\sqrt{2N})\) gives \(|b_{pq}|\le|a_{pq}|/2\). Orthogonality is exact, and invariance of the Frobenius norm gives

\[
\mathrm{off}(B)^2=\mathrm{off}(A)^2-2a_{pq}^2+2b_{pq}^2
\le(1-3/(4N))\mathrm{off}(A)^2 .
\]

Thus the cap in the Jacobi method increases by at most a constant factor. The rank and capture bounds apply to the resulting rational basis.

Next, each completed view is rounded independently. For a dyadic \(h>0\) define \(S_h(y)=h\lfloor y/h+U\rfloor\) with \(U\) uniform on \([0,1)\). Use fresh uniforms in all coordinates and all views \cite{CFH+22}. Conditional on \(y\) the error has mean zero, absolute value at most \(h\) and variance at most \(h^2/4\). Two inputs rounded with the same uniform disagree with probability at most \(|y-y'|/h\), by counting the grid boundaries between them. In the comparison experiment construct all the Gaussian views of a raw observation first, including its selection view or held-out view, and then apply \(S_h\) separately to each. Conditional on the component, the views remain independent. A view of mean \(v\) and covariance \(\sigma^2 I\) has rounded second-moment operator bounded above by \(vv^\top+(\sigma^2+h^2/4)I\). Tensorizing this bound proves both product estimates of \cref{lem:parts} with the same means. For the fit replace \(\Omega\) by \(\Omega+1\) and enlarge the corresponding batch sizes. Since \(\Xi\ge32A\), \((\Omega+1)/\Xi\le2.5\) still holds. Cell assignment uses its own rounded selection view. If \(D=|\mu_i-a_m|\) and \(r_i\) is the nearest-anchor distance, \cref{prop:cells} gives

\[
\pi_i(m)\le\exp\big(-(D-r_i-2\sqrt d\,h)_+^2/16\big).
\]

The extra displacement comes from the error in the projection onto the unit bisector normal. The fitting views and the assignment remain independent given the component.

For the pair test take \(h\sqrt d\le1\) and a dyadic \(S_0\in[\sqrt{17n},2\sqrt{17n}]\). Use the two independent groups of rounded training views and the capture tolerance \(a\) of \cref{sec:pair-test}. Take a dyadic \(a'\in[a/2,a]\), \(H=\kappa d5^n/a'\) and \(\theta=\varrho(a')^2/2\). Clip \(p\) by multiplying by \(2^{-b}\) for the least nonnegative integer \(b\) putting its norm below \(H\) or at \(H\). The clipped Gram mean has rank at most \(k^2\), since its factors are conditionally independent given the ordered pair of labels. The normalized difference noise has covariance at most \((4n+h^2/2)I/S^2\preceq I\). The capture proof of \cref{lem:pair-test} therefore gives \(|(I-\Pi)u^{\otimes n}|^2\le200^{-n}|u|^{2n}\) after a constant-factor increase in training batch size.

Use a rational norm gate \(G_0\in[\rho_g+2h\sqrt d,\rho_g+2h\sqrt d+1]\) and scales through \(2G_0\). Each real and imaginary coordinate of the fresh score comes from rounded completed views, including a rounded \(\sigma G\). For null or matched atoms couple to the unrounded \(W\). Then \(|W_h-W|\le3h\sqrt d/S\), and \(|W|\le K_W=4+2\sqrt d+2\sqrt{2(n+1)}\) except with probability \(2e^{-n}\) by the two Gaussian norm tails. Choose \(h\) so that \(n(3h\sqrt d/S)(K_W+1)^{n-1}\le2^{-n}/8\) and \(3h\sqrt d/S\le1\). The telescoping tensor product bound gives the same tolerance for the change in the projected norm. Use the squared rational threshold \((3\cdot2^{-n}/8)^2\). At a matched atom the unrounded projected norm is at least \(2^{-n}/2\) except with probability \(5e^{-n}\) by \cref{lem:pair-test}. At a null atom the probability it reaches \(2^{-n}/4\) is at most \(128k^2e^{-n}\) by the same first-moment bound. The norm gate absorbs the rounding displacement of \(Y\) at both the null and separations beyond \(2G_0\). At most \(2E\) scales suffice, and multiplying the degree criterion of \cref{sec:tier} by eight covers the enlarged error bounds. This preserves \(n=O(\log(dk/(w\alpha)))\). Dyadic \(S_0\) increases by at most a factor two. Rounding the panel averages and snapping their anchors to the lattice add at most \(2h\sqrt d\) to the reference-to-anchor error. The eligible-reference rule of \cref{lem:tier} now yields \(r_i\le3S_0\). In \cref{lem:degree} the factor eight increases the coefficient of \(\ln Q_c\) from 26 to at most 27, since \(Q_c\ge8\). Thus \(r_i^2\le612 n_u\le C\log(Q_c/w_i)\) with \(C=20000\). With \(b=2h\sqrt d\), the inequalities \((x-b)_+^2\ge x_+^2/2-b^2\) for \(x\ge0\), and \(r^2/C+(D-r)_+^2/32\ge D^2/(C+32)\) give

\[
w_i\pi_i(m)\le Q_c e^{b^2/16}\exp(-D^2/(C+32)).
\]

Choose the rounded run's cutoff using its own sample budget. Since \(b\le2\), on \(Z\ge1/2\) the retained cells have profile \((4Q_c/P(m),0,v_b)\) with \(v_b=(C+32)/2=10016\). For \(M=k(U+1)\) let \(T_b(U)\) be the deterministic fitting budget of \cref{sec:reduction,sec:trains} with this profile, the enlarged moment bound \(\Omega+1\) and the base accuracy and failure allocations of \cref{thm:main}. Use the least \(U\ge1\) with \(k2^{-U}\le e_b/4\) and \(k2^{-U}T_b(U)\le\delta_b/4\), where \(e_b,\delta_b\) are the base accuracy and failure probability. The rational taper with the fixed \(v_b\) gives \(T_b(U)\le(Q_0(U+1))^{C_b}\) for a universal integer \(C_b\) after enlarging the power for rounding. Hence \(U=O(\log Q_0)\), and coupling the entire rounded fitting batch costs at most \(\delta_b/4\). A dyadic \(h\) with \(\log_2(1/h)=O(\log^2 Q_0)\) suffices for every degree and scale in a base run.

We now pass from the reported observations to the comparison experiment. Let \(N_0\) and \(N_{\mathrm v}\) be deterministic polynomial bounds in \(Q_0\) for the raw sample count and scalar view count of one run. Jitter every reported coordinate uniformly inside its rounding bin of width \(h_0=2^{-p_0}\). The density of the resulting observation is the bin average of \(f\). In one coordinate, integrating the fundamental theorem of calculus over two points in the same bin bounds the \(L^1\) change by \(h_0\|\partial_j f\|_1\). Coordinatewise bin averaging is an \(L^1\) contraction and \(\|\partial_j f\|_1\le\sqrt{2/\pi}\). Taking \(h_0\le1/(4096dN_0)\) couples the batch to exact observations except with probability \(1/4096\). Generate digits of the jitter and Gaussian auxiliaries as needed. A Gaussian can be generated as a refinable real from fair bits, with polynomial expected cost in the precision \cite{Kar16}. Reject the base run if an auxiliary exceeds a Gaussian tail cutoff of order \(\sqrt{\log(2N_{\mathrm v})}\), chosen to make the rejection probability at most \(1/4096\). A fixed cap on the digit-generation work at \(4096\) times a uniform expected polynomial bound adds at most \(1/4096\) failure probability. The jitter and the thinning coefficients can then be approximated so that the absolute error in each completed scalar is at most \(h/(4096N_{\mathrm v})\). Their required fractional precision is \(O(\log^2 Q_0)\) since the coefficient of the raw observation is one. Using the first \(t\) bits of its rounding uniform, with \(2^{-t}\le1/(4096N_{\mathrm v})\), adds at most \(2^{-t}\) probability that a given rounding comparison differs. A union bound couples all the simulated lattice views with those of the comparison experiment except with probability less than \(1/64\), including the preceding exceptional events. The rounded values can now be processed with exact rational operations.

It remains to bound the word lengths. Use the deterministic auxiliary count from the envelope \(V_{\mathrm{env}}\) in \eqref{eq:evaluator-bound}, and round the coordinate tolerance down to a dyadic number within a factor two of its stated value. Increase the auxiliary count \(L\) to an integer square. The rational weights in \eqref{eq:rational} apply for every rational leakage variance. In the fit \(R=0\) for the spatial profile and \(\Xi=4v'A\) is rational. Use upper integer bounds for the radii \(D_1\) and for \(\tau-2D_1\), then compute \(c_m=\exp^+(\tau D_1)\) with the same bounds. Thresholds based on squared norms use rational comparisons. The taper has rational coefficients by its finite product formula. The Jacobi pivot parameter has \(O(1+\log(m\Gamma/\xi))\) bits, independent of the denominators of the current matrix. A common denominator for the new basis divides the product of the rotation denominators in that one call, and the conjugated matrix uses their product squared times its input denominator. Every call starts from an identity basis. A feature of level \(j\) contains each previous basis once and at most \(j\) lattice inputs. Its denominator divides the product of their common denominators. The dyadic clipping shifts over a whole batch have a common denominator given by their maximum shift. That shift is bounded by the bit length of the unscaled feature, and averaging adds only the bit length of the sample count. Numerator bounds follow by multiplying at most \(j\) view norms and using that all basis maps are contractions. The same argument applies to the coordinate averages and evaluator recursion. Hence all rational words have polynomial bit length in \(Q_0\) and the largest encoding length \(\beta\) of a reported coordinate. The smaller \(O(\log^2 Q_0)\) bound refers to the stochastic training mesh.

\subsection{Computing the amplified sampler}\label{sec:sampler}

\begin{theorem}[logarithmic confidence on a Turing machine]\label{thm:bit}
For \(Q_0=2dk/\varepsilon\) a randomized Turing machine reads \(O(\log(1/\delta))\,\mathrm{poly}(Q_0)\) observations rounded coordinatewise to \(p_0=O(\log Q_0)\) fractional bits. The training time and the output length are at most \(O(\log(1/\delta))\) times a polynomial in \(Q_0\) and \(\beta\), with a sequential list encoding of the independent fits and the integer repetition count. With probability at least \(1-\delta\) the output program defines a continuous law within \(\varepsilon\) of \(F\) in total variation. A draw can be generated to any requested absolute coordinate precision \(2^{-p}\) in expected time polynomial in \(Q_0,\beta,\log(1/\delta)\) and \(p\). The distribution in the guarantee is that of the underlying continuous draw.

\end{theorem}

\begin{proof} First freeze rational approximations of the Gaussian residuals. For each cell the untruncated score is a sum of products of at most \(A\) linear factors with train coefficients. If two arguments to each factor differ coordinatewise by at most \(\ell\) and every factor has coordinates at most \(T\) before perturbation, telescoping bounds the score change by

\[
\ell\sum_{j=1}^A \|\hat M_j\| L^{-j/2}\binom Lj
 j\sqrt d\,\big(\sqrt d(T+1)\big)^{j-1},\qquad 0\le\ell\le1 .
\]

Clipping is a contraction. On a successful fit \(\|\hat M_j\|/\sqrt{j!}\le kK e^{\Xi/2}+\eta_0\) by the weight estimate and \cref{thm:levels}. The bound can be checked on the coordinate vectors, and the run may return a failure score if it fails. For \(y\) distributed as \(\gamma\), the residuals used in \(P_m(y)\) and the argument \(y\) have Gaussian coordinate tails. Choose \(T=O(\sqrt{\log Q_0})\) so that, for a cell, all factors and the Gaussian values used to freeze them are bounded as above outside a set of joint probability at most \(\varepsilon/(8192c)\). This follows by the union bound over a polynomial number of coordinates. Approximate each generating Gaussian \(g_j\) to \(\ell/2\) before subtracting its average, with a fixed cutoff on exceptionally large values. Since the logarithm of the displayed sensitivity is \(O(\log^2 Q_0)\) on a successful fit, the expected \(L^1\) difference between the exact and dyadically frozen scores is at most \(\varepsilon/2048\) for a suitable \(\ell=2^{-O(\log^2 Q_0)}\). Here the averaging over each cell uses its own translated \(\gamma\), and the weights sum to at most one. A constant-probability work cap as in the training construction contributes at most \(1/64\) failure. Combining Markov's inequality, the base fit bound, and the training coupling leaves single-run success probability at least \(13/16\) at error \(\varepsilon/16\). Run independent copies and apply \cref{thm:confidence}. The training operations concatenate their output descriptions and do not compare the fitted densities.

It remains to implement the accept decision for their median. Discard zero-weight terms from the representation of \(\bar q\), let \(b_*>0\) be the smallest remaining weight, and write \(a_l\) for all the corresponding centers, with repetitions allowed. Put

\[
t_l(x)=\tfrac12|x-a_l|^2-\min_i\tfrac12|x-a_i|^2.
\]

Factor the common Gaussian value out of both numerator and denominator of \(\alpha_{\mathrm{acc}}(x)\). Each envelope is now \(e^{-t_l(x)}\in[0,1]\), and the scaled proposal is at least \(b_*\) since a minimizing center has exponential equal to one. Scaling all the scores by the same positive number scales their median by that number. To approximate \(e^{-t}\) at a rational \(t\ge0\) to \(2^{-b}\), return zero for \(t>b\), and otherwise use its Taylor polynomial with \(8(b+1)\) terms. The real Taylor remainder is bounded by \(t^j/j!\le(eb/j)^j\) for the first omitted degree \(j\). The number of operations is independent of the squared distance itself. This also implements the evaluations in \cref{thm:confidence} on the real-arithmetic machine, where \(x\) is exact.

For a continuous proposal on the Turing machine first enclose each coordinate in a rational unit interval. On this enclosed box and a unit margin, let \(2^{\Theta}\) bound the absolute values of coordinates, centers, residuals and train coefficient norms. Such a \(\Theta\) is computable from the input words and interval endpoints. The squared distances and their minimum have maximum-norm Lipschitz constants at most \(4d2^{\Theta}\). The exponential on the negative half-line and clipping are contractions. The product bound above with \(T=2^{\Theta+2}\) shows that a Lipschitz constant for each clipped score has logarithm \(O((A+1)(\Theta+\log(dL+2))+\log(A+2))\), after increasing \(\Theta\) by a constant. A median is \(1\)-Lipschitz for maximum-norm perturbations of its entries. The scaled denominator bound \(b_*\) and the numerator bound \(c+\lambda\) therefore give a computable upper bound for the logarithm of the Lipschitz constant of \(\alpha_{\mathrm{acc}}\), polynomial in the input word lengths, \(\Theta\), \(Q_0\) and \(\log N_{\mathrm{rep}}\). Evaluate at a dyadic approximation to \(x\) sufficiently fine for this bound, and choose exponential errors smaller than \(2^{-j}b_*^2/(16(c+1))\). The convex sums and denominator bound then yield a certified enclosure of \(\alpha_{\mathrm{acc}}(x)\) of width at most \(2^{-j}\) after increasing the precision by a fixed number of bits. The train is evaluated by the recursion of \cref{sec:trains}, without expanding its tensors. Its intermediate rational words obey the same product bound as in training, now with the additional input precision. All costs at stage \(j\) are polynomial in these quantities and \(j\).

Compare this enclosure with the first \(j\) bits of an independent uniform variate. Conditional on the proposal and the digit stream used to generate it, the probability that the decision is still unresolved is at most \(3\cdot2^{-j}\). The integer lengths of Gaussian coordinates in a proposal have moments of every fixed order after accounting for the encoded center. Summing the polynomial refinement costs against this geometric tail gives a polynomial expected cost per proposal. An accepted proposal retains its Gaussian digit stream, from which any further requested digits can be read. The acceptance floor in \cref{thm:confidence} bounds the expected total cost by the mean trial cost divided by \(\lambda/(2c+\lambda)\). This proves the assertion even for atypical trained scores. 
\end{proof}

\begin{proposition}[confidence lower bound]\label{prop:confidence-lower}
For \(0<\varepsilon,\delta\le1/4\), TV accuracy \(\varepsilon\) at confidence \(1-\delta\) requires \(\Omega(\varepsilon^{-2}\log(1/\delta))\) observations, even for one Gaussian on the line.
\end{proposition}

\begin{proof} The unit-variance Gaussians with means \(\pm6\varepsilon\) are more than \(2\varepsilon\) apart in TV. The output law therefore yields a test with each error at most \(\delta\). The optimal equal-prior error from \(N\) observations is \(\Phi(-6\varepsilon\sqrt N)\), giving the lower bound.\end{proof}

\begin{corollary}[a calibrated noise unit]\label{cor:noise-unit}
Suppose the instrument reports a positive dyadic scale \(a\) and the coordinates of \(X/a\) rounded as in \cref{thm:bit}. Let \(N\) be the deterministic sample cap of that theorem at accuracy \(\varepsilon/2\) and failure probability \(\delta/2\). Conditional on a fixed or independently calibrated \(a\), assume \(|\sigma/a-1|\le c\min(\varepsilon/\sqrt d,\delta/\sqrt{Nd})\) for a sufficiently small universal \(c\). Applying that theorem and scaling its output by \(a\) gives TV error at most \(\varepsilon\) with probability \(1-\delta\). The additional bit cost is polynomial in the encoding length of \(a\) and the requested output precision.

\end{corollary}

\begin{proof} Compare mixtures with identical latent labels and means and respective noise scales \(\sigma/a\) and one in scaled coordinates. Their one-sample relative entropy is at most \(d(t^2-1-2\log t)/2\) by marginalization, where \(t=\sigma/a\). For \(|t-1|\le1/2\) this is at most \(C d|t-1|^2\). Rounding cannot increase relative entropy. Independence and Pinsker's inequality bound the training-record TV by \(C\sqrt{Nd}|t-1|\), and the unrounded target discrepancy by \(C\sqrt d|t-1|\). Transfer the success event and apply the triangle inequality. Binary rescaling of an output digit stream uses the bits of \(a\) and, if required, extra digits proportional to its positive binary exponent, which we charge in the ordinary rational encoding. 
\end{proof}

\section{Complements}\label{sec:complements}

This section is independent of the main line. It treats degree-zero fits on a fitted partition, certificates computed from held-out responses, positive reconstruction of the mixing law, and the directions that carry mixing information when the common covariance is unknown. In this section \(n\) is a number of fresh observations, \(M\) a number of cells or sites, and \(\varsigma^2=1+1/t\) the variance of a response view.

\subsection{Posterior centroids}\label{sec:centroids}

Write \(U\) for the latent location, so \(F\) has density \(\mathbb{E}[\gamma_U]\). For \(t>0\), form \(S_\pm=X\pm\sqrt t\,g\) and \(Y_\pm=X\mp g/\sqrt t\). Given \(U\), each pair has independent Gaussian noises with covariances \((1+t)I\) and \(\varsigma^2I\), where \(\varsigma^2=1+1/t\), since its cross-covariance vanishes. Reversing the sign swaps the views at \(t=1\). Choose any measurable partition \(c\) into at most \(M\) cells, possibly fitted on an independent pilot. Condition on that pilot throughout. For \(C=c(S_+)\) define \(p_m=\Pr[C=m]\), \(b_m=\mathbb{E}[U\mid C=m]\) and \(v_m=\mathbb{E}[|U-b_m|^2\mid C=m]\) on positive-mass cells. Write \(V=\sum_m p_mv_m\) for the \emph{dispersion}.

\begin{proposition}[posterior centroid approximation]\label{prop:centroid}
The mixture \(H=\sum_m p_m\gamma_{b_m}\) satisfies \(d_{\mathrm{TV}}(F,H)\le\varphi(1)V\). The constant is sharp in every dimension. Given \(n\ge1\) fresh observations, fit weights by observed cell frequencies and centers by cell averages of the response view \(Y_+\), omitting empty cells. The resulting mixture \(\widehat H\) satisfies

\[
\mathbb{E}[d_{\mathrm{TV}}(F,\widehat H)]\le \varphi(1)V+
\frac{1}{2\sqrt n}\sum_m\sqrt{p_m(1-p_m)}+
\frac{\varphi(0)}{\sqrt n}\sum_m\sqrt{p_m(v_m+\varsigma^2 d)}.
\]

The last two terms sum to at most \(\tfrac12\sqrt{(M-1)/n}+\varphi(0)\sqrt{M(V+\varsigma^2 d)/n}\). These bounds hold for any mixing law of finite second moment.

Set \(L_m^\pm=\mathbf1_{\{c(S_\pm)=m\}}\) and \(B_m^\pm=L_m^\pm(Y_\pm-b_m)\), let a bar denote the average of the two signs for one observation, and write \(\operatorname{sd}(Z)^2=\mathbb{E}[|Z-\mathbb{E}[Z]|^2]\). If both sign orientations are pooled, the sampling terms can be replaced by

\[
\frac{1}{\sqrt n}\sum_m\left(\tfrac12 \operatorname{sd}(\bar L_m)+\varphi(0)\operatorname{sd}(\bar B_m)\right),
\]

which is no larger. Indeed \(\operatorname{sd}(\bar Z)^2=\operatorname{sd}(Z^+)^2-\tfrac14\mathbb{E}[|Z^+-Z^-|^2]\) for \(Z=L_m\) and \(Z=B_m\).
\end{proposition}

\begin{proof} Put \(h=U-b_C\). Expand the translated Gaussian density about \(b_C\) along \(h\). The linear term averages to zero conditional on \(C\). For \(Z\sim N(0,1)\), rotation and integration give \(\|D_h^2\gamma\|_1=|h|^2\mathbb{E}[|Z^2-1|]=4\varphi(1)|h|^2\). Translation preserves this norm. Integrating the second derivative with Taylor weight \(1-u\) on \(0\le u\le1\), then taking half the \(L^1\) norm, proves the bound. The scalar case also follows by Gaussian regularization from Lemma 2 of Christofides and Vaggelatou \cite{CV09}, since contraction to the mean has Zolotarev distance of order two equal to half the variance.

For one cell take \(U=rW\) with \(\mathbb{E}[W]=0\) and covariance \(\Sigma_W\). Continuity of translations in \(L^1\) and the preceding integrable bound on \(D_W^2\gamma\) give, for \(G\sim\gamma\),

\[
\lim_{r\downarrow0}r^{-2}d_{\mathrm{TV}}(\mathbb{E}[\gamma_{rW}],\gamma)
=\tfrac14\mathbb{E}[|G^\top \Sigma_WG-\operatorname{tr}\Sigma_W|].
\]

Choosing \(W=\pm e_1\) with equal probabilities gives \(\varphi(1)\). Equal-covariance Gaussian translates at distance \(r\) have total variation \(2\Phi(r/2)-1\le\varphi(0)r\). The weight error costs at most \(\tfrac12\sum_m|\widehat p_m-p_m|\). For pooled centers, \(\widehat p_m(\widehat b_m-b_m)\) is the sample average of \(\bar B_m\), interpreted as zero for empty cells. The independent observations therefore give the stated standard deviations divided by \(\sqrt n\). With one sign the same argument uses \(\operatorname{sd}(L_m^+)^2=p_m(1-p_m)\) and \(\operatorname{sd}(B_m^+)^2=p_m(v_m+\varsigma^2d)\) by conditional independence. Cauchy--Schwarz gives the simplification, and exchangeability of the two signs gives the variance identity.
\end{proof}

The sampling bound holds for arbitrarily small cell masses. Increasing \(t\) lowers the response noise and changes \(V\) through the noisier selection view. In one cell the pooled response is \(X\), which reduces \(V+\varsigma^2d\) to \(V+d\).

The degree-zero approximation of \cref{sec:reduction} uses the anchors in place of the centroids. It equals \(F\) precisely when \(\sum_m p_m\delta_{a_m}\) is the law of \(U\). On an equally weighted regular simplex with anchors at the means this holds by symmetry.

\subsection{Selecting and assessing a fitted partition}\label{sec:certificate}

The constant in \cref{prop:centroid} is sharp for small rank-one displacements. For a dispersed or high-rank cell the trace bound exceeds the error. For fitted centers \(z_m\), an independent response gives

\[
\mathbb{E}[|Y_+-z_C|^2]-d\varsigma^2
=V+\sum_m p_m|z_m-b_m|^2 .
\]

Thus squared response loss estimates a population upper bound for \(V\), consistently under a finite second moment. A finite-sample upper limit for \(V\) needs additional tail information. A mass \(p\) at an arbitrarily distant point makes \(V\) arbitrarily large, and \(n\) observations miss it with probability \((1-p)^n\).

A density certificate can tolerate such components. Condition on a fitted partition and a probability mixture \(H_q=\sum_m q_m\gamma_{z_m}\). Its centers and weights may be arbitrary. Choose a bandwidth \(0<\lambda<1/(2\varsigma^2)\), and let \(a=1-2\lambda\varsigma^2\), \(A_\lambda=a^{-d/2}\), \(c_0=\varphi(0)\) and \(c_1=\varphi(1)\). On a fresh pair use \(C=c(S_+)\) and the residual \(R=Y_+-z_C\) to compute the kernel value \(\kappa_\lambda\), a vector \(h\) and a scalar \(\ell\),

\[
\begin{aligned}
\kappa_\lambda&=A_\lambda e^{-\lambda |R|^2/a},\qquad h=\kappa_\lambda R/a,\\
\ell&=1-\kappa_\lambda+c_1 \kappa_\lambda\big(|R|^2/a^2-d\varsigma^2/a\big).
\end{aligned}
\]

For the empirical average \(P_n\) over \(n\) independent pairs and \(I_m=\mathbf1_{\{C=m\}}\), define the \emph{certificate}

\[
\widehat{\mathcal B}=P_n\ell+c_0\sum_m |P_n(I_m h)|
                  +\tfrac12\sum_m|P_n I_m-q_m| .
\]

\begin{proposition}[observable partition selection]\label{prop:certificate}
Fit \(L\) candidate specifications before examining a common independent validation sample, counting each bandwidth separately. Write \(\mathcal B\) for the population version of \(\widehat{\mathcal B}\). For every mixing law, \(d_{\mathrm{TV}}(F,H_q)\le\mathcal B\). Simultaneously for all candidates, with probability at least \(1-\delta\),

\[
\begin{aligned}
d_{\mathrm{TV}}(F,H_q)&\le\mathcal B\le\widehat{\mathcal B}+
 \min_{\theta\in\Theta}\left(\theta S_n+\frac{\log(JL/\delta)}{n\theta}\right),\\
S_n&=\left(\sqrt{P_n\ell^2}+c_0\sqrt{P_n|h|^2}+\tfrac12\right)^2 .
\end{aligned}
\]

Here \(\Theta\) is any predetermined grid of \(J\) positive values at most \(1/(2D)\), where

\[
D=1+A_\lambda+\frac{c_1 A_\lambda}{a}\left(\frac1{e\lambda}+d\varsigma^2\right)
       +\frac{c_0 A_\lambda}{\sqrt{2e\lambda a}}+\tfrac12 .
\]

Selecting the candidate with the smallest upper endpoint preserves the guarantee.
\end{proposition}

\begin{proof} Put \(W=U-z_C\) and \(w=e^{-\lambda|W|^2}\). Conditional on \(U,C\), \(R\) has law \(N(W,\varsigma^2I)\). Gaussian integration gives \(\mathbb{E}[\kappa_\lambda\mid U,C]=w\). Differentiation in \(W\) and \(\lambda\) gives \(\mathbb{E}[h\mid U,C]=wW\) and \(\mathbb{E}[-\partial_\lambda \kappa_\lambda\mid U,C]=w|W|^2\). In the translated density difference use Taylor's formula from \cref{prop:centroid} on the fraction \(w\) and charge at most one in total variation on the remaining fraction. The resulting quadratic cost plus the discarded mass equals \(\mathbb{E}[\ell]\). Within each cell the integrated linear term costs \(c_0|\mathbb{E}[I_m h]|\), since \(\|D_u\gamma\|_1=2c_0|u|\). Changing the weights costs at most half their absolute discrepancy. This proves the population bound.

Choose unit-ball vectors \(u_m\) and signs \(s_m\) attaining the population norms and absolute values. Set \(Z=\ell+c_0\langle u_C,h\rangle+s_C/2\). The empirical supremum over these choices gives \(\mathcal B-\widehat{\mathcal B}\le\mathbb{E}[Z]-P_n Z\). One has \(|Z|\le D\) by maximizing \(r^j e^{-\lambda r^2/a}\) for \(j=1,2\), and \(P_n Z^2\le S_n\) by the triangle inequality. Since \(e^{-x-x^2}\le1-x\) for \(|x|\le1/2\), independence bounds the expectation of \(\exp(n\theta(\mathbb{E}[Z]-P_n Z)-n\theta^2 P_n Z^2)\) by one. Markov's inequality and a union bound over the grid and the candidates prove the claim.
\end{proof}

The term \(P_n|h|^2\) can be reduced to \(\sum_m\lambda_{\max}(P_n(I_mhh^\top))\). Taking \(\lambda\varsigma^2\) of order \(1/d\) keeps \(A_\lambda\) bounded. At true centroids and weights with \(V<\infty\), dominated convergence sends \(\mathcal B\) to \(c_1 V\) as \(\lambda\downarrow0\). For fixed candidates the endpoints converge to their population bounds, by the law of large numbers for the bounded observables, when the grid includes a value of order \(n^{-1/2}\) and \(\log(JL)=o(\sqrt n)\). Selection then asymptotically attains the smallest population certificate in the menu. In the experiments of \cref{app:fifty} the selected upper limits are about four times the realized error.

\begin{corollary}[comparison within a fixed menu]\label{cor:partition-menu}
Under the conditions of \cref{prop:certificate}, for each specification put \(H_0=A_\lambda/\sqrt{2e\lambda a}\) and \(e_n=D\sqrt{2\log(2L/\delta)/n}\). Selecting the smallest \(\widehat{\mathcal B}+e_n\) gives TV loss at most the minimum, over the menu, of

\[
\mathcal B+(c_0 H_0+1/2)\sqrt{M/n}+2e_n
\]

with probability at least \(1-\delta\).
\end{corollary}

\begin{proof} Convexity and Cauchy--Schwarz give \(0\le\mathbb{E}[\widehat{\mathcal B}]-\mathcal B\le(c_0 H_0+1/2)\sqrt{M/n}\) by summing the cellwise standard deviations. Changing one validation observation moves \(\widehat{\mathcal B}\) by at most \(2D/n\). Bounded differences gives \(|\widehat{\mathcal B}-\mathbb{E}[\widehat{\mathcal B}]|\le e_n\) simultaneously over the menu. The selected endpoint bounds its TV loss and is at most the displayed expression for each candidate.
\end{proof}

We turn to the construction of a partition. For separated mixtures a pilot can supply the small dispersion required by \cref{prop:centroid}. Distance concentration also underlies classical algorithms for separated mixtures \cite{DS07}. Here the common noise term \(2d\) cancels in comparing squared distances, yielding a dispersion bound for fresh, noisier selection views.

\begin{corollary}[one update from separated pilot seeds]\label{cor:pilot}
Suppose the \(k\ge2\) distinct means have minimum distance \(\Delta\) and positive weights at least \(w\). On \(n_0\) raw pilot observations start with any one observation. Repeatedly select the observation farthest from its closest seed until there are \(k\) seeds. Assign each observation to its nearest seed and replace the seeds once by the cell averages. For \(0<\alpha<1\) put \(x=\log(4n_0^2/\alpha)\), \(y=\log(4k/\alpha)\) and \(u=(\sqrt d+\sqrt{2y})/\sqrt{n_0w/2}\). Assume

\[
n_0w\ge8y,\qquad \Delta^2>4\sqrt{dx}+4x+4\sqrt{(d+\Delta^2)x}.
\]

With probability at least \(1-\alpha\) the updated sites are within \(u\) of their respective means. For their Voronoi partition and \(D_{ij}=|\mu_i-\mu_j|\), the dispersion on the selection view satisfies

\[
V\le\sum_i w_i\sum_{j\ne i}D_{ij}^2
\exp\big[-(D_{ij}-2u)_+^2/(8(1+t))\big].
\]

There is no upper diameter in the assumptions.
\end{corollary}

\begin{proof} Conditional on pilot labels, the squared distance of two observations from means at distance \(D\) has the law of \(2T\), with \(T\) noncentral chi-square of \(d\) degrees of freedom and noncentrality \(D^2/2\). For noncentrality \(b\), its moment generating function is \((1-2s)^{-d/2}\exp(bs/(1-2s))\). The centered log moment at \(-s\) is at most \((d+2b)s^2\) for \(s>0\), and at \(s\in(0,1/2)\) with \(b=0\) is at most \(ds^2/(1-2s)\). Chernoff bounds thus give the within-pair upper endpoint \(2d+4\sqrt{dx}+4x\) and between-pair lower endpoint \(2d+D^2-4\sqrt{(d+D^2)x}\), each with failure probability at most \(e^{-x}\). The latter increases for \(D\ge\Delta\) under the hypothesis. A union bound gives a strict gap for every pair. All component counts are at least \(n_0w/2\) except with probability \(ke^{-n_0w/8}\). The gap forces one seed per component and pure assignments. Conditional only on the labels, Gaussian concentration places each component's sample average within \(u\) except with probability \(e^{-y}\). Intersecting the events gives the claim about sites.

Let \(z_i,z_j\) be the sites of means \(i,j\) and put \(a=\mu_i-z_i\), \(c=\mu_i-z_j\). The signed distance from \(\mu_i\) to the bisector of the two sites is \((|c|^2-|a|^2)/(2|c-a|)\). Since \(|c|+|a|\ge|c-a|\), we have \(|c|^2-|a|^2\ge(|c|-|a|)\,|c-a|\) when \(|c|\ge|a|\), and \(|c|-|a|\ge D_{ij}-2u\). The distance is therefore at least \((D_{ij}-2u)/2\) when \(D_{ij}>2u\). The Gaussian half-space bound for the selection view of variance \(1+t\) controls wrong assignments. Using the corresponding true means as cell representatives bounds the squared error of the optimal centroids.
\end{proof}

Substitution in \cref{prop:centroid} gives the expected error of the fresh centroid fit conditional on this pilot. The pair gap holds if \(\Delta^2>16\sqrt{dx}+24x\). The additional scale \(u\) describes the precision of the estimated sites. \Cref{thm:main} treats arbitrary separations and weights by a different construction.

\subsection{Positive mixtures and isolated components}\label{sec:positive}

A local tensor fit also identifies a mixing law when its radial tail is controlled. The accuracy obtained for the mixing law is logarithmic in the moment error, which is the form of the minimax rate for Gaussian deconvolution in Wasserstein distance \cite{DM13}. The relevant dimension is the rank retained, and its approximation cost depends on the total discarded second moment. In this subsection \(m\) is the number of fitted moment orders.

\begin{proposition}[positive reconstruction]\label{prop:positive}
Suppose \(\int e^{|u|^2/(2s)}\,d\nu(u)\le B\), with \(s>0\), and \(\|\widehat M_j-M_j(\nu)\|_F\allowbreak\le\eta\sqrt{j!}\) for \(1\le j\le m\), where \(m\ge2\). For any rank-\(r\) orthogonal projection \(P\), \(r\ge1\), put \(D_P=\int|(I-P)u|^2\,d\nu(u)\). A finite positive probability measure \(\pi\) on \(\operatorname{ran}P\) can be computed from the projected moments such that

\[
\begin{aligned}
 W_1(\nu,\pi)&\le \sqrt{D_P}+C_{r,s,B}(m^{-1/2}+e^{2m}\eta),\\
 d_{\mathrm{TV}}(\nu*g_1,\pi*g_1)&\le \varphi(0)\sqrt{D_P}
     +C_{r,s,B}(e^{-c_sm}+e^{2m}\eta).
\end{aligned}
\]

Here \(W_1\) uses Euclidean cost, \(c_s>0\), and \(\pi\) uses at most \(\binom{m+r}{r}\) atoms. The search grid has at most \(\exp(O_s(rm+r\log(r+1)))\) points. Choosing \(m=\lfloor\frac14\log(1/\eta)\rfloor\ge2\) gives logarithmic accuracy for the mixing law when these moments are available.

\end{proposition}

\begin{proof} Write \(\nu_P=P_\#\nu\) in \(r\) orthonormal coordinates. Take \(R^2=4s(4+\max(1,s))m\) and \(h=2^{-\lceil 8m+R^2\rceil}\). Round each coordinate inside the radius-\(R\) ball toward zero on a grid of spacing \(h/\sqrt r\), and move its exterior to zero. Keep only lattice sites in the radius-\(R\) ball. The grid has at most \((1+2R\sqrt r/h)^r\) points. At each grid site write \(u=|x|^2/(2s)\) and \(b_x=2\sum_{l=0}^{\lceil 2u\rceil}u^l/l!\). Markov's inequality for a Poisson variable of mean \(u\) gives \(e^u\le b_x\le 2e^u\). Impose \(\sum_x p_x b_x\le2B\) on grid probabilities. For \(j\le m\), the normalized Frobenius moment error from rounding is at most \(\sqrt j\,h e^{R^2/2}\). The exterior contributes at most \(B e^{-R^2/(4s)}(2s)^{j/2}\), by maximizing \(t^je^{-t^2/(4s)}\) and using \(j!\ge(j/e)^j\). Both bounds hold for the rounded comparator, which is feasible because radii decrease.

Fit grid probabilities by minimizing the squared Euclidean residual of all projected moment tensors divided by \(\sqrt{j!}\), to additive squared-objective accuracy \(e^{-8m}\). The triangle inequality bounds the stacked difference from \(\nu_P\) by \(2\sqrt m\,\eta+O_B(m e^{-4m})\). The quadratic objective uses the Gram entries \(\sum_{j=1}^m(x\cdot y)^j/j!\) and contractions of \(\widehat M_j\) against \(x^{\otimes j}\), directly available from the train. For example, a conditional-gradient step minimizes a linear function over probabilities with one additional budget constraint. Each extreme point uses at most two grid sites, so enumeration suffices. The squared feature diameter is at most \(4e^{R^2}\). The step sizes \(2/(t+2)\) and the quadratic expansion give objective gap at most \(16e^{R^2}/(t+2)\) after \(t\ge1\) steps. To compress the output, form the matrix of monomials through degree \(m\), including the constant, on its support. While the support exceeds the row count \(\binom{m+r}{r}\), move the weights along a null vector, choosing its sign to decrease the budget weakly, until a weight reaches zero. All fitted moments are preserved.

Put \(T=\sqrt{m/(4e(1+s))}\) and let \(\Delta\) be the sup norm of the difference of the two characteristic functions on \(|t|\le T\). The real-argument Taylor remainder for either law is bounded by \(2B(es|t|^2/(m+1))^{(m+1)/2}\) by the exponential budget. Cauchy--Schwarz applied to the retained tensors gives

\[
\Delta\le C_B(2^{-m}+\sqrt m\,e^{T^2/2}\eta).
\]

Gaussian Fourier inversion bounds the smoothed density difference uniformly by \(C_r(\Delta+e^{-T^2/4})\). Integrate on the ball of radius \(\sqrt m\). Outside this ball, both Gaussian convolutions have mass at most \(2B e^{-m/(8s)}+2^{r/2}e^{-m/16}\) by splitting the radius between the two summands.

For \(W_1\), convolve both laws instead with \(T^r\psi(Tx)\), where \(\psi\) is the normalized squared modulus of the inverse Fourier transform of a smooth bump supported in the half-unit ball. The transport cost is \(O_r(1/T)\) and the density difference has sup norm at most \(C_rT^r\Delta\). A unit-Lipschitz test vanishing at zero contributes at most \(C_rm^{(r+1)/2}T^r\Delta\) inside the ball. Outside, its contribution is \(O_{r,s,B}(1/\sqrt m)\) by the uniform second moments. Absorbing polynomial factors gives the bounds for \(\nu_P\). Coupling \(u\) with \(Pu\) costs at most \(\sqrt{D_P}\) and Gaussian translation costs at most \(\varphi(0)\sqrt{D_P}\) in TV.
\end{proof}

Taking the leading \(r\) eigenvectors of the symmetrized \(\widehat M_2\) gives \(D_P\le\sum_{j>r}\lambda_j(M_2)+2\sqrt2\,r\eta\) by the variational principle. A cell containing at most \(k\) atoms with per-atom bound \(p_i e^{|u_i|^2/(2s)}\le K\) satisfies the radial budget with \(B=kK\).

\begin{corollary}[a list for isolated components]\label{cor:isolated}
Condition on \(M\) fixed, distinct pilot sites and assign a selector \(S\mid i\sim N(\mu_i,2I)\) to its nearest site. Enumerate the groups of sites within radius \(\rho\) of each pilot \(a\), including radius zero and all intersite distances. There are at most \(M^2\) distinct indexed groups. For component \(i\) write \(r_i=|a-\mu_i|\), \(r_j=\min_b|b-\mu_j|\) for \(j\ne i\), and \(D_{ij}=|\mu_i-\mu_j|\). Put

\[
\begin{aligned}
 h_i&=\min(1,M e^{-(\rho-2r_i)_+^2/16}),\\
 \beta_i&=\sum_{j\ne i}w_j\min(1,M e^{-(D_{ij}-\rho-r_i-r_j)_+^2/16}).
\end{aligned}
\]

The group probability \(p\) satisfies \(w_i(1-h_i)\le p\le w_i+\beta_i\). On \(n\) independent response pairs, estimate \(p\) by group frequency and \(\mu_i\) by coordinatewise response medians. With probability at least \(1-\alpha\), all frequencies differ from their group probabilities by at most \(\sqrt{\log(4M^2/\alpha)/(2n)}\). For any nonempty group with \(b\) responses put \(\xi=\beta_i/p\) and \(z=\sqrt{\log(4dM^2/\alpha)/(2b)}\). On the same event its location error is at most \(\sqrt{2d}\,\Phi^{-1}((1/2+z)/(1-\xi))\) whenever \(z+\xi<1/2\).

\end{corollary}

\begin{proof} Comparison with the home site in the Gaussian nearest-site inequality gives leakage at most \(h_i\) to sites outside the group, and incoming weight at most \(\beta_i\). Conditional on group membership the response law contains \(N(\mu_i,2I)\) with mass at least \(1-\xi\). Apply Hoeffding's inequality to frequencies and the Dvoretzky--Kiefer--Wolfowitz inequality to each conditional response coordinate, followed by a union bound. Inverting the two Gaussian CDF inequalities at the median gives the displayed radius. Pooling allows several pilots to share an isolated component.
\end{proof}

\subsection{Visible directions with a common unknown covariance}\label{sec:visible}

Let \(X=U+Z\), where \(U\) has at most \(k\) support points of positive mass and \(Z\) is an independent \(N(0,\Sigma)\) with \(\Sigma\succ0\). Write \(S=\operatorname{im}\operatorname{Cov}[U]\), \(r=\dim S\) and \(T=\operatorname{Cov}[X]\). After whitening, \(P=T^{-1/2}\operatorname{Cov}[U]T^{-1/2}\) has image \(T^{-1/2}S\), and the noise covariance is \(I-P\succ0\). Hence the eigenvalues of \(P\) on its image lie in \((0,1)\) and the orthogonal complement carries an independent standard Gaussian. This is a non-Gaussian component analysis problem \cite{TV18}. Higher cumulants identify the subspace before whitening.

\begin{theorem}[visible cumulant order]\label{thm:visible}
For \(3\le j\le m\), concatenate the first-mode unfoldings of the order-\(j\) joint cumulant tensors of \(X\) into a matrix \(A_m\). For \(k\ge2\), \(\operatorname{im}A_{2k}=S\). No smaller order suffices uniformly over \(k\)-atomic laws. Suppose \(\operatorname{rank} A_m=r>0\) and its smallest positive singular value is \(g\). Put \(D=\sum_{j=3}^m d^j\), \(L=\max(1,\max_i(\mathbb{E}[|X_i|^{2m}])^{1/(2m)})\), and let \(c_m\) be the sum of \((|\pi|-1)!\) over partitions \(\pi\) of \(\{1,\ldots,m\}\). For \(0<\eta<g/2\), with

\[
n\ge C m c_m^2 L^{2m}D\,\eta^{-2}\log(2D/\delta)
\]

observations one can estimate the projector onto \(S\) within \(2\eta/g\) in operator norm with probability \(1-\delta\), given \(r\) or the tolerance \(\eta\). For fixed \(m\) the arithmetic cost is polynomial in the sample size and \(d^m\).

\end{theorem}

\begin{proof} Cumulants add under independent convolution. All Gaussian cumulants of order at least three vanish, and all centered latent cumulants lie in tensor powers of \(S\). If \(a^\top A_{2k}=0\), the scalar \(a^\top U\) matches the first \(2k\) moments of a Gaussian with its mean and variance, by the moment-cumulant identity. The squared product of the distinct linear factors vanishing on its support has degree at most \(2k\). Its integral vanishes on the support and is positive for every Gaussian of positive variance. Thus \(a^\top U\) is constant, proving equality of the spaces. Conversely the positive \(k\)-node Gaussian quadrature law matches the standard normal through degree \(2k-1\). It is supported on the zeros of the monic Hermite polynomial of degree \(k\), whose squared Gaussian norm is \(k!\). Its cumulant of order \(2k\) is therefore \(-k!\). Embed the quadrature variate \(V\) as \(U=bVa\) for a unit vector \(a\) and \(0<b<1\), with independent noise covariance \(I-b^2aa^\top\). Its observed moments through degree \(2k-1\) agree with \(N(0,I_d)\), which has latent span zero.

For each cumulant entry use independent groups of \(m\) observations. Within a group estimate each term of the moment-cumulant formula using a different observation for each block of the partition. The product is unbiased. By independence across blocks and Hölder's inequality within each block its second-moment norm is at most \(L^m\). The triangle inequality bounds the second-moment norm of the signed sum by \(c_m L^m\). Divide the groups into \(O(\log(2D/\delta))\) batches and take the coordinatewise median of their averages. Chebyshev's inequality and a binomial tail bound give entrywise error at most \(\eta/\sqrt D\) simultaneously, hence matrix error at most \(\eta\). Singular values above \(\eta\) recover \(r\) when \(g>2\eta\). For the leading left projector \(\widehat P\) we have \(\|(I-\widehat P)A_m\|_{\mathrm{op}}\le2\eta\). Applying a right inverse on the image bounds the sine of the largest principal angle by \(2\eta/g\).
\end{proof}

This reduction identifies the directions that carry mixing information, and the unknown covariance remains in the reduced problem. Both the gap \(g\) and the moment bound \(L\) enter the sample requirement, so a light, remote mean makes high cumulants costly. Gaussian quadrature is also the source of statistical-query lower bounds \cite{DKS17}, and a quasi-polynomial obstruction holds there even with uniform weights \cite{DKKLP25}. List-decodable covariance estimation \cite{DKLPP23} applies to differences of observations, with relative Frobenius error polynomial in the inverse inlier fraction \(k\).

\section{Numerical illustrations}\label{sec:numerics}

This section illustrates the three steps of the argument at constants smaller than those of \cref{rem:exponent}. \Cref{sec:num-taper} shows the criticality barrier and the approximation error of the tapers as a function of degree. \Cref{sec:num-cover} runs the cover of \cref{sec:cover} at small scale and tests \cref{prop:profile}. \Cref{sec:num-recovery} evaluates the recovery rate of \cref{thm:recovery,thm:line}. \Cref{app:baseline} records experiments with a baseline variant of the learner and diagnostics for \cref{sec:complements}. The random experiments use base seed \(42\).

\subsection{Leakage and cancellation}\label{sec:num-taper}

In \cref{fig:leak} the component mean is the origin and the two anchors are \(2e_1\) and \(De_2\) in dimension 30. The exact probability of assignment to the foreign anchor is

\[
\Phi\!\left(-\frac{D^2-4}{2\sqrt{2(D^2+4)}}\right).
\]

For \(D=4,6,8,10\), independent projected samples gave counts \(258160\), \(55186\), \(7594\) and \(643\) in \(1.5\times10^6\) trials per distance. The exact probabilities are \(0.17139\), \(0.03682\), \(0.005049\) and \(0.000437\). The four counts lie \(2.33\), \(0.19\), \(0.24\) and \(0.49\) standard errors from their expectations. The bars in \cref{fig:leak} show this first run, so the interval at \(D=4\) excludes the exact value. A second run of the same size gave \(257188\), \(55464\), \(7639\) and \(667\), within \(0.22\), \(1.02\), \(0.76\) and \(0.45\) standard errors, and the two runs pooled lie \(1.80\) standard errors from the exact value at \(D=4\). The lower bound applies since a Gaussian centered on the foreign anchor enters its own half-space with probability at least \(1/2\). \Cref{fig:energy} gives the squared coefficients for the two-atom construction of \cref{thm:criticality}. With \(s=1.5\) and \(N=80\), the displaced atom weighs \(e^{-40}\) and the peaks occur at degrees \(119\) and \(120\).

\Cref{fig:tapers} fixes the mixing law at each leak variance so that eventual convergence of ordinary truncation is visible. Five atoms occupy \(-2,-1,0,1,2\) and the other offsets are \((-1)^j(2+12j/25)\) for \(1\le j\le25\). We normalized the weights \(\exp[-(|\delta|-2)_+^2/(2v')]\) and evaluated the mean absolute density ratio error at the 40 midpoint quantiles of a standard Gaussian conditioned on \([-3,3]\), using 140-digit arithmetic. At degree \(160\) the base-ten logarithms of the truncation errors were \(18.56\), \(31.46\) and \(35.28\) for \(v'=1.5,4,8\). The original taper gave \(-18.43\), \(-7.19\) and \(-4.12\), and the rational taper \eqref{eq:rational} gave \(-2.30\), \(-0.92\) and \(-1.11\). At degree \(800\) ordinary truncation was accurate for all three fixed laws. Uniform failure in \cref{thm:criticality} requires a sequence of laws indexed by degree.

\textbf{Damping of low levels.} The taper used in \cref{thm:main} is the least accurate of the three at these degrees, and the weights explain why. Both tapers multiply level \(n\) by a weight that tends to one only as the degree grows. At degree 16 the original taper has \((\Lambda_1,\Lambda_2,\Lambda_4)=(0.995,0.983,0.894)\) for \(v'=1.5\) and \((0.824,0.610,0.221)\) for \(v'=8\). The low levels are damped before the series has converged. This is why the oracle error on the ball D1 in \cref{app:fifty} rises from \(0.0127\) to \(0.0687\) as \(v'\) goes from \(1.5\) to \(8\) and exceeds both the degree-zero error and the hard cutoff there. The rational taper at degree 160 has \(\Lambda^{(p)}_1=0.983\) for \(p=5\) (\(v'=1.5\)) and \(0.777\) for \(p=7\). The latter reaches \(0.990\) only near degree 800. Since \(v'=4\) and \(v'=8\) both give \(p=7\), their rational weights coincide, and the two curves in \cref{fig:tapers}(c) differ only through the mixing law. As a rule, level \(n\) is undamped once \(A\gtrsim n(1+v')\) for the original taper and \(A\gtrsim n2^{p}\) for the rational one. The better dependence of \eqref{eq:rational} on \(v'\) thus comes with a larger degree threshold, which is the factor \(2^{p+1}\) in \cref{rem:exponent}.

The distinction persists under the integrated density loss. At \(N=16\) and \(B=2\), the Fourier witnesses in \cref{thm:criticality} were \(0.000220\), \(0.0701\) and \(20.90\) for \(s=1/2,1,2\). Integration of the absolute density error gave \(0.000283\), \(0.0899\) and \(26.74\). At criticality the witness decreased to \(0.0352\) at degree \(64\) and \(0.0176\) at degree \(256\).

\begin{figure}
\centering
\includegraphics[width=0.67\textwidth]{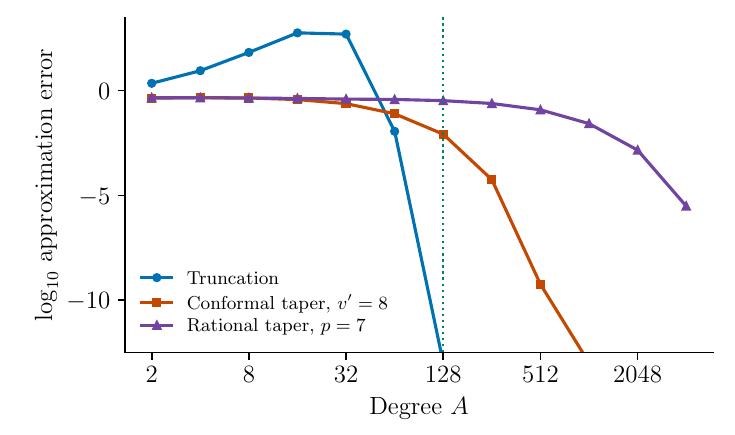}
\caption{Approximation error against degree for a cell with four components, from exact moments. The dotted line marks degree \(2^p=128\). Values below \(10^{-13}\) are shown at that level.}\label{fig:degree}
\end{figure}

\textbf{A cell with several components.} \Cref{fig:degree} shows the same effect for a cell law of the kind the cover produces: mass \(0.5\) at the anchor and masses \(0.25,0.15,0.10\) at displacements \(3.8,-4.5,5.0\) on the line. The error is half the \(L^1(\gamma)\) distance between \(r_\nu\) and its degree-\(A\) reconstruction from exact moments. Truncation first diverges, to \(10^{2.8}\) at degree 16, and converges for this fixed law once the degree exceeds the largest squared displacement. The conformal taper at \(v'=8\) reaches \(10^{-2}\) at degree 128. The rational taper at \(p=7\) stays above \(10^{-0.5}\) through degree 128 and reaches \(10^{-1.6}\), \(10^{-2.8}\) and \(10^{-5.5}\) at degrees 1024, 2048 and 4096. Resolving a cell of several components with the taper of \cref{thm:main} therefore takes a degree in the thousands, and by \cref{thm:levels} the sample cost is exponential in the degree, which is the source of the exponent in \cref{rem:exponent}.

\textbf{Perturbing the moments.} On the same 40 points take \(Y=U+\varsigma Z\) with the \cref{fig:tapers} displacement law and \(\varsigma^2=2\). For each observation point \(h\) the population score is \(\mathbb{E}[P_h(U)]\), with

\[
P_h(u)=\sum_{j=0}^A\lambda_j \operatorname{He}_j(h)u^j/j!.
\]

Each fit uses the same response batch. Moment \(j\) is estimated by the average of \(\varsigma^j\operatorname{He}_j(Y/\varsigma)\). \Cref{prop:variance} gives the score variance including correlations between orders.

For \(v'=4\) and \(n=4096\), the root mean square sampling errors of the original taper at degrees 8 and 16 were \(0.1513\) and \(0.8829\), with population biases of \(0.09151\) and \(0.02324\) in the same norm on the grid. This is the bias--variance trade in the degree: adding squared bias and variance, degree 16 first beats degree 8 at \(n=395{,}598\). At degree 16 the rational taper reduced the sampling error to \(0.01815\) with bias \(0.4720\), while the hard cutoff had sampling error \(2.41\times10^6\). These errors precede clipping and normalization.

\subsection{The weight-adaptive cover at small scale}\label{sec:num-cover}

This run uses the cover of \cref{alg:cover} itself in \(d=2\). The resolution \(S_0=\sqrt{17n}\), the views, the gate \(\rho_g\), the complex statistic \eqref{eq:pair-statistic} with its threshold, the tier schedule with \(\delta_c=1/4\) and \(U=5\), the reference counts and the keep rule are those of \cref{sec:cover}. We relax the group size \(m_0\), train on \(2\times10^5\) pairs per level, retain eigenvalues above \(10^{-3}\) times the largest, clip level \(j\) at \(2\cdot5^{j/2}\), and report the number of kept references where \cref{alg:tier} would abort. Each row of \cref{tab:endtoend} has 20 independent runs.

\emph{Fixed design.} Here \(k=4\), the weights are \((0.5,0.3,0.15,0.05)\) and the means \((0,0)\), \((1.5,0)\), \((0,8)\), \((30,0)\). Two heavy means are not separated and the lightest is remote. The degree \(n\in\{2,3,8\}\) is the same at every tier and \(m_0=1024\). The degree prescribed in \cref{sec:tier} is \(31\) at the first tier and \(43\) at the last.

\emph{Ladder design.} Here \(k=5\) and the weights are \((0.5,0.25,0.13,0.07,0.05)\). Tier \(u\) uses degree \(n_u=6+2u\), so the resolution grows with the tier as in \cref{thm:cover}, and \(m_0=512\). The heavy mean is the origin. The mean of weight \(w_i\) lies at distance \(\rho_g(n_{u_i})-4.5\) from it, where \(u_i\) is the first tier with \(2^{-u_i}\le w_i\). These distances are \(3.84\), \(4.45\), \(5.00\) and \(5.52\), in four orthogonal directions. A light component then sits inside the gate of the heavy one at its own tier and is not resolved by the pair test.

\begin{table}
\centering\small
\setlength{\tabcolsep}{3pt}
\begin{tabular*}{\linewidth}{@{\extracolsep{\fill}}lrrrrrrrr@{}}
\toprule
& & \multicolumn{2}{c}{Pair-test error} & \multicolumn{2}{c}{Anchors kept} & & \multicolumn{2}{c}{TV error} \\
\cmidrule(lr){3-4}\cmidrule(lr){5-6}\cmidrule(l){8-9}
Design & \(S_0\) & same far & separated near & tier 0 & tier 5 & \(\max_i r_i\) & taper & degree 0 \\
\midrule
Fixed, \(n=2\) & 5.83 & 0.714 & \(8\times10^{-7}\) & 3.1 & 39.6 & 0.26 & 0.058 & 0.071 \\
Fixed, \(n=3\) & 7.14 & 0.350 & \(2\times10^{-5}\) & 2.5 & 17.3 & 0.48 & 0.078 & 0.081 \\
Fixed, \(n=8\) & 11.66 & 0.0016 & 0 & 1.8 & 3.6 & 3.73 & 0.215 & 0.238 \\
Ladder & 10.1--16.5 & \(2\times10^{-5}\) & -- & 2.4 & 1.7 & 5.44 & 0.358 & 0.375 \\
\bottomrule
\end{tabular*}
\caption{The cover of \cref{alg:cover} at relaxed panel sizes, means over 20 runs. Error rates are the fractions of same-component pairs declared far and of pairs with \(|\Delta|\ge S_0\) declared near. The ladder design has no pair with \(|\Delta|\ge S_0\). The radius column is the largest over components and runs. Standard errors of the TV columns are at most \(0.012\).}\label{tab:endtoend}
\end{table}

\textbf{The pair test.} At \(n=2\) and \(n=3\) the test declares \(71\%\) and \(35\%\) of same-component pairs far, so the tiers keep duplicates. A tier kept more than \(k\) references in 104 and 91 of the 120 tiers, and the bound \((32k^2E_0+1)e^{-n}\) of \cref{lem:pair-test} exceeds one at these degrees. At \(n=8\) the rate is \(0.16\%\), and in the ladder design, with degrees \(6\) to \(16\), it is \(2\times10^{-5}\). In both, no tier exceeded \(k\) and the covers had 11 to 19 anchors against the limits \(k(U+1)=24\) and \(30\). The error of the test decays with the degree as \cref{lem:pair-test} states.

\begin{figure}
\centering
\includegraphics{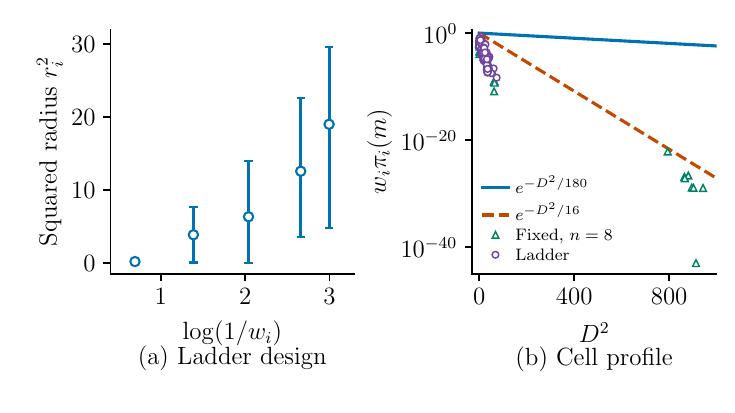}
\caption{The cover at small scale. Panel (a) gives the squared radius of each component in the ladder design against \(\log(1/w_i)\): mean over 20 runs, with bars from the smallest to the largest. Panel (b) gives the joint assignment mass \(w_i\pi_i(m)\) against squared distance to the anchor in the first run of two designs. The solid line is the envelope of \eqref{eq:constant-leak} with prefactor one. The dashed line is \cref{prop:cells}(3) at zero radius.}\label{fig:cover}
\end{figure}

\textbf{Radii grow as weights shrink.} In the fixed design every component lay within \(3.73\) of an anchor. In the ladder design the mean radii of the five components, in decreasing order of weight, were \(0.40\), \(1.83\), \(2.27\), \(3.47\) and \(4.25\) (\cref{fig:cover}(a)). The cover locates the heavy component precisely and leaves the light ones at a distance that increases with \(\log(1/w_i)\), which is the behaviour \cref{thm:cover} permits. All radii are below the bounds \((2+1/32)S_0\ge11.8\) of \cref{lem:tier} and \(\sqrt{164\log(Q_c^{12}/w_i)}\ge91\) of \cref{thm:cover}.

\textbf{The profile.} We computed every \(\pi_i(m)\) by quadrature on the selection view. \Cref{fig:cover}(b) plots \(w_i\pi_i(m)\) against \(D^2=|\mu_i-a_m|^2\). Over all 80 runs, components and anchors, the largest value of \(w_i\pi_i(m)e^{D^2/180}\) was \(0.25\), against the prefactor \(Q_c^{12}\ge2^{72}\) allowed by \eqref{eq:constant-leak}. In the ladder design the unresolved light components appear as the points of mass \(10^{-3}\) to \(0.2\) at \(D^2\) between 5 and 26. They lie below the envelope because their weight is small, which is the mechanism of \cref{prop:profile}.

\textbf{The fit.} On \(2\times10^6\) fresh observations we fitted each kept cell by the unclipped train of \cref{sec:trains} at degree \(A=4\), with the rational taper at \(p=7\), frozen residuals, at most 64 auxiliary vectors and scores clipped at 40. The weights are \((\Lambda^{(7)}_1,\ldots,\Lambda^{(7)}_4)=(0.178,0.029,0.0020,0.00024)\), so the fit is close to its degree-zero part, as the last two columns show. TV errors are midpoint rules on a mesh of \(0.05\). EM with \(2\times10^5\) observations had error \(0.0025\) on the fixed design and \(0.0028\) on the ladder, and the centroid fit of \cref{prop:centroid} on a \(k\)-cell pilot had \(0.083\) and \(0.206\). The fit is best at \(n=2\), where the many duplicate anchors act as a fine net, and worst on the ladder, where the cells near the origin contain several components. Resolving such a cell is the task of the taper at the degrees of \cref{fig:degree}.

\subsection{Moment recovery}\label{sec:num-recovery}

\textbf{A one-dimensional dual calculation.} At \(s=B=2\) and \(N=15,31,63,99,100\) in this order, let \(m=N+1\), \(t=\sqrt{m/2}\), and take \(h(x)=\cos(tx)\) for even \(m\) and \(h(x)=\sin(tx)\) for odd \(m\). Define the weighted approximation error

\[
D_N=\inf_{\deg p\le N}\sup_{u\in\mathbb R} e^{-u^2/4}|e^{-t^2/2}h(u)-p(u)|.
\]

Gaussian convolution multiplies this test by \(e^{-t^2/2}\). Alternating weighted residuals on a mesh of spacing \(1/80\) give nodes \(\pm x_i\), and barycentric weights on them define a signed measure \(\sigma\) that annihilates all polynomials through degree \(N\). A lower bound for \(D_N\) is the ratio

\[
\Big|\int e^{-t^2/2}h\,d\sigma\Big|\Big/\int e^{u^2/4}\,d|\sigma| .
\] An alternating polynomial on the same nodes, with its weighted residual enclosed on the whole line, gives an upper bound. The upper-to-lower ratios are below \(1.0041\) at all five degrees.

\begin{figure}
\centering
\includegraphics{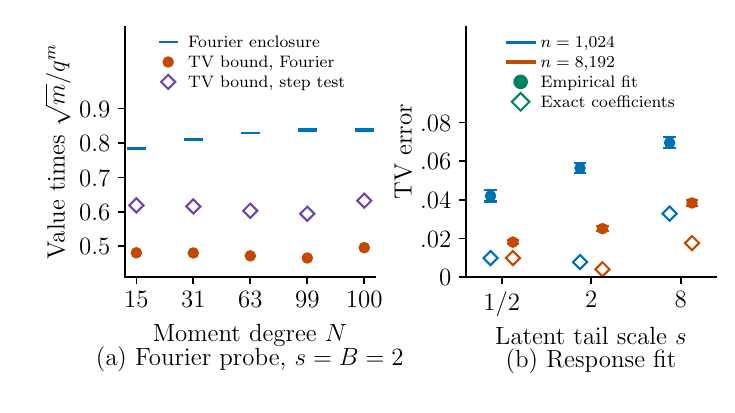}
\caption{In (a), the weighted approximation value for a Fourier test and two TV lower bounds from its alternating witness, all multiplied by \(\sqrt{m}/q^m\) with \(m=N+1\). The endpoints of each Fourier enclosure nearly coincide. Panel (b) gives mean TV errors from 48 response fits. Bars show one Monte Carlo standard error. Diamonds give errors at the same degrees using exact coefficients.}\label{fig:recovery}
\end{figure}

Completing \(a\sigma_+\) and \(a\sigma_-\) to probabilities by a common atom at zero, with the largest multiplier \(a\) the budget allows, gives a pair with identical input moments and the bound \(L_N=a|\int e^{-t^2/2}h\,d\sigma|/4\le E_{N,1}(2,2)\). At the five degrees \(\sqrt m L_N/q^m\) ranges from 0.465 to 0.496 (\cref{fig:recovery}). The convolved witness has exactly \(m\) sign changes, and a test that alternates at them, integrated on a mesh of spacing \(1/320\), gives the larger lower bounds \(0.6183,0.6153,0.6026,0.5939,0.6321\) on \(\sqrt m\,E_{N,1}(2,2)/q^m\). For comparison, the derivative witness in the proof of \cref{thm:recovery} gives \(0.175\) at \(m=64\). The upper bound is \cref{thm:line}.

\textbf{Noisier responses at fixed dimension.} For \(s\in\{1/2,2,8\}\) let the mixing law put equal mass at \(\pm\sqrt{2s\log 2}\), so the exponential budget equals two. In 48 fits at each \(n\in\{1024,8192\}\), the estimator of \cref{cor:responses} used variance-two responses and the prescribed degree. \Cref{fig:recovery}(b) shows TV error declining at all three tail scales as \(n\) increases.

\section{Conclusion}\label{sec:conclusion}

This paper separates the information carried by the moments of a Gaussian location mixture from the cost of using it. For information, optimal recovery from moments through degree \(N\) converges geometrically with base \(\sqrt{s/(s+2)}\) where Hermite truncation diverges, and parity extrapolation gives the exact order on the line. For computation, a cover whose radius depends on the component weight and a conformal taper give polynomial complexity of density estimation for arbitrary means and weights, in the real-number and bit models.

The exponent of \cref{thm:main} exceeds \(10^{11}\). It comes from the taper degree at a constant leak variance (\cref{fig:degree}) and from the variance of product estimates of each Hermite level (\cref{rem:exponent}). Lowering it is the main open problem, and at feasible degrees EM is the more accurate method (\cref{app:baseline}). Three further questions are open: whether \(\sqrt m\,E_{N,1}(s,B)/q^m\) converges as \(m=N+1\to\infty\), which we conjecture; the sharp base when the dimension grows in proportion to the degree; and a proper learner with at most \(k\) components at joint polynomial cost.

\appendix

\section{Integer choices for the cover}\label{app:integers}

\begin{lemma}\label{lem:degree}
Let \(Q_c=2dk/\delta_c\ge8\). For \(w_i\ge2^{-U}\) let \(u\) be the first tier at which \(2^{-u}\le w_i\). Its degree satisfies

\[
n_u\le 1+\frac{26\ln Q_c+(9/4)\ln(1/w_i)}{\ln(8/3)} .
\]

\end{lemma}

\begin{proof} Set \(t=1+u\) and \(\alpha=\delta_c/(2t^2)\). Since \(k/\delta_c\le Q_c/2\) and \(\lg x\le1+\log_2x\), one has \(\lg(8k/\alpha)\le4Q_ct\) and \(N_r\le4Q_ct2^u\). Also

\[
\lg(8N_r/\alpha)\le7+2\log_2 Q_c+3\log_2 t+u\le8Q_ct .
\]

These inequalities give \(m_0\le2^{20}Q_ct\), \(N_g\le2^{24}Q_ct2^u\) and \(P\le2^{27}Q_c^2t^2 2^{2u}\). Using \(E\le2Q_c\), \(k\le Q_c\) and \(1/\alpha\le Q_ct^2\), we obtain \(512k^2EP/\alpha\le2^{37}Q_c^6t^4 2^{2u}\). The elementary bound \(t\le32\cdot2^{u/16}\) follows by maximizing \((1+u)2^{-u/16}\) for \(u\ge0\). Next \(2^{57}\le Q_c^{19}\) and \(2^u\le2/w_i\), so the degree criterion is met when \((8/3)^n\ge Q_c^{26}w_i^{-9/4}\). Rounding up proves the degree bound. Since \(\ln Q_c\ge\ln8\), multiplying it by \(17(65/32)^2\) gives at most \(164(12\ln Q_c+\ln(1/w_i))\).
\end{proof}

\section{Baseline experiments and further diagnostics}\label{app:baseline}

The experiments of this appendix use a baseline variant of the learner with its own pair test and pruning rule. They measure that variant and are separate from the cover of \cref{sec:cover}.

\textbf{The baseline variant.} The baseline uses the original taper and a product statistic. It thins each observation into \(J=9\) parts of variance \(9\) and uses degree \(n=4\). The test view of an observation is the mean of four parts, of variance \(9/4\), and its held-out view is the ninth part. Two observations are declared near when their test views differ by at most the gate \(\rho_b=\tfrac12\sqrt{18}(\sqrt d+\sqrt8)\), which is \(21.0\) for \(d=50\) and \(100.87\) for \(d=2000\). We write \(\rho_c=2\rho_b\) for the resolution of this gate. Where a tensor statistic is used (\cref{app:moments,app:highdim}), it is a product of independent parts at the single scale \(S_0=32\sqrt{18}\simeq135.8\), thirty-two standard deviations of a part difference, in place of the complex statistic \eqref{eq:pair-statistic} at \(S_0=\sqrt{17n}\). A reference needs \(m_0=50\) near neighbours, and a candidate anchor is pruned when it lies within \(4S_0\simeq543\) of an anchor already kept in its tier. Fits use the original taper of \cref{def:taper}. Every design in \cref{app:chains,app:fifty} has diameter less than \(4S_0\), so pruning leaves one anchor per tier there and the norm gate is the entire pair test. E1 through E3 are therefore single-anchor fits per tier, and their errors measure this pruning rule and the low fitted degree.

\subsection{Estimating moments and finding anchors}\label{app:moments}

We used the three means \(8e_1\), \(8e_2\) and \((5,5,2,0,\dots,0)\) in dimension eight with weights \((0.5,0.3,0.2)\). With six parts of variance \(12\) and a threshold of \(0.003\) times the largest eigenvalue, batches of \(20{,}000\), \(200{,}000\) and \(2{,}000{,}000\) produced ranks \((5,12,25)\), \((3,3,3)\) and \((3,3,3)\). The rank cap of \cref{alg:tt} rejects the first fit. At \(2{,}000{,}000\) the relative Frobenius errors of orders one to three were \(0.00098\), \(0.00178\) and \(0.00282\).

For the baseline independent-product pair test, take \(d=30\), \(n=4\), \(S_0=32\sqrt{18}\) and means at radii \(0\), \(1.05S_0\), \(1.3S_0\) and \(2000S_0\), with independent uniform directions and weights \((0.35,0.30,0.30,0.05)\). Each order used \(300{,}000\) training pairs at scale \(S_0\), clipping radius \(2\cdot5^{j/2}\) and relative eigenvalue threshold \(10^{-3}\). The clipped ranks were \((3,4,4,4)\). Among \(60{,}000\) fresh pairs, the tensor statistic rejected none of \(18{,}390\) same-component pairs and accepted none of \(35{,}819\) pairs with separation in \([S_0,2S_0]\). Without clipping the ranks were \((1,1,1,1)\), and the test accepted \(12{,}439\) of \(36{,}097\) matched far pairs, so clipping is needed when one remote component dominates the moments.

\subsection{From local scores to density loss}\label{app:chains}

\Cref{tab:protocol} specifies the information and degree used in each fit. \Cref{tab:chain-oracle} isolates approximation with oracle information. E1 through E3 use practical cover schedules. Each cutoff uses the corresponding taper's moments. We clip scores to \([0,40]\), multiply by the anchor Gaussians and normalize their sum. We fix the fitted degree because a remote atom of conditional weight \(p\) can still contribute \(p|\delta|^{2n}\) to the directional second moment in \cref{lem:parts}.

The fitting batch size is \(N'\). The gate \(\rho_b\), its resolution \(\rho_c=2\rho_b=\sqrt{18}(\sqrt d+\sqrt8)\) and the pruning distance \(4S_0\) are those of the baseline variant.

For the chain design D3, take means \(ie_1\) for \(1\le i\le\lfloor8\rho_c\rfloor\) with equal weights. For D4, assign weights proportional to \(2^{-i/40}\) with total mass \(1-(N')^{-1/2}\) and put the remaining mass at distance \(4\rho_c\) beyond the last mean. The oracle experiment reduces exactly to one dimension. \Cref{tab:chain-oracle} gives the degree-16 errors at $N'=10^5$.

\begin{table}
\centering\small
\setlength{\tabcolsep}{3pt}
\begin{tabular*}{\linewidth}{@{\extracolsep{\fill}}lrrrr@{}}
\toprule
& \multicolumn{2}{c}{Uniform chain (D3)} & \multicolumn{2}{c}{Light component (D4)} \\
\cmidrule(lr){2-3}\cmidrule(l){4-5}
Radius $R_0$ & Taper & Hard cutoff & Taper & Hard cutoff \\
\midrule
0 & 0.00056 & 0.09835 & 0.00137 & 0.10764 \\
2 & 0.00953 & 0.49941 & 0.01015 & 0.50407 \\
4 & 0.05138 & 0.54419 & 0.04865 & 0.55698 \\
8 & 0.22091 & 0.72714 & 0.22154 & 0.73073 \\
\bottomrule
\end{tabular*}
\caption{Total variation error at degree 16 for the chain designs with oracle anchors, $N'=10^5$ and taper parameter $v'=8$.}\label{tab:chain-oracle}
\end{table}

The degree-zero D3 fit at \(R_0=0\) already has error \(0.00051\), since an evenly spaced chain with exact anchors has nearly uniform cell frequencies.

For a sampled chain we used the one-dimensional design, where the conditional tensors are scalars. Each tier used 32 references and 4,096 neighbours, with \(m_0=50\) and dyadic thresholds through the first tier with \(w\le100/N'\). Each Gram, moment and frequency estimate used \(N'\) fresh observations. We integrated densities on a mesh of width \(1/32\), and halving the mesh changed losses by at most \(0.0020\).

\begin{figure}
\centering
\includegraphics{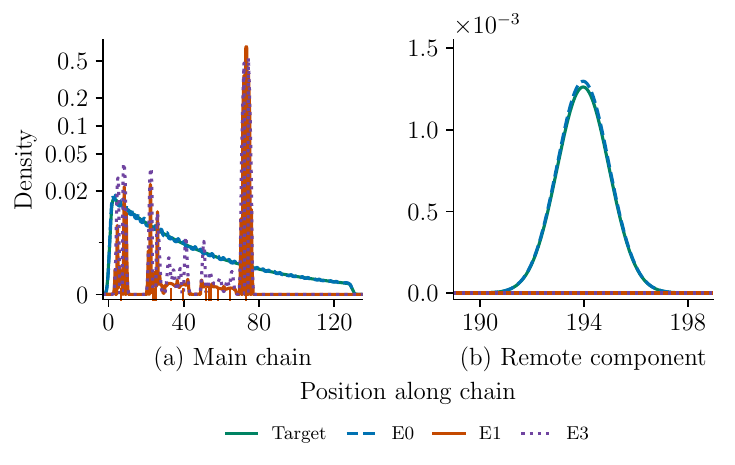}
\caption{D4 at \(N'=10^5\), \(v'=4\) in the first replicate. Ticks in (a) mark the learned anchors. The vertical scale there is linear up to density \(0.02\) and logarithmic above it. Panel (b) resolves the rare component, covered by E0 and missed by E1 and E3.}\label{fig:chain}
\end{figure}

At \(N'=10^5\) each of 50 fits per design returned 11 anchors, one per tier. With \(v'=4\) and oracle radius \(R_0=2\), E0 through E3 had mean losses \((0.016,0.866,0.866,0.730)\) on D3 and \((0.016,0.872,0.872,0.726)\) on D4. Standard errors were at most \(0.009\) for E1 and E2 and \(0.008\) for E3. On D4 the remote atom lay \(68.2\) to \(136.1\) from its closest anchor, beyond \(\rho_c\simeq16.2\). \Cref{fig:chain} shows the main chain and the missed remote component in the first D4 fit. Eleven anchors cannot cover a chain of 129 unit-spaced means at radius \(2\), and the gap between E0 and E1 is the cost of that pruning rule.

\textbf{Parametric comparison.} On \(10^5\) observations per chain, expectation maximization fitted 129 D3 and 130 D4 locations with free weights and unit variance. At 50 updates from 20 starts the highest-likelihood fits had TV errors \(0.01197\) and \(0.01876\). EM is far more accurate than the baseline fits on these designs.

\subsection{A fifty-dimensional comparison}\label{app:fifty}

Both designs use equal weights. In D1 we fixed one draw of ten independent uniform means in the ball of radius \(\sqrt{\log 10}\) in \(d=50\). In D2 the twenty means are \(D e_i\) with \(D=10.5\) or \(168\) and separations \(\sqrt2D\). E0 uses the true means as anchors.

E1 through E3 used one cover tier with 32 references and 4,096 neighbours. Here \(\rho_b=21.0\), and pruning at \(4S_0\simeq543\) leaves a single anchor in each design. Each subspace step used 4,096 observations and each coordinate step used 100,000. We kept at most \(k\) directions with empirical eigenvalues above \(10^{-3}\) times the largest.

\begin{table}
\centering\small
\setlength{\tabcolsep}{2pt}
\begin{tabular*}{\linewidth}{@{\extracolsep{\fill}}lrrrrrrrr@{}}
\toprule
& \multicolumn{3}{c}{E0 (oracle)} & \multicolumn{2}{c}{Partly oracle} & \multicolumn{3}{c}{Learned} \\
\cmidrule(lr){2-4}\cmidrule(lr){5-6}\cmidrule(l){7-9}
Design & Degree 0 & Taper & Cutoff & E1 & E2 & E3 & Centroid & EM \\
\midrule
D1 & 0.00681 & 0.0687 & 0.0119 & 0.0874 & 0.2196 & 0.2320 & 0.1922 & 0.0753 \\
D2, \(D=10.5\) & \(0\) & 0.00047 & 0.5014 & \(>0.999\) & \(>0.999\) & \(>0.999\) & 0.0603 & 0.0862 \\
D2, \(D=168\) & \(0\) & \(<10^{-1000}\) & \(<10^{-1000}\) & 0.9754 & 0.9754 & 0.9827 & 0.0565 & 0.0400 \\
\bottomrule
\end{tabular*}
\caption{Total variation error for the ball (D1) and the two simplices (D2). The centroid fit uses one thinning sign at $t=1$. Oracle degree-zero entries for the two simplices are exact.}\label{tab:ball-simplex}
\end{table}

\Cref{tab:ball-simplex} gives TV errors for normalized densities, with \(R_0=0\) in E0 and \(v'=8\) for the tapers. On each simplex permutation symmetry gives oracle cell probabilities \(1/20\), so the degree-zero density equals the target. At \(D=10.5\) adding higher orders with the taper costs \(0.00047\), compared with \(0.5014\) for the hard cutoff. For \(D=168\) the bounds are analytic, since the foreign mass in a simplex cell is at most \(19e^{-D^2/8}\).

TV errors were estimated by stratified importance sampling with equal numbers of target and proposal draws. E1--E3 used 50 fits and 8,192 evaluation points per fit, with standard errors below \(0.00089\). The oracle standard errors were at most \(0.0011\). Identity-covariance EM used 20 distance-weighted starts, 25 updates on 4,096 draws, then ten on 100,000 fresh draws for the highest-likelihood start. Its standard errors were \((0.0013,0.0069,0.00012)\).

On the near simplex the E0 errors at \(v'=1.5,4,8\) were \(0.5519,0.0274,0.00047\). On D1 they were \(0.0127,0.0330,0.0687\), increasing with \(v'\) for the reason given in \cref{sec:num-taper}. In 50 independent cover runs per design the pruning rule always retained one of 32 candidates. The mean maximum distance from a component to that anchor was \(1.575\), \(10.677\) and \(237.884\). The learned errors in \cref{tab:ball-simplex} are those of a single anchor, and the centroid fit and EM are far more accurate on both simplices.

\Cref{prop:rejection} explains the near-simplex errors. Any single anchor and any acceptance rule with rate at least one percent have total variation error above \(0.9495\), by a case distinction on the distance \(6.2\) to the nearest mean. At the population barycenter the bound is \(0.999989\).

We learned \(k\) Voronoi sites from 4,096 observations by \(k\)-means++ with eight starts and at most 25 Lloyd updates. On 100,000 fresh observations, \(X+g\) selected the cell and \(X-g\) estimated its centroid, with empirical cell weights. In order D1, near simplex, far simplex, the single-anchor Gaussian gave \((0.2370,{>}0.999,0.9526)\) over 50 fits per design, and refitting all \(k\) centroids gave \((0.1922,0.0603,0.0565)\) with standard errors \((0.00023,0.00271,0.00016)\). On the two near-simplex pilots where two true means shared a nearest site, mean error was \(0.1529\), compared with \(0.0564\) in the other 48.

On the first D1 pilot, 49,152 independent selection draws per atom gave \(V=1.9616\) (standard error \(0.0006\)), hence \(\varphi(1)V\simeq0.4746\). The estimated population-centroid mixture had TV error \(0.1907\), so the trace bound of \cref{prop:centroid} exceeds the error by a factor \(2.5\) on this dispersed cell law. For the 48 near-simplex pilots with distinct nearest sites and all 50 far-simplex pilots, \cref{prop:centroid} gave a conditional expected-error bound below \(0.0634\) at \(t=1\).

\begin{table}
\centering\small
\setlength{\tabcolsep}{2pt}
\begin{tabular*}{\linewidth}{@{\extracolsep{\fill}}>{\raggedright\arraybackslash}p{.155\linewidth}>{\raggedright\arraybackslash}p{.235\linewidth}>{\raggedright\arraybackslash}p{.355\linewidth}>{\raggedright\arraybackslash}p{.18\linewidth}@{}}
\toprule
Protocol & Centers or partition & Cell masses and moments & Degree or fit \\
\midrule
E0 & Greedy net of true means & Population cell law, integrated numerically on D1 & Zero or prescribed degree \\
E1 & Sample cover & Population cell law given the cover & 16 \\
E2 & Same sample cover & Empirical weights and moments through 2. Oracle orders 3 to 16 & 16 \\
E3 & Same sample cover & Empirical weights and moments through 2 & 2 \\
Centroid & Pilot \(k\)-means cells & Empirical masses and response centroids & Gaussian per cell \\
EM & Raw-data initializations & Likelihood updates of means and weights & \(k\) Gaussians \\
Small-scale run & Cover of \cref{alg:cover} with relaxed panels & Fresh batches for retained cells, orders 1 to 4 & 4, rational taper \\
Full learner & Weight-adaptive sample cover & Fresh batches for retained cells and every required order & Profile-dependent degree \\
\bottomrule
\end{tabular*}
\caption{Information supplied to each fit. E1 through E3 use the baseline cover and clip the scores at 40. The small-scale run is that of \cref{sec:num-cover}. The last row is the procedure of \cref{thm:main}.}\label{tab:protocol}
\end{table}

\textbf{Choosing a partition from response views.} On the same 50 pilots per design, we compared one cell, the eight-start fit and Lloyd iteration seeded by farthest-first traversal from the pilot point nearest the sample mean. After one Lloyd update, distinct sites represented all twenty means in every simplex pilot. At \(\alpha=0.05\) the squared-distance condition in \cref{cor:pilot} holds on the far simplex (\(56{,}448>4{,}573\)) and misses the near simplex (\(220.5<515.3\)).

On 65,536 fresh validation observations per pilot, \cref{prop:certificate} was applied at simultaneous level 95 percent, with \(\lambda\varsigma^2d\) equal to \(1/8\), \(1/4\), \(1/2\) or \(1\) and \(\theta=2^{-j}/(2D)\) for \(0\le j\le6\). On the near simplex, validation switched the two merged fits to farthest-first and kept the other 48. Mean TV was \(0.05659\) (standard error \(0.00018\)), versus \(0.06035\) for eight starts and \(0.05648\) for farthest-first alone. The mean selected upper limit was \(0.229\). On the far simplex it was \(0.233\) at mean TV \(0.05655\). On D1 the certificate always chose one cell, with mean upper limit \(0.652\) and TV \(0.2002\). Squared response loss selected multicell fits, at mean TV \(0.1921\). The selected upper limits on the simplices are about four times the realized error, so \cref{prop:certificate} serves to rule out failed partitions.

\subsection{High-dimensional cover and unknown scale}\label{app:highdim}

\textbf{A trained baseline cover in 2,000 dimensions.} With \(S_0=135.76\) and \(\rho_b=100.87\), the test uses one scale. The means were \(0,0.15S_0e_1,6S_0e_1\) with weights \((0.50,0.35,0.15)\). Each of 50 independent runs used 512 training pairs per level, 30 references, 768 neighbours, clipping levels \(2\cdot5^{j/2}\), relative eigenvalue threshold \(10^{-3}\) and minimum group size 50. Every run returned two anchors, with mean maximum component-to-anchor distance \(0.1597S_0\). The test rejected 16 of 461,444 same-label comparisons and all 285,604 above \(S_0\). The \(4S_0\) pruning rule grouped the two nearby means.

\textbf{Unknown scale.} Give locations \(\sigma(-1,0,3)\) weights \((0.2,0.5,0.3)\). Integrating the nine Gaussian difference laws gives probability \(0.00999\) that the minimum norm of 36 independent pairs falls outside \([\sigma/1152,2\sigma]\). Among 2,048 independent blocks per scale, failure counts were 20, 18 and 16 for \(\sigma=10^{-7},1,10^5\), consistent with that probability at every scale.

\subsection{Complements and the complex pair test}\label{app:complements}

\textbf{Cumulant visibility.} In three dimensions take independent latent coordinates \(0.8A\), \(0.85B\) and zero, where \(A\) is a symmetric random sign and \(B\) takes \(0,\pm\sqrt3\) with probabilities \(2/3,1/6,1/6\). Add independent Gaussian noise making total covariance the identity. The six-component mixture has one visible direction through order four and both through order six, since the fourth cumulants are \(-2,0\) and the sixth are \(16,-6\) before scaling. Over 20 replicates with the population ranks, the median projector errors in operator norm at orders four and six were \(0.0498,0.0644\) at \(n=4096\) and \(0.00763,0.00615\) at \(n=131072\).

\textbf{Positive reconstruction from noisy moments.} Take means \((-2,0.12),\allowbreak(0,-0.20),\allowbreak(1.7,0.12)\) with weights \(0.25,0.45,0.30\). From \(20{,}000\) responses with conditional covariance \(2I\), estimate the second moment, select its leading directions and fit positive weights to Gaussian-corrected moments on a grid of spacing \(0.5\) and radius \(2.5\), with radial budget \((s,B)=(2,3)\). In the order \((r,m)=(0,0),(1,2),(1,4),(2,2),(2,4)\), latent \(W_1\) errors were \(1.1022\), \(0.4345\), \(0.4142\), \(0.4653\), \(0.4441\) and density TV errors were \(0.2819\), \(0.0622\), \(0.0430\), \(0.0619\), \(0.0310\). Moment-preserving compression reduced the degree-four fits from \(10\) to \(5\) atoms at rank one and from \(17\) to \(15\) at rank two, with \((W_1,\mathrm{TV})\) errors \((0.3603,0.0412)\) and \((0.4306,0.0301)\). This experiment uses one law and a coarser grid than \cref{prop:positive}.

\textbf{Single-scale complex pair test.} In \(d=12\) take two equally weighted means separated by \(S_0=\sqrt{17n}\). At each level retain the leading eigenvector of the symmetrized clipped cross-Gram matrix from 6,000 training pairs, using the known rank one. Four independent trains each applied \eqref{eq:pair-statistic} to 50,000 fresh pairs of each type at one scale, with the norm gate disabled. At \(n=2,4,8\) the false-positive counts were \(69462,10098,162\) out of 200,000. The matched false-negative counts were \(431,46,0\). For an exactly captured direction the null probability is \(\exp[-4^{-1-1/n}S_0^2/(2\sigma^2)]\), giving \(0.3470,0.0501,0.000809\). Both error rates decrease with the degree.

\bibliographystyle{emss}
\bibliography{references}
\end{document}